\documentclass[10pt,journal,compsoc]{IEEEtran}

\usepackage{amsfonts}
\usepackage{balance}
\usepackage{booktabs}
\usepackage{amsmath}
\usepackage{tabularx}
\usepackage{subfigure}
\usepackage{graphicx}
\usepackage{multirow}
\usepackage{rotating}
\usepackage{algorithmic}
\usepackage{footnote}
\usepackage{threeparttable}
\usepackage{color}

\ifCLASSOPTIONcompsoc
  \usepackage[nocompress]{cite}
\else
  \usepackage{cite}
\fi

\ifCLASSINFOpdf
\else
\fi

\begin{document}

\title{Ultra-Scalable Spectral Clustering and Ensemble Clustering}

\author{Dong~Huang,~\IEEEmembership{Member,~IEEE, }
        Chang-Dong~Wang,~\IEEEmembership{Member,~IEEE, }
        Jian-Sheng Wu,~\IEEEmembership{Member,~IEEE, }\\
        Jian-Huang~Lai,~\IEEEmembership{Senior Member,~IEEE, }
        and~Chee-Keong Kwoh,~\IEEEmembership{Senior Member,~IEEE}
\IEEEcompsocitemizethanks{\IEEEcompsocthanksitem D. Huang is with the College of Mathematics and Informatics, South China Agricultural University, Guangzhou, China, and also with the School of Computer Science and Engineering, Nanyang Technological University, Singapore. E-mail: huangdonghere@gmail.com.
\IEEEcompsocthanksitem C.-D. Wang and J.-H. Lai are with the School of Data and Computer Science,
Sun Yat-sen University, Guangzhou, China, and also with Guangdong Key Laboratory of Information Security Technology, Guangzhou, China, and also with Key Laboratory of Machine Intelligence and Advanced Computing, Ministry of Education, China.\protect\\
E-mail: changdongwang@hotmail.com, stsljh@mail.sysu.edu.cn.
\IEEEcompsocthanksitem J.-S. Wu is with the School of Information Engineering, Nanchang University, Nanchang, China. E-mail: jiansheng4211@gmail.com
\IEEEcompsocthanksitem C.-K. Kwoh is with the School of Computer Science and Engineering, Nanyang Technological University, Singapore. \protect\\
E-mail: asckkwoh@ntu.edu.sg.}
}

\markboth{IEEE Transactions on Knowledge and Data Engineering}%
{Shell \MakeLowercase{\textit{et al.}}: Bare Demo of IEEEtran.cls for Computer Society Journals}

\IEEEtitleabstractindextext{
\begin{abstract}

This paper focuses on scalability and robustness of spectral clustering for extremely large-scale datasets with limited resources. Two novel algorithms are proposed, namely, ultra-scalable spectral clustering (U-SPEC) and ultra-scalable ensemble clustering (U-SENC). In U-SPEC, a hybrid representative selection strategy and a fast approximation method for $K$-nearest representatives are proposed for the construction of a sparse affinity sub-matrix. By interpreting the sparse sub-matrix as a bipartite graph, the transfer cut is then utilized to efficiently partition the graph and obtain the clustering result. In U-SENC, multiple U-SPEC clusterers are further integrated into an ensemble clustering framework to enhance the robustness of U-SPEC while maintaining high efficiency. Based on the ensemble generation via multiple U-SEPC's, a new bipartite graph is constructed between objects and base clusters and then efficiently partitioned to achieve the consensus clustering result. It is noteworthy that both U-SPEC and U-SENC have nearly linear time and space complexity, and are capable of robustly and efficiently partitioning ten-million-level nonlinearly-separable datasets on a PC with 64GB memory. Experiments on various large-scale datasets have demonstrated the scalability and robustness of our algorithms. The MATLAB code and experimental data are available at {https://www.researchgate.net/publication/330760669}.

\end{abstract}

}

\maketitle

\IEEEdisplaynontitleabstractindextext

\IEEEpeerreviewmaketitle

\IEEEraisesectionheading{\section{Introduction}\label{sec:introduction}}

\IEEEPARstart{D}{ata} clustering is a fundamental problem in the field of data mining and machine learning \cite{jain10_survey}, whose purpose is to partition a set of objects into a certain number of homogeneous groups, each referred to as a cluster. Out of the large number of clustering algorithms that have been developed, spectral clustering in recent years has been gaining increasing attention due to its promising ability in dealing with nonlinearly separable datasets \cite{vonLuxburg2007,chen11_nystrom,cai15_LSC,he18_tcyb}. However, a critical limitation to conventional spectral clustering lies in its huge time and space complexity, which significantly restricts its application to large-scale problems.

Conventional spectral clustering typically consists of two time- and memory-consuming phases, namely, affinity matrix construction and eigen-decomposition. It generally takes $O(N^2d)$ time and $O(N^2)$ memory to construct the affinity matrix, and takes $O(N^3)$ time and $O(N^2)$ memory to solve the eigen-decomposition problem \cite{vonLuxburg2007}, where $N$ is the data size and $d$ is the dimension. As the data size $N$ increases, the computational burden of spectral clustering grows dramatically. For example, given a dataset with one million objects, the $N\times N$ affinity matrix alone will consume 7450.58 GB of memory (with each entry in the matrix stored as a double-precision value), which prohibitively exceeds the memory capacity of a general-purpose machine, not to mention the next phase of eigen-decomposition.

To alleviate the huge computational burden of spectral clustering, a commonly used strategy is to sparsify the affinity matrix and solve the eigen-decomposition problem by some sparse eigen-solvers \cite{vonLuxburg2007}. The matrix sparsification strategy can reduce the memory cost of storing the affinity matrix and facilitate the eigen-decomposition, but it still requires the computation of all entries in the original affinity matrix. Besides matrix sparsification, another widely-studied strategy is based on sub-matrix construction \cite{chen11_nystrom,cai15_LSC}. The Nystr\"{o}m method \cite{chen11_nystrom} randomly selects $p$ representatives from the original dataset and builds an $N\times p$ affinity sub-matrix. Cai et al. \cite{cai15_LSC}  extended the Nystr\"{o}m method and proposed the landmark based spectral clustering (LSC) method, which performs $k$-means on the dataset to get $p$ cluster centers as the $p$ representatives. However, these sub-matrix based spectral clustering methods \cite{chen11_nystrom,cai15_LSC} are typically restricted by an $O(Np)$ complexity bottleneck, which has been a critical hurdle for them to deal with extremely large-scale dataset where a larger $p$ is often desired for achieving better approximation \cite{cai15_LSC}. Moreover, the clustering results of these methods heavily rely on their one-shot approximation via the sub-matrix, which places an unstable factor on their clustering robustness. Despite the considerable efforts that have been made in recent years \cite{vonLuxburg2007,chen11_nystrom,cai15_LSC,he18_tcyb}, it remains a highly challenging problem how to enable spectral clustering to \emph{efficiently and robustly} cluster extremely large-scale datasets (which may even be nonlinearly separable) with rather limited computing resources.

In light of this, this paper focuses on scalability and robustness of spectral clustering for extremely larger-scale datasets. Specifically, this paper proposes two novel large-scale algorithms, namely, \textbf{u}ltra-\textbf{s}calable s\textbf{pec}tral clustering (U-SPEC) and \textbf{u}ltra-\textbf{s}calable \textbf{en}semble \textbf{c}lustering (U-SENC). In U-SPEC, a new hybrid representative selection strategy is presented to efficiently find a set of $p$ representatives, which reduces the time complexity of $k$-means based selection from $O(Npdt)$ to $O(p^2dt)$. Then, a fast approximation method for $K$-nearest representatives are designed to efficiently build a sparse sub-matrix with $O(Np^{\frac{1}{2}}d)$ time and $O(Np^{\frac{1}{2}})$ memory. With the sparse sub-matrix serving as the cross-affinity matrix, a bipartite graph is constructed between the dataset and the representative set. By taking advantage of the bipartite graph structure, the transfer cut \cite{CVPR12_Li} is utilized to solve the eigen-decomposition problem with $O(NK(K+k)+p^3)$ time, where $k$ is the number of clusters and $K$ is the number of nearest representatives. Finally, the $k$-means discretization is adopted to construct the clustering result from a set of $k$ eigenvectors, which takes $O(Nk^2t)$ time. As it generally holds that $k,K\ll p\ll N$, the time and space complexity of our U-SPEC algorithm are respectively dominated by $O(Np^{\frac{1}{2}}d)$ and $O(Np^\frac{1}{2})$.  Further, to go beyond the one-shot approximation of U-SPEC and provide better clustering robustness, the U-SENC algorithm is proposed by integrating multiple U-SPEC clusterers into a unified ensemble clustering framework, whose time and space complexity are respectively dominated by $O(Nmp^{\frac{1}{2}}d)$ and $O(Np^{\frac{1}{2}})$. Extensive experiments have been conducted on ten large-scale datasets (including five synthetic datasets and five real datasets), which have shown the superiority of our U-SPEC and U-SENC algorithms over the state-of-the-art in terms of both clustering robustness and scalability.

To summarize, the main contributions of this paper are listed as follows:

\begin{enumerate}
  \item A hybrid representative selection strategy is proposed to strike a balance between the efficiency of random selection and the effectiveness of $k$-means based selection.
  \item A fast approximation method for $K$-nearest representatives is designed,  which is time- and memory-efficient for constructing the sparse affinity sub-matrix between objects and representatives.
  \item A large-scale spectral clustering algorithm termed U-SPEC is developed based on efficient affinity sub-matrix construction and bipartite graph formulation. Its time and space complexity are dominated by $O(Np^{\frac{1}{2}}d)$ and $O(Np^\frac{1}{2})$ respectively.
  \item By integrating multiple U-SPEC clusterers, a new large-scale ensemble clustering algorithm termed U-SENC is developed, which significantly enhances the robustness of U-SPEC while maintaining high scalability. Its time and space complexity are dominated by $O(Nmp^{\frac{1}{2}}d)$ and $O(Np^\frac{1}{2})$ respectively.
\end{enumerate}

The notations that are used throughout the paper are summarized in Table~\ref{table:notations}. The rest of the paper is organized as follows. The related work on large-scale spectral clustering and ensemble clustering is reviewed in Section~\ref{sec:related_work}. The proposed U-SPEC and U-SENC algorithms are described in Section~\ref{sec:framework}. The experimental results are reported in Section~\ref{sec:experiment}. Finally, the paper is concluded in Section~\ref{sec:conclusion}.

\begin{table}[!t]
\caption{Summary of notations}
\label{table:notations}
\begin{center}
\begin{tabular}{c|l}
\toprule
$\mathcal{X}$   &A dataset of $N$ objects\\
$x_i$       &The $i$-th object in $\mathcal{X}$\\
$N$         &Number of objects in $\mathcal{X}$\\
$d$         &Dimension\\
$t$         &Number of iterations in the $k$-means method\\
$k$         &Number of clusters in the clustering result\\
$p'$        &Number of candidate representatives\\
$p$         &Number of representatives\\
$\mathcal{R}$ &The set of representatives\\
$r_i$       &The $i$-th representatives in $\mathcal{R}$\\
$\mathcal{RC}$ &The set of rep-clusters\\
$rc_i$      &The $i$-th rep-cluster in $\mathcal{RC}$\\
$y_i$       &Center of $rc_i$\\
$z_1$       &Number of rep-clusters in $\mathcal{RC}$\\
$z_2$       &Average number of objects in each rep-cluster\\
$K$         &Number of nearest representatives\\
$K'$        &Candidate neighborhood size around a representative\\
$Dist(x_i,rc_j)$    &Distance between object $x_i$ and rep-cluster $rc_j$\\
$G$         &A bipartite graph between $\mathcal{X}$ and $\mathcal{R}$\\
$B$         &Cross-affinity matrix of graph $G$.\\
$b_{ij}$    &The $(i,j)$-th entry of $B$\\
$E$         &Full affinity matrix of graph $G$\\
$L$         &Graph Laplacian of graph $G$\\
$D$         &Degree matrix of graph $G$\\
$u_i$       &The $i$-th eigenvector of graph $G$\\
$\gamma_i$  &The $i$-th eigenvalue of graph $G$\\
$G_{\mathcal{R}}$ &A small graph with $\mathcal{R}$ as the node set\\
$E_{\mathcal{R}}$         &Affinity matrix of graph $G_{\mathcal{R}}$\\
$L_{\mathcal{R}}$         &Graph Laplacian of graph $G_{\mathcal{R}}$\\
$D_{\mathcal{R}}$         &Degree matrix of graph $G_{\mathcal{R}}$\\
$v_i$       &The $i$-th eigenvector of graph $G_{\mathcal{R}}$\\
$\lambda_i$  &The $i$-th eigenvalue of graph $G_{\mathcal{R}}$\\
$D_{\mathcal{X}}$ &Diagonal matrix with its $(i,i)$-th entry being the\\
                    &sum of the $i$-th row of $B$\\
$T$         &Transition probability matrix\\
$\Pi$       &The ensemble of $m$ base clusterings\\
$\pi^i$       &The $i$-th base clustering in $\Pi$\\
$m$         &Number of base clusterings in $\Pi$\\
U-SPEC$_i$          &The clusterer to generate the $i$-th base clustering\\
$\mathcal{R}^i$ &The set of representatives in U-SPEC$_i$\\
$r_j^i$       &The $j$-th representatives in $\mathcal{R}^i$\\
$k^i$       &Number of clusters in $\pi^i$\\
$k_{min}$   &Minimum number of clusters in a base clustering\\
$k_{max}$   &Maximum number of clusters in a base clustering\\
$\tau$      &Random variable in $[0,1]$\\
$\mathcal{C}$ &Set of all clusters in $\Pi$\\
$C_i$       &The $i$-th cluster in $\mathcal{C}$\\
$k_c$       &Number of clusters in $\mathcal{C}$\\
$\tilde{G}$ &A bipartite graph between $\mathcal{X}$ and $\mathcal{C}$\\
$\tilde{B}$         &Cross-affinity matrix of graph $\tilde{G}$.\\
$\tilde{b}_{ij}$    &The $(i,j)$-th entry of $\tilde{B}$\\
$\tilde{u}_i$       &The $i$-th eigenvector of graph $\tilde{G}$\\
$\tilde{D}_{\mathcal{X}}$ &Diagonal matrix with its $(i,i)$-th entry being the\\
                    &sum of the $i$-th row of $\tilde{B}$\\
$G_{\mathcal{C}}$ &A small graph with $\mathcal{C}$ as the node set\\
$E_{\mathcal{C}}$         &Affinity matrix of graph $G_{\mathcal{C}}$\\
$L_{\mathcal{C}}$         &Graph Laplacian of graph $G_{\mathcal{C}}$\\
$D_{\mathcal{C}}$         &Degree matrix of graph $G_{\mathcal{C}}$\\
$\tilde{v}_i$       &The $i$-th eigenvector of graph $G_{\mathcal{C}}$\\
$\tilde{\lambda}_i$  &The $i$-th eigenvalue of graph $G_{\mathcal{C}}$\\
\bottomrule
\end{tabular}
\end{center}
\end{table}

\section{Related Work}
\label{sec:related_work}

In this section, we review the literature related to spectral clustering and ensemble clustering, with special emphasis on their recent large-scale extensions.

\subsection{Spectral Clustering}

Given a dataset of $N$ objects, conventional spectral clustering \cite{vonLuxburg2007} first computes an $N\times N$ affinity matrix, in which each entry corresponds to the similarity of two objects according to some similarity metrics. Then, the eigen-decomposition is performed on the graph Laplacian of the affinity matrix to obtain the $k$ eigenvectors associated with the first $k$ eigenvalues. By embedding the datasets into the low-dimensional space via the obtained $k$ eigenvectors, the final clustering can be achieved via $k$-means or some other discretization techniques \cite{vonLuxburg2007}.

Although spectral clustering has shown promising advantages in finding clusters of arbitrary shapes from complex data, its $O(N^3)$ time complexity and $O(N^2)$ space complexity significantly restrict its application in large-scale tasks. To alleviate the huge computational cost, some researchers sparsified the affinity matrix by considering $K$-nearest neighbors or $\epsilon$-neighbors, and then solved the eigen-decomposition problem by some sparse eigen-solvers \cite{vonLuxburg2007}, which, however, still requires the computation of all the entries in the original affinity matrix.

To avoid the computation of the full affinity matrix, the sub-matrix based approximation has emerged as a powerful and efficient tool for spectral clustering \cite{chen11_nystrom,cai15_LSC,he18_tcyb}. The Nystr\"{o}m approximation \cite{chen11_nystrom} randomly selects $p$ representatives from the dataset and builds an $N\times p$ affinity sub-matrix between the $N$ objects and the $p$ representatives. The sub-matrix construction takes $O(Npd)$ time and $O(Np)$ memory, which are much lower than the full affinity matrix construction. Although the random representative selection is very efficient, it is often unstable with regard to the quality of the selected representatives (see Fig.~\ref{fig:cmpSelStrategy_all3}). Moreover, while it has been shown that a larger $p$ is often favorable for better approximation \cite{chen11_nystrom}, the $O(Np)$ memory cost of the sub-matrix construction can still be a critical bottleneck when dealing with very large datasets. To address the potential instability of random selection, Cai and Chen \cite{cai15_LSC} proposed the LSC algorithm, which first partitions the dataset into $p$ clusters via $k$-means and then uses the $p$ cluster centers as the representatives. With the $N\times p$ sub-matrix constructed, they further sparsified it by preserving the $K$-nearest representatives for each row and zeroing out the others \cite{cai15_LSC}. Despite its progress over the previous methods, there are still three computational bottlenecks in the LSC algorithm \cite{cai15_LSC}. First, although the $k$-means based selection often provides a better set of representatives, it comes with the time complexity of $O(Npdt)$. Second, the calculation of all possible entries in the $N\times p$ sub-matrix is still required before the sparsification, which comes with the time complexity of $O(Npd)$. Third, the computation of the $K$-nearest representatives for all objects comes with the time complexity of $O(NpK)$. More recently, instead of using $p$ representatives, He et al. \cite{he18_tcyb} used Fourier features to represent data objects in kernel space, and built an $N\times p$ sub-matrix between the $N$ objects and the $p$ selected Fourier features, upon which the efficient eigen-decomposition can be performed. The time and space complexity of the fast explicit spectral clustering (FastESC) algorithm in \cite{he18_tcyb} are respectively $O(Npd+p^3)$ and $O(Np)$, which are still restricted by the $O(Np)$ complexity bottleneck. By incorporating a newly-designed positive Euler kernel, Wu et al. \cite{wu17_Euler} proposed the Euler spectral clustering (EulerSC) method and proved that the EulerSC is equivalent to the weighted positive Euler k-means, which can be iteratively optimized with $O(Ndkt)$ time. However, EulerSC can only use the positive Euler kernel to define the pair-wise similarity, and is not feasible for the general spectral clustering formulation with other similarity metrics. Moreover, its clustering robustness heavily relies on the proper selection of the Euler kernel parameter, which is difficult to find without prior knowledge.

\subsection{Ensemble Clustering}

Ensemble clustering has been a popular technique in recent years, which aims to combine multiple base clusterings into a better and more robust consensus clustering \cite{Fred05_EAC,iam_on11_linkbased,wu15_TKDE,Huang16_TKDE,huang17_tcyb,huang18_tsmcs,liu17_bioinformatics,liu17_tkde,huang14_weac,huang15_ecfg,strehl02,Zheng14_TKDD,Zhong15_pr}. The existing ensemble clustering algorithms can be mainly classified into three categories.

The first category is the pair-wise co-occurrence based methods \cite{Fred05_EAC,iam_on11_linkbased,yi_icdm12}. Fred and Jain \cite{Fred05_EAC} proposed the evidence accumulation clustering (EAC) method, which makes use of the co-association matrix by considering the frequency of pair-wise co-occurrence among multiple base clusterings. With the co-association matrix treated as the similarity matrix, the agglomerative clustering algorithms \cite{jain10_survey} were then performed to obtain the consensus clustering. Iam-On et al. \cite{iam_on11_linkbased} presented the weighted connected triple (WCT) method, which extends the EAC method by refining the co-association matrix via the common neighborhood information between clusters.

The second category is the graph partitioning based methods \cite{strehl02,fern04_bipartite,Huang16_TKDE,huang17_tcyb}. Strehl and Ghosh \cite{strehl02} transformed the multiple base clusterings into a hypergraph representation, based on which three graph partitioning based ensemble clustering methods were presented. Fern and Brodley \cite{fern04_bipartite} built a bipartite graph structure by treating both base clusters and data objects as graph nodes, and then partitioned the graph via the METIS algorithm \cite{karypis98_METIS}.

The third category is the median partition based methods \cite{franek13_pr,huang15_ecfg}, which cast the ensemble clustering problem into an optimization problem that aims to find a median  clustering (or partition) by maximizing the similarity between this clustering and the multiple base clusterings. Franek and Jiang \cite{franek13_pr} formulated the median partition problem into a Euclidian median problem and solved it by the Weiszfeld algorithm \cite{Weiszfeld09}. Huang et al. \cite{huang15_ecfg} cast the median partition problem into a binary linear programming problem and solved it by the factor graph model.

These ensemble clustering algorithms have shown their advantages in improving clustering accuracy and robustness. However, due to the efficiency bottleneck, most of them are not suitable for very large-scale applications. Recently some efforts have been made to (partially) address the scalability problem for ensemble clustering. To reduce the problem size, Huang et al. \cite{Huang16_TKDE} exploited the microcluster representation, which maps the $N$ data objects onto $N'$ microclusters ($N'\ll N$). Then, the set of microclusters are treated as the primitive objects, based on which two novel algorithms, i.e., the probability trajectory accumulation (PTA) and the probability trajectory based graph partitioning (PTGP), are proposed. Wu et al. \cite{wu15_TKDE} transformed the ensemble clustering problem into a $k$-means based consensus clustering (KCC) framework, which significantly facilitated the computation of the consensus function. Liu et al. \cite{liu17_tkde} proved that the spectral clustering of the co-association matrix is equivalent to an instance of weighted $k$-means clustering, and presented the spectral ensemble clustering (SEC) algorithm. While there are two phases in ensemble clustering (i.e., ensemble generation and consensus function), these algorithms \cite{Huang16_TKDE,wu15_TKDE,liu17_tkde} generally focus on the efficiency of the consensus function. In ensemble generation, they mostly exploited $k$-means to produce $m$ base clusterings \cite{Huang16_TKDE,wu15_TKDE,liu17_tkde}. Note that the time complexity of ensemble generation by $k$-means is $O(Nmdkt)$, which can still be computationally expensive when dealing with very large-scale datasets. Moreover, the performance of $k$-means may significantly deteriorate when handling nonlinearly separable datasets, which has a critical influence on the robustness of the ensemble clustering algorithms. Unlike the common practice that typically exploits multiple $k$-means clusterers as base clusterers, the proposed U-SENC algorithm integrates a diverse set of large-scale U-SPEC clusterers into a highly efficient ensemble clustering framework, which for the first time, to our knowledge, simultaneously tackles the scalability and nonlinear separability issues in both the ensemble generation and consensus function phases in ensemble clustering.

\section{Proposed Framework}
\label{sec:framework}

In this section, we describe the proposed U-SPEC and U-SENC algorithms in Sections~\ref{sec:USPEC} and \ref{sec:USENC}, respectively.

\subsection{Ultra-Scalable Spectral Clustering (U-SPEC)}
\label{sec:USPEC}

To deal with extremely large-scale datasets, the proposed U-SPEC algorithm complies with the sub-matrix based formulation \cite{chen11_nystrom,cai15_LSC} and aims to break through the efficiency bottleneck of previous algorithms via three phases.
Specifically, in the first phase, we present a hybrid representative selection strategy to strike a balance between the efficiency of the random selection and the effectiveness of the $k$-means based selection.
In the second phase, we develop a coarse-to-fine method to efficiently approximate the $K$-nearest representatives for each data object, and construct a sparse affinity sub-matrix between the $N$ objects and the $p$ representatives. In the third phase, the $N\times p$ sub-matrix is interpreted as a bipartite graph, which can be efficiently partitioned to obtain the final clustering result. These three phases of U-SPEC will be described in Sections~\ref{sec:hybrid_select}, \ref{sec:approx_knn}, and \ref{sec:bipartite_partition}, respectively.

\subsubsection{Hybrid Representative Selection}
\label{sec:hybrid_select}

\begin{figure}[!tb]
\begin{center}
{\subfigure[]
{\includegraphics[width=0.281\linewidth]{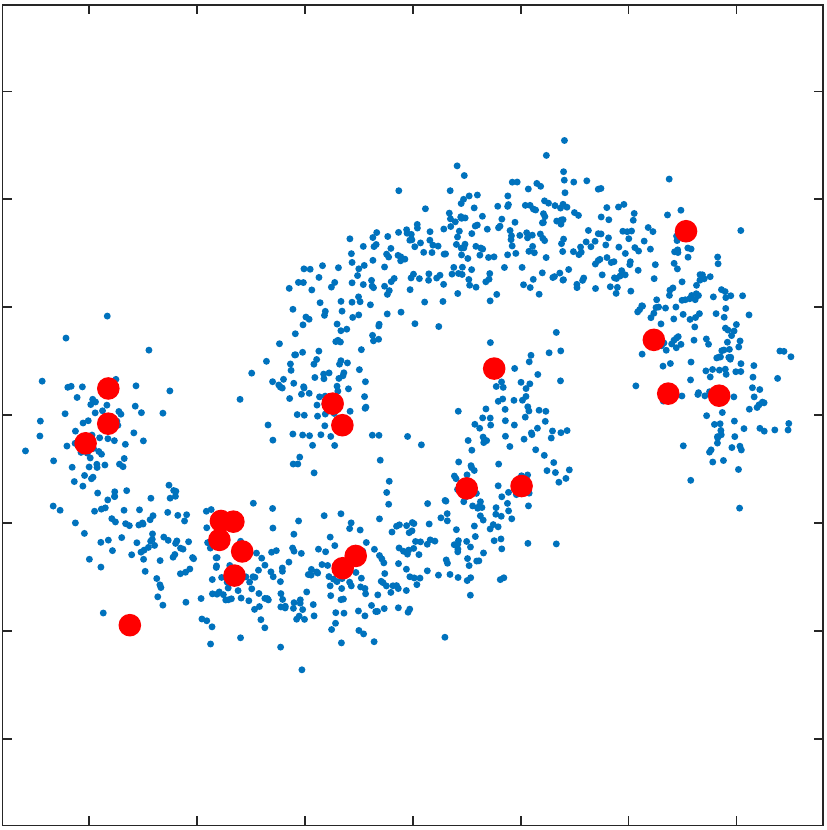}\label{fig:cmpSelStrategy_random}}}
{\subfigure[]
{\includegraphics[width=0.281\linewidth]{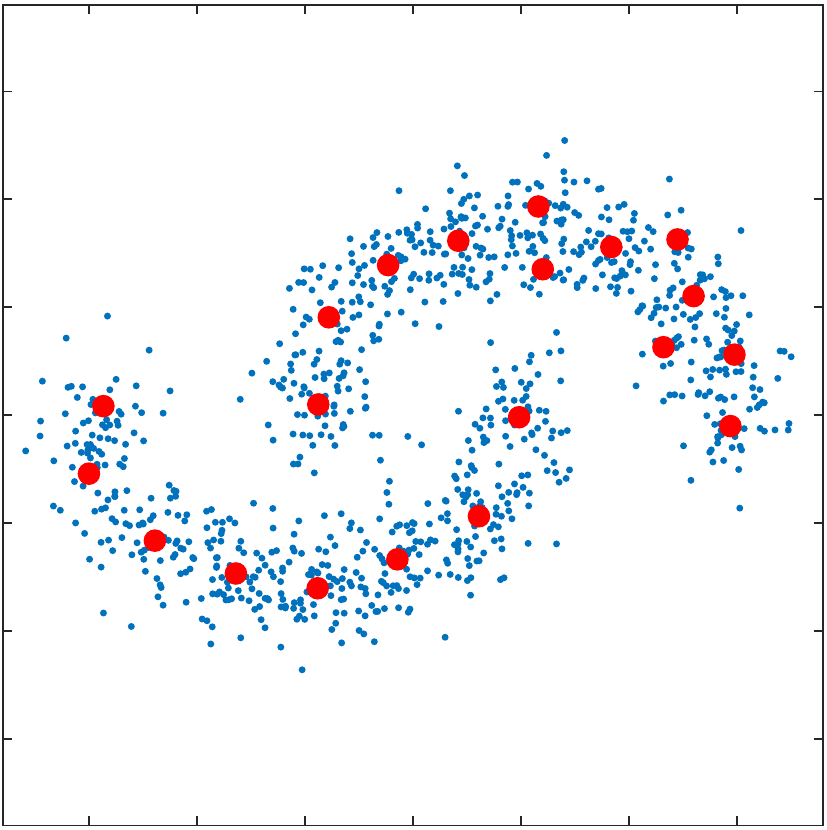}}}
{\subfigure[]
{\includegraphics[width=0.281\linewidth]{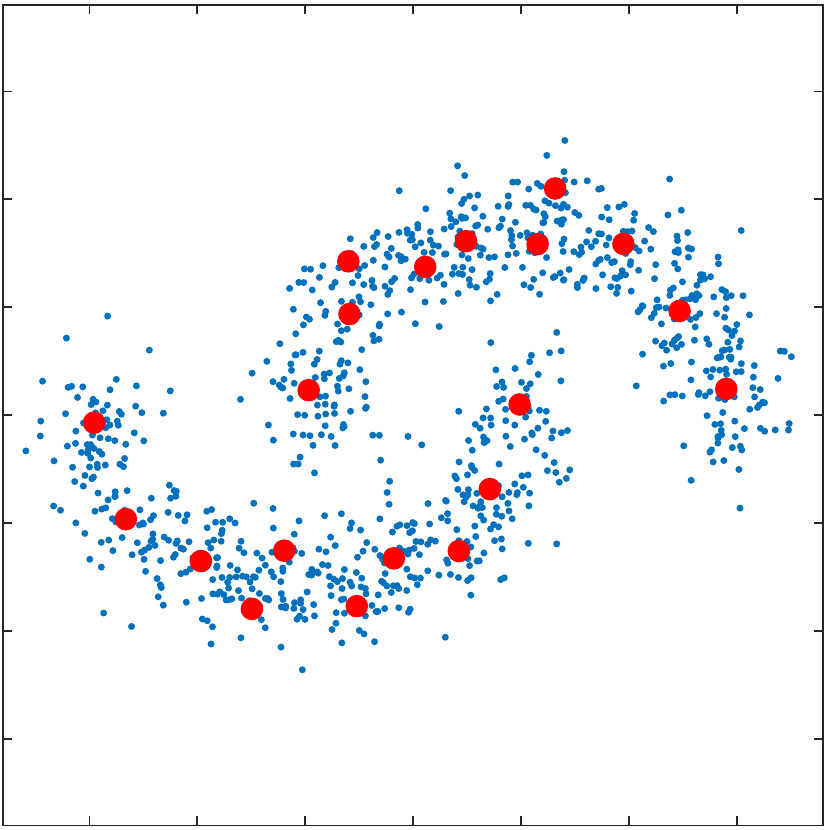}}}
\caption{Comparison of the representatives produced by (a) random selection, (b) $k$-means based selection, and (c) hybrid selection.}
\label{fig:cmpSelStrategy_all3}
\end{center}
\end{figure}

Let $\mathcal{X}=\{x_1, x_2, \cdots, x_N\}$ denote a dataset with $N$ objects, where $x_i\in \mathbb{R}^d$ is the $i$-th object and $d$ is the dimension. To capture the relationship between all objects in $\mathcal{X}$, an $N\times N$ affinity matrix can be constructed in conventional spectral clustering \cite{vonLuxburg2007}, which consumes $O(N^2d)$ time and $O(N^2)$ memory and is not feasible for large-scale datasets. To avoid the computation of the full affinity matrix, the sub-matrix representation is often adopted in the literature of large-scale spectral clustering \cite{chen11_nystrom,cai15_LSC}. The sub-matrix representation generally exploits a set of representatives to encode the overall structure of the dataset. These representatives play a crucial role in the sub-matrix representation, and can be selected by random selection \cite{chen11_nystrom} or $k$-means based selection \cite{cai15_LSC}. Though the random selection strategy \cite{chen11_nystrom} is highly efficient, it suffers from the inherent randomness and may lead to a set of low-quality representatives (see Fig.~\ref{fig:cmpSelStrategy_random}). To deal with the instability of random selection, the $k$-means based selection \cite{cai15_LSC} first groups the entire dataset into $p$ clusters via $k$-means and then uses the $p$ cluster centers as the representatives. However, the $k$-means based selection brings in an extra time cost of $O(Npdt)$, which restricts its feasibility for very large-scale datasets.

\begin{figure}[!tb]
\begin{center}
{\subfigure[]
{\includegraphics[width=0.281\linewidth]{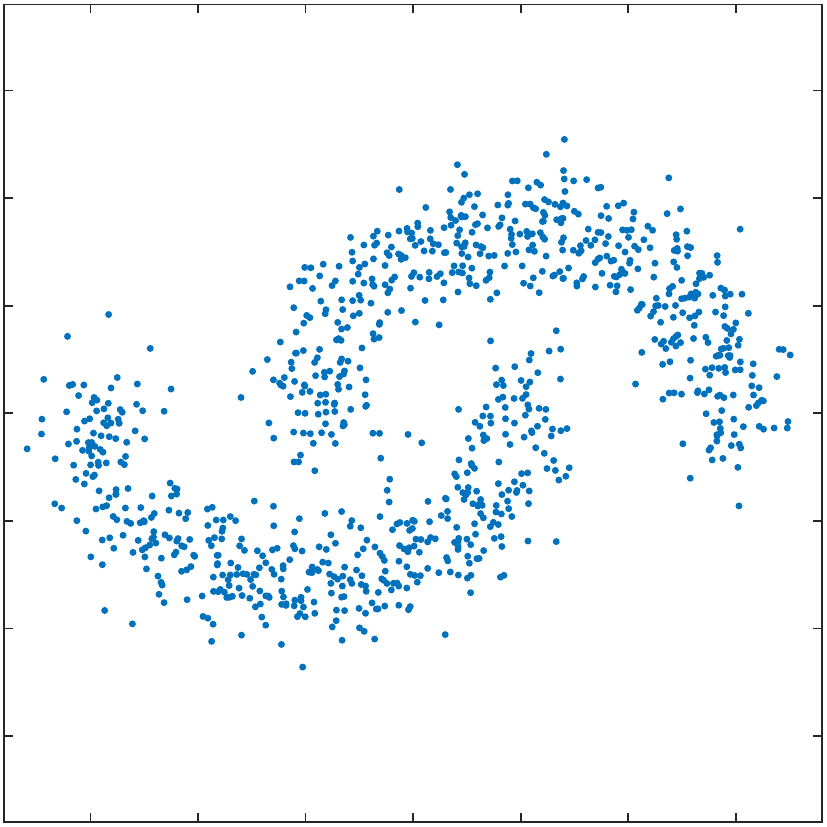}}}
{\subfigure[]
{\includegraphics[width=0.281\linewidth]{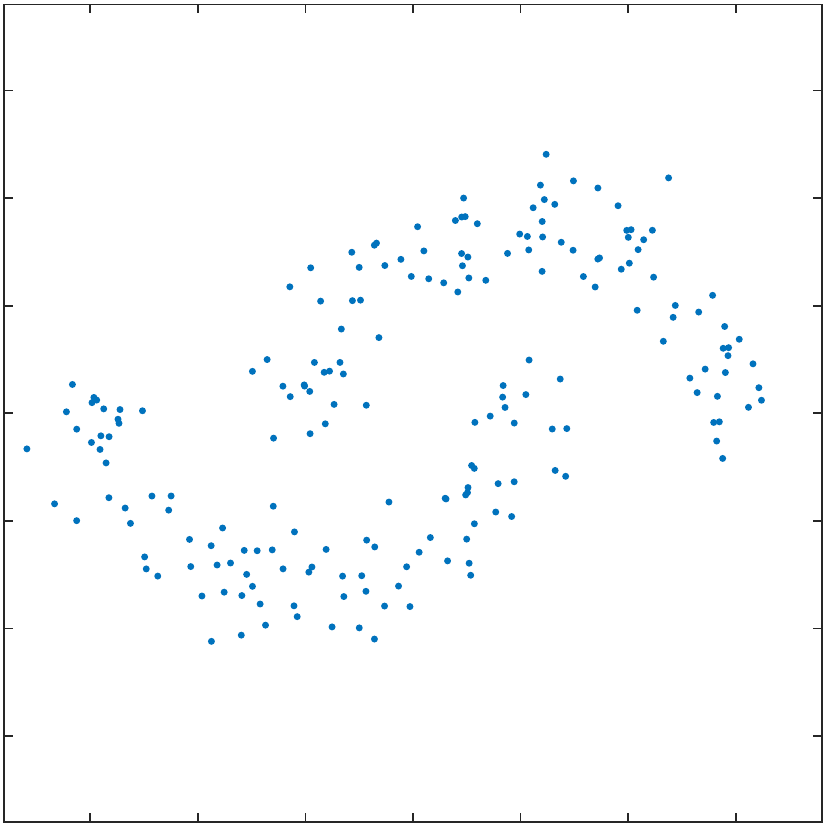}}}
{\subfigure[]
{\includegraphics[width=0.281\linewidth]{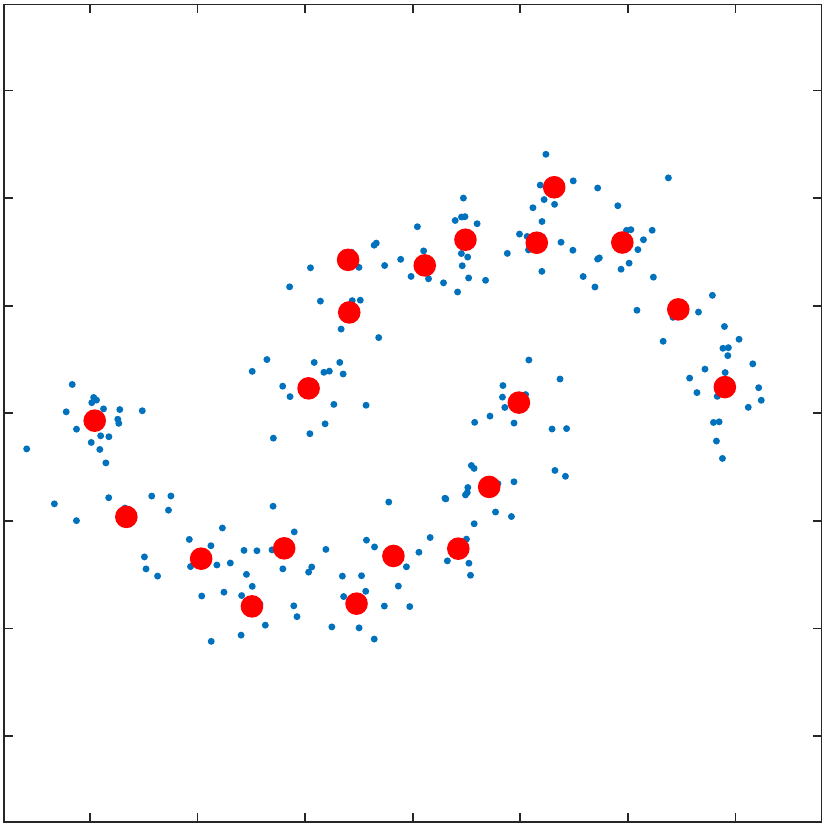}}}
\caption{Illustration of hybrid representative selection. (a) The dataset. (b) Randomly select $p'$ candidates ($p'>p$). (c) Obtain $p$ representatives from $p'$ candidates via $k$-means.}
\label{fig:hybridSelStrategy}
\end{center}
\end{figure}

In this paper, we propose a hybrid representative selection strategy, which is designed to find a balance between the efficiency of random selection and the effectiveness of $k$-means based selection. The process of the hybrid representative selection strategy is illustrated in Fig.~\ref{fig:hybridSelStrategy}. Different from the $k$-means based selection which attempts to cluster the entire dataset even when the data size $N$ is extremely large, the proposed hybrid selection strategy first randomly samples a set of $p'$ candidate representatives such that $p<p'\ll N$. Then, upon the $p'$ candidates, we perform the $k$-means method to obtain $p$ clusters and exploit the $p$ cluster centers as the set of representatives. Empirically, the number of candidates $p'$ is suggested to be several times larger than $p$, e.g., $p'=10p$, so as to provide enough candidates while still keeping $p'$ much smaller than $N$ in large-scale datasets. Formally, we denote the set of selected representatives as
\begin{align}
\mathcal{R} = \{r_1, r_2, \cdots, r_p\},
\end{align}
where $r_i$ is the $i$-th representative in $\mathcal{R}$.

By introducing an intermediate stage of random \emph{pre-sampling}, the computational complexity of the $k$-means based selection is reduced from $O(Npdt)$ to $O(p^2dt)$. As illustrated in Fig.~\ref{fig:cmpSelStrategy_all3}, the set of representatives produced by the hybrid selection can better reflect the data distribution than the random selection while requiring much less computational cost than the $k$-means based selection. To discuss this in more detail, quantitative evaluation of the performance of the proposed hybrid selection strategy against random selection and $k$-means based selection will be provided in Section~\ref{sec:cmpSelStrat}.

\subsubsection{Approximation of $K$-Nearest Representatives}
\label{sec:approx_knn}

With the $p$ representatives obtained, the next objective is to encode the pair-wise relationship of the entire dataset via the small set of representatives.

In the sub-matrix formulation of the Nystr\"{o}m algorithm \cite{chen11_nystrom}, the construction of the $N\times p$ affinity sub-matrix between objects and representatives takes $O(Npd)$ time and $O(Np)$ memory, which is the main efficiency bottleneck of the overall algorithm \cite{chen11_nystrom}. Given a dataset with ten million objects and a set of one thousand representatives, the storage of the $N\times p$ sub-matrix alone takes 74.51GB of memory, while the later manipulations of the sub-matrix even require more memory consumption. Cai and Chen \cite{cai15_LSC} proposed to sparsify the $N\times p$ affinity matrix by $K$-nearest representatives (with $K\ll p$), which, however, still requires the computation of all the distances between the $N$ objects and the $p$ representatives. Moreover, besides the calculation of the total of $Np$ entries, the sparsification step also consumes $O(NpK)$ time \cite{cai15_LSC}.

Before introducing our facilitation strategy, we first investigate the characteristics of the sparse sub-matrix between $N$ objects and $p$ representatives, where each object is only connected to its $K$-nearest representatives. It is obvious that there are $K$ non-zero entries in each row of the matrix, and $NK$ non-zero entries in the entire matrix. Assume we have $p=1,000$ and $K=5$, the proportion of the non-zero entries in the matrix will be $0.5\%$. However, to exactly identify such a small proportion of useful entries via $K$-nearest representatives, the entire matrix should first be calculated, which unfortunately consists of $99.5\%$ of \emph{intermediate} entries. To break the efficiency bottleneck, the key problem here is how to significantly reduce the calculation of these intermediate entries when building the sub-matrix with $K$-nearest representatives.

In this section, our aim is to alleviate the computational cost of the exact $K$-nearest representative calculation \cite{cai15_LSC} by designing a time- and memory-efficient approximation method. Though the $K$-nearest representative approximation problem and the classical $K$-nearest neighbor ($K$-NN) approximation problem \cite{Dong11_www,zhang13_knn,Bryant18_tkde} have some characteristics in common, they are faced with very different computational issues in actual applications. Different from the conventional $K$-NN approximation scenarios, which mostly deal with a general graph with an $N\times N$ affinity matrix, our aim here is to find the $K$-nearest representatives in a heavily imbalanced bipartite graph with an $N\times p$ affinity sub-matrix, where $p$ is generally far smaller than $N$. This imbalanced nature is crucial to our $K$-nearest representative approximation problem. On the one hand, it makes the conventional $K$-NN approximation methods \cite{Dong11_www,zhang13_knn,Bryant18_tkde} (which are typically designed for general graphs with $N\times N$ affinity matrices) inappropriate here. On the other hand, it may as well contribute to the design of our $K$-nearest representative approximation strategy. To take advantage of the imbalanced structure, it is intuitive to pre-process the graph on the side of the $p$ representatives and minimize the computation on the other side of the $N$ objects.

In particular, we present a new $K$-nearest representative approximation method based on the coarse-to-fine mechanism, and build the sparse affinity sub-matrix with $O(Np^{\frac{1}{2}}d)$ complexity. The main idea of our $K$-nearest representative approximation is to first find the nearest \emph{region}, then find the nearest representative (denoted as $r_l$) in the nearest region, and finally find the $K$-nearest representatives in the neighborhood of $r_l$. To efficiently implement the approximation, two preprocessing steps are required, that is
\begin{itemize}
  \item \textbf{Pre-step 1.} The set of representatives are grouped into $z_1$ rep-clusters via $k$-means (with $z_1\ll p$). The time complexity is $O(pz_1dt)$.
  \item \textbf{Pre-step 2.} For each representative in $\mathcal{R}$, its $K'$-nearest neighbors are computed and stored (with $K'>K$). The time complexity is $O(p^2(d+K'))$.
\end{itemize}

\begin{figure}[!tb]
\begin{center}
{\subfigure[]
{\includegraphics[width=0.2426\linewidth]{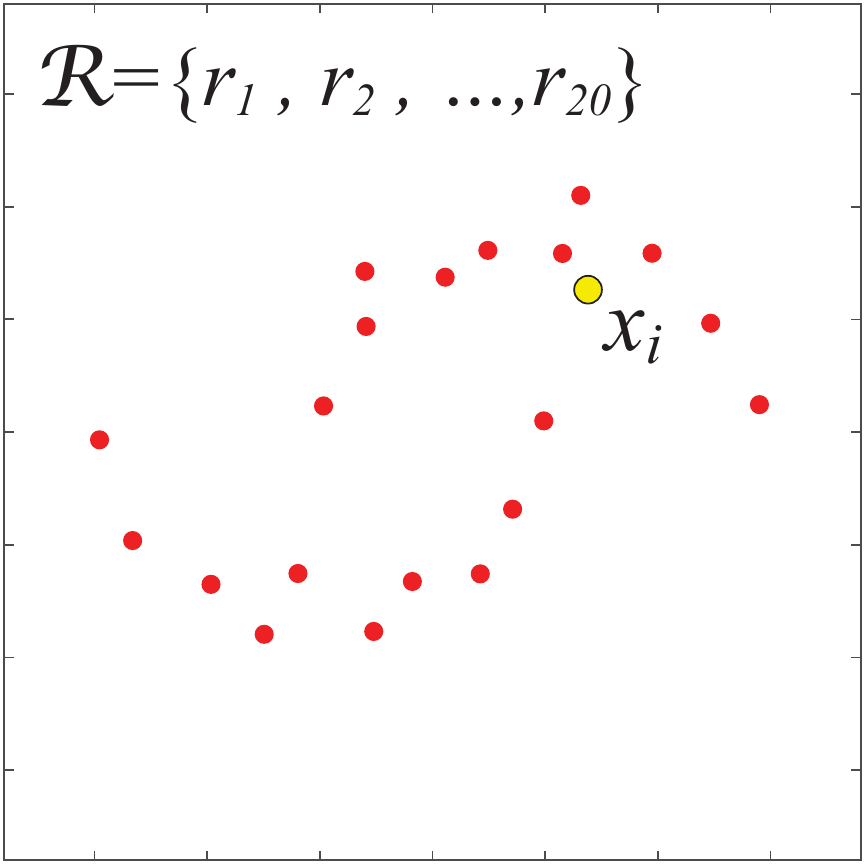}\label{fig:example_ApproxKNN_a}}}
{\subfigure[]
{\includegraphics[width=0.2426\linewidth]{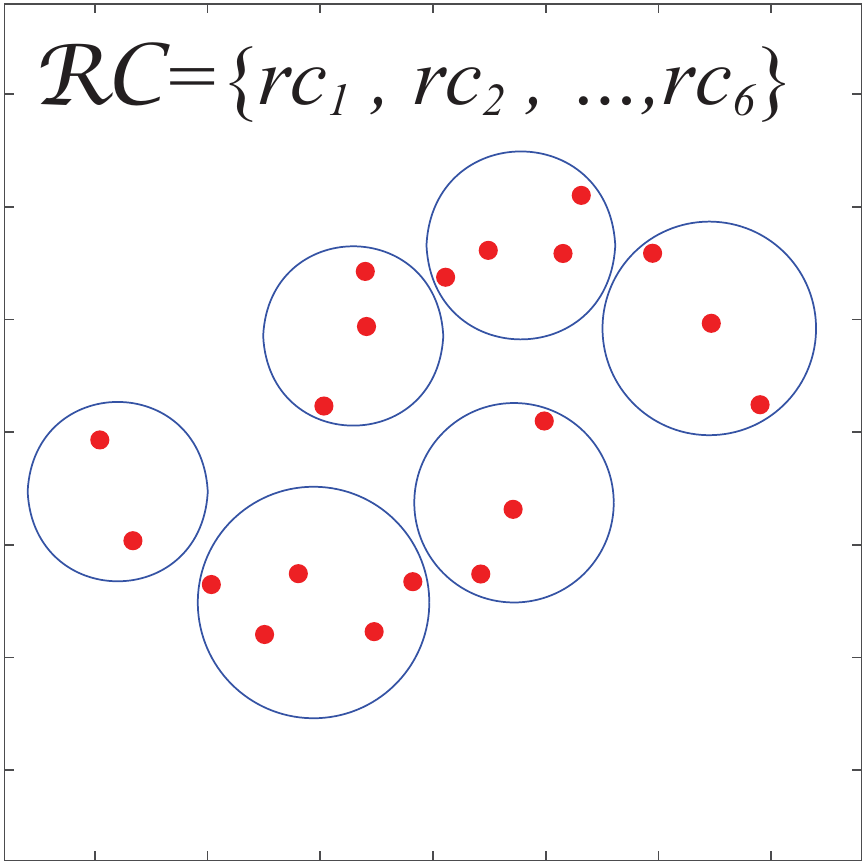}\label{fig:example_ApproxKNN_b}}}
{\subfigure[]
{\includegraphics[width=0.2426\linewidth]{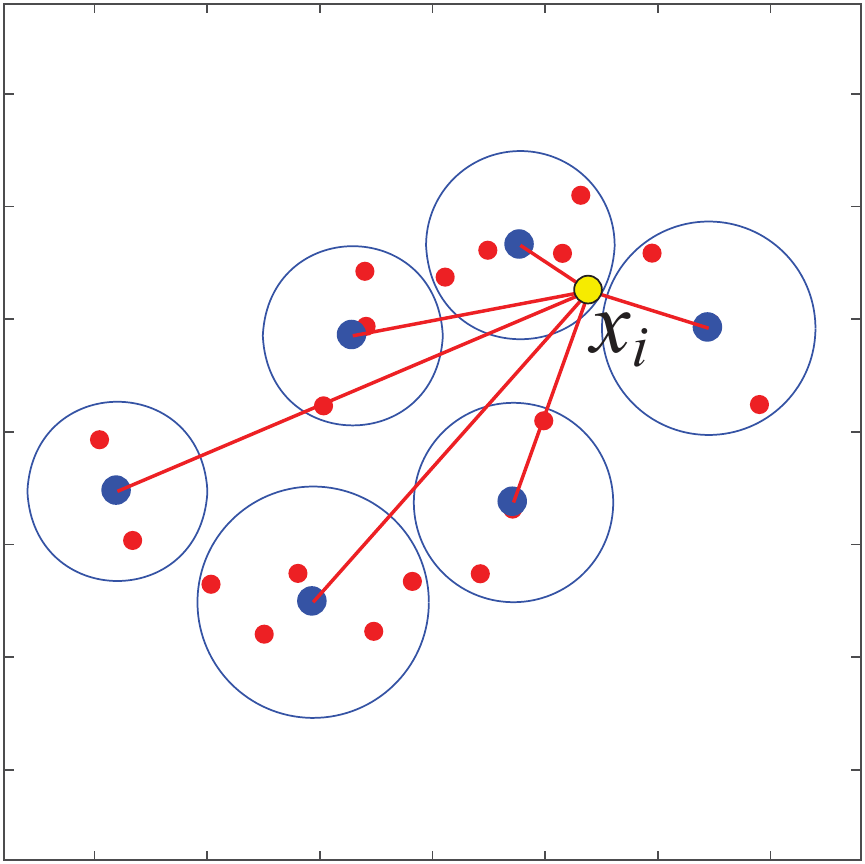}\label{fig:example_ApproxKNN_c}}}
{\subfigure[]
{\includegraphics[width=0.2426\linewidth]{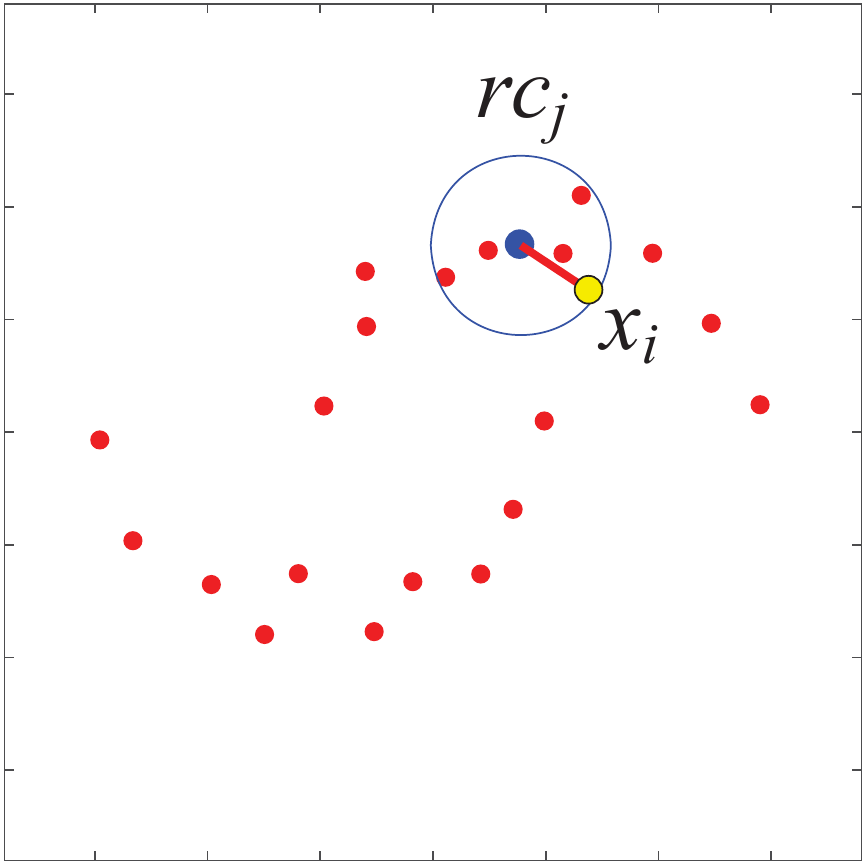}\label{fig:example_ApproxKNN_d}}}
{\subfigure[]
{\includegraphics[width=0.2426\linewidth]{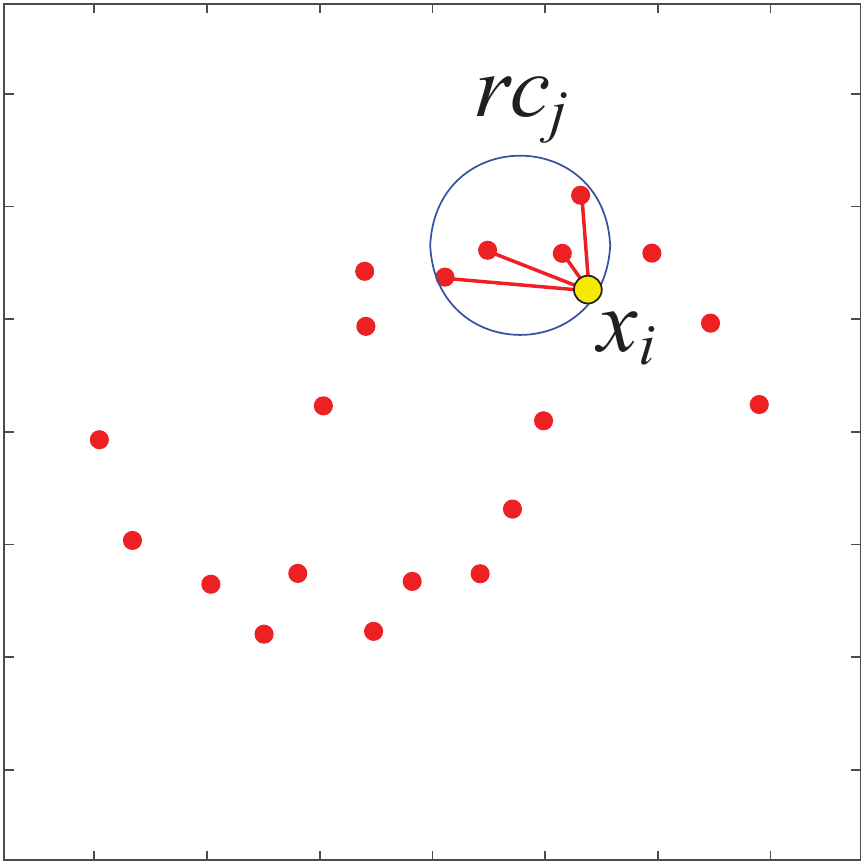}\label{fig:example_ApproxKNN_e}}}
{\subfigure[]
{\includegraphics[width=0.2426\linewidth]{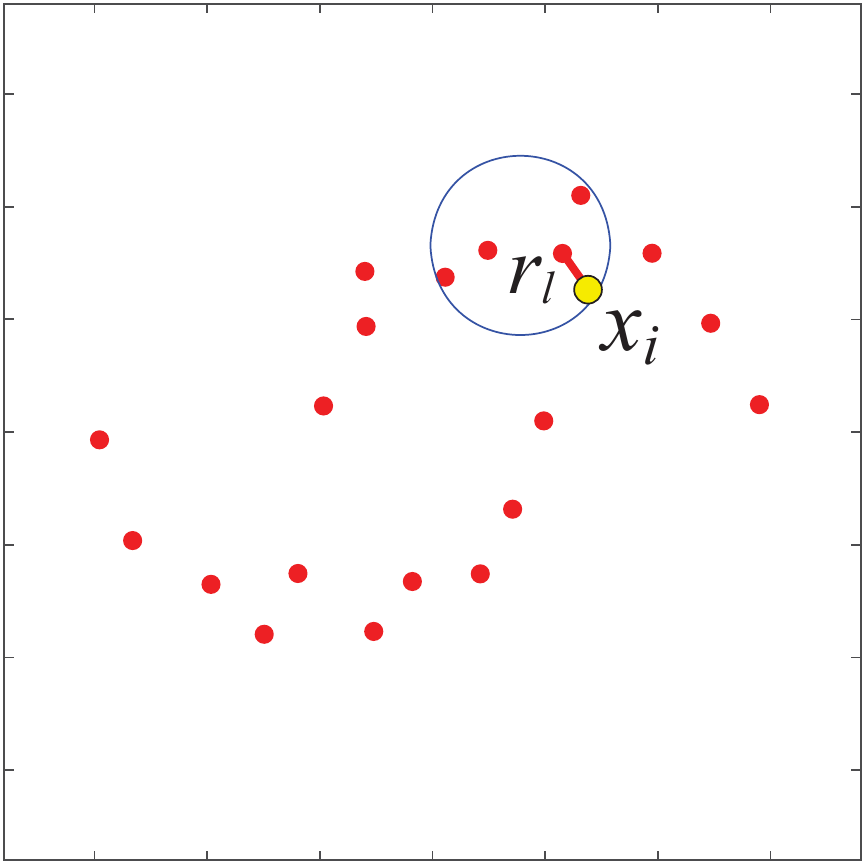}\label{fig:example_ApproxKNN_f}}}
{\subfigure[]
{\includegraphics[width=0.2426\linewidth]{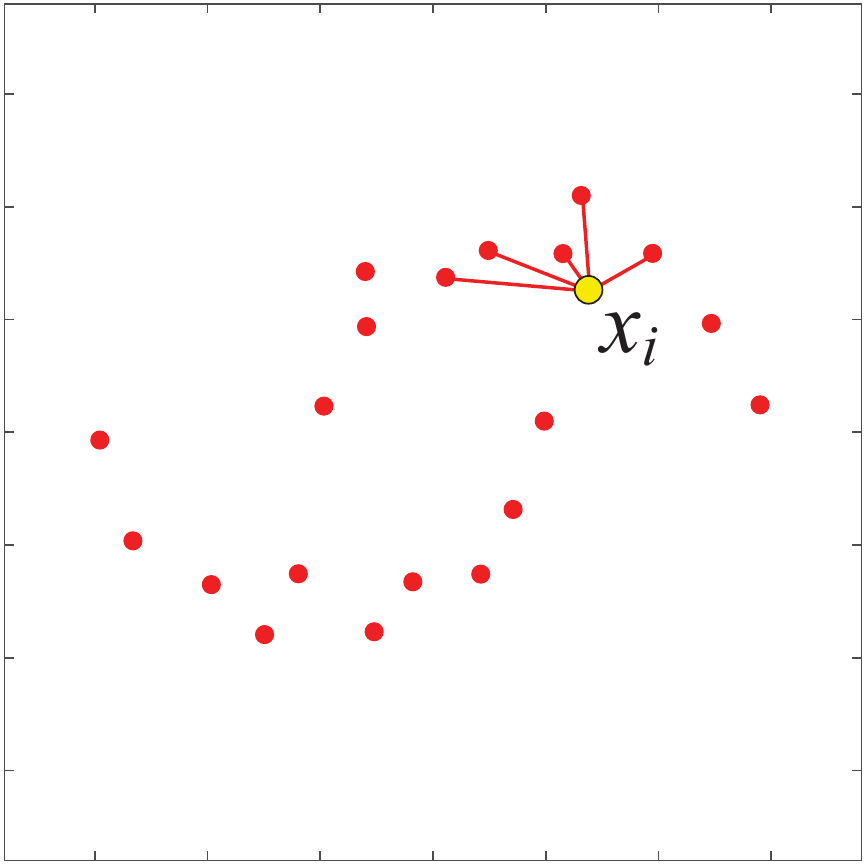}\label{fig:example_ApproxKNN_g}}}
{\subfigure[]
{\includegraphics[width=0.2426\linewidth]{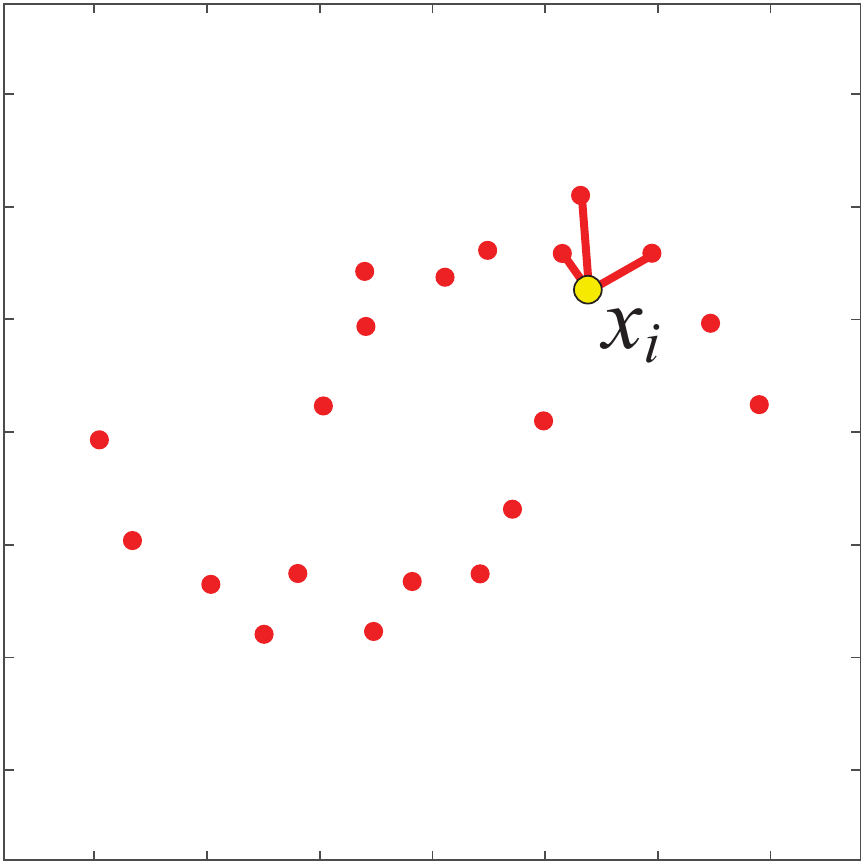}\label{fig:example_ApproxKNN_h}}}
\caption{Approximate $K$-nearest representatives. (a) The representative set $\mathcal{R}$ and an object $x_i\in \mathcal{X}$. (b) Partition the representatives into several rep-clusters. (c) Compute the distances between $x_i$ and all the rep-cluster centers. (d) Find the nearest rep-cluster $rc_j$. (e) Compute the distances between $x_i$ and all the representatives in $rc_j$. (f) Find the nearest $r_l\in rc_j$. (g) Compute the distances between $x_i$ and the representatives in the $K'$-nearest neighborhood of $r_l$ ($K'>K$). (h) Obtain the approximate $K$-nearest representatives ($K=3$).}
\label{fig:example_ApproxKNN}
\end{center}
\end{figure}

In pre-step 1, each rep-cluster consists of a certain number of representatives, and can be regarded as a local region of the representative set (see Fig.~\ref{fig:example_ApproxKNN_b}). Formally, the obtained $z_1$ rep-clusters are denoted as
\begin{align}
\mathcal{RC}=\{rc_1, rc_2, \cdots, rc_{z_1}\},
\end{align}
where $rc_i$ is the $i$-th rep-cluster in $\mathcal{RC}$. Given an object $x_i\in\mathcal{X}$ and a rep-cluster $rc_j\in \mathcal{RC}$, their distance is defined as the distance between $x_i$ and the center of $rc_j$. That is
\begin{align}
Dist(x_i,rc_j)=\|x_i-y_j\|,\\
y_j = \frac{1}{|rc_j|}\sum_{r_l\in rc_j}r_l,
\end{align}
where $|rc_j|$ denotes the number of representatives in the rep-cluster $rc_j$ and $\|x_i-y_j\|$ computes the Euclidean distance between two vectors $x_i$ and $y_j$.

With the distance between objects and rep-clusters defined, for each object $x_i\in\mathcal{X}$, we approximately find its $K$-nearest representatives according to three main steps:
\begin{description}
  \item[\textbf{Step 1}] Find the nearest rep-cluster of $x_i$, denoted as $rc_j$.
  \item[\textbf{Step 2}] Find the nearest representative of $x_i$ inside the rep-cluster $rc_j$, denoted as $r_l$.
  \item[\textbf{Step 3}] Out of $r_l$ and its $K'$-nearest neighbors, find the $K$-nearest representatives of $x_i$.
\end{description}

More details are illustrated in Fig.~\ref{fig:example_ApproxKNN}. For a dataset with $N$ objects, the time cost of step 1 is $O(Nz_1d)$. The time cost of step 2 is $O(Nz_2d)=O(N({p}/{z_1})d)$, where $z_2 = {p}/{z_1}$ denotes the average size of the rep-clusters. The time cost of step 3 is $O(NK'd + NK'K)$. It is obvious that $z_1 + {p}/{z_1}$ reaches its minimum when $z_1=z_2={p}^{\frac{1}{2}}$. Thus, to minimize the cost, $z_1=\lfloor{p}^{\frac{1}{2}}\rfloor$ is used in this work, where $\lfloor\cdot\rfloor$ denotes the floor of a value. The candidate neighborhood size $K'$ is suggested to be several times larger than $K$, which can be set to $K'=10K$ in practice. Then, the total time complexity of the $K$-nearest representative approximation is $O(Nz_1d+N({p}/{z_1})d+NK'd + NK'K)$, which can be re-written as $O(N(p^{\frac{1}{2}}d+Kd+K^2))$. As $K\ll p \ll N$, the dominant term in the complexity is $O(Np^{\frac{1}{2}}d)$.

With the $K$-nearest representatives of each object obtained, a sparse $N\times p$ affinity sub-matrix can thereby be constructed. In this paper, the Gaussian kernel is used as the similarity kernel. Thus the sparse affinity sub-matrix can be represented as
\begin{align}
B &= \{b_{ij}\}_{N\times p},\\
b_{ij} &= \begin{cases}\exp{(-\frac{\|x_i-r_j\|^2}{2\sigma^2})},&\text{if $r_j\in N_K(x_i)$},\label{eq:gaussian_kernel}\\
0,&\text{otherwise,}
\end{cases}
\end{align}
where $N_K(x_i)$ denotes the set of $K$-nearest representatives of $x_i$ and the kernel parameter $\sigma$ is set to the average Euclidean distance between the objects and their $K$-nearest representatives. Note that $B$ is a sparse matrix which only contains $NK$ non-zero entries.

\subsubsection{Bipartite Graph Partitioning}
\label{sec:bipartite_partition}

\begin{figure}[!tb]
\begin{center}
{
{\includegraphics[width=0.48\linewidth]{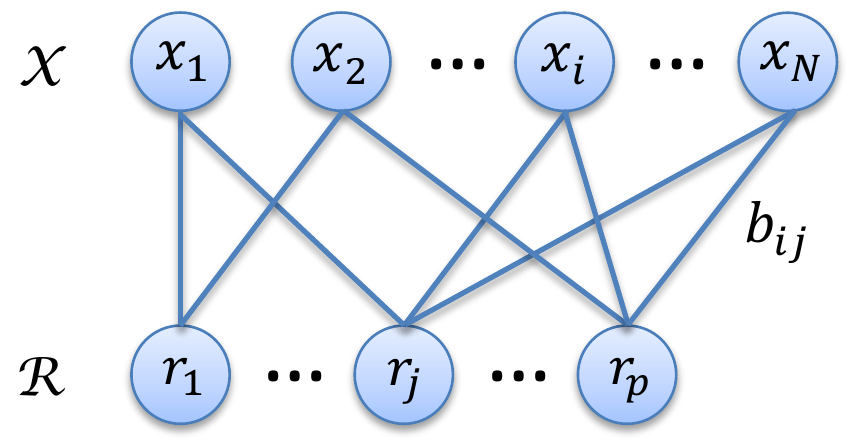}}}
\caption{Illustration of the bipartite graph $G$.}
\label{fig:BGP}
\end{center}
\end{figure}

The affinity sub-matrix $B$ reflects the relationship between the objects in $\mathcal{X}$ and the representatives in $\mathcal{R}$, which can be naturally interpreted as a bipartite graph $G=\{\mathcal{X}, \mathcal{R}, B\}$, where $\mathcal{X}\cup\mathcal{R}$ is the node set and $B$ is the cross-affinity matrix (as shown in Fig.~\ref{fig:BGP}). By taking advantage of the bipartite graph structure, the transfer cut \cite{CVPR12_Li} can thereby be used to efficiently partition the graph and achieve the final clustering result.

To start, if we view the graph $G$ as a general graph with $N+p$ nodes, then its full affinity matrix can be denoted as
\begin{align}
E=\left[\begin{matrix}
   0 & B^\top \\
   B & 0
  \end{matrix}\right].
\end{align}
Spectral clustering seeks to partition the graph by solving the following generalized eigen-problem \cite{jm00_ncut}:
\begin{align}
\label{eq:general_eigen_problem}
Lu=\gamma Du,
\end{align}
where $L=D-E$ is the graph Laplacian and $D\in \mathbb{R}^{(N+p)\times (N+p)}$ is the degree matrix. By treating $G$ as a general graph, it takes $O((N+p)^3)$ time to solve the eigen-problem~(\ref{eq:general_eigen_problem})  \cite{golub2012matrix}, which is not computationally feasible for very large-scale datasets.

By exploiting the bipartite structure, we resort to the transfer cut \cite{CVPR12_Li} to reduce the eigen-problem~(\ref{eq:general_eigen_problem}) on the graph $G$ (with $N+p$ nodes) to an eigen-problem on a much smaller graph $G_{\mathcal{R}}$ (with $p$ nodes). Specifically, the graph $G_{\mathcal{R}}$ is constructed as $G_{\mathcal{R}}=\{\mathcal{R}, E_{\mathcal{R}}\}$, where $\mathcal{R}$ is the node set, $E_{\mathcal{R}}=B^\top {D_{\mathcal{X}}}^{-1}B$ is the affinity matrix (whose computation takes $O(NK^2)$ time), and $D_{\mathcal{X}}\in \mathbb{R}^{N\times N}$ is a diagonal matrix with its $(i,i)$-th entry being the sum of the $i$-th row of $B$. Let $L_{\mathcal{R}}=D_{\mathcal{R}}-E_{\mathcal{R}}$ be the graph Laplacian, where $D_{\mathcal{R}}\in \mathbb{R}^{p\times p}$ is the degree matrix of $G_{\mathcal{R}}$. Then, the generalized eigen-problem on the graph $G_{\mathcal{R}}$ can be represented as
\begin{align}
\label{eq:general_eigen_problem_reduced}
L_{\mathcal{R}}v=\lambda D_{\mathcal{R}}v.
\end{align}
It has been proved by Li et al. \cite{CVPR12_Li} that solving the eigen-problem~(\ref{eq:general_eigen_problem}) on the graph $G$ is equivalent to solving the eigen-problem~(\ref{eq:general_eigen_problem_reduced}) on the graph $G_{\mathcal{R}}$. Let the first $k$ eigen-pairs for the eigen-problem~(\ref{eq:general_eigen_problem_reduced}) be denoted as $\{(\lambda_i,v_i)\}_{i=1}^k$ with $0=\lambda_1\leq\lambda_2\leq\cdots\leq\lambda_k<1$, and the first $k$ eigen-pairs for the eigen-problem~(\ref{eq:general_eigen_problem}) denoted as $\{(\gamma_i,u_i)\}_{i=1}^k$ with $0=\gamma_1\leq\gamma_2\leq\cdots\leq\gamma_k<1$. It has been shown that \cite{CVPR12_Li}
\begin{align}
&\gamma_i(2-\gamma_i)=\lambda_i,\label{eq:uv_1}\\
&u_i=\left[\begin{matrix}
   h_i\\
   v_i
  \end{matrix}\right]\label{eq:uv_2}\\
  &h_i=\frac{1}{1-\gamma_i}Tv_i,\label{eq:uv_3}
\end{align}
where $T=D_{\mathcal{X}}^{-1}B$ is the transition probability matrix. It takes $O(p^3)$ time to compute the first $k$ eigen-pairs for the eigen-problem~(\ref{eq:general_eigen_problem_reduced}). As $B$ is a sparse matrix with $NK$ non-zero entries, it takes $O(NK)$ time to compute $u_i$ from $v_i$ according to Eqs.~(\ref{eq:uv_1}), (\ref{eq:uv_2}), and (\ref{eq:uv_3}). Therefore, the total cost of computing the first $k$ eigenvectors for the eigen-problem~(\ref{eq:general_eigen_problem}) will be $O(NK^2)+O(NKk)+O(p^3)=O(NK(K+k)+p^3)$.

With the eigen-problem solved, the obtained $k$ eigenvectors are stacked to form an $(N+p)\times k$ matrix. By treating each row of this matrix as a new feature vector, the $N$ rows corresponding to the $N$ original objects are used, upon which the $k$-means discretization can be performed to obtain the final clustering result with $O(Nk^2t)$ time complexity.

\subsubsection{Computational Complexity}
\label{sec:USPEC_complexity}

In this section, we summarize the time and memory cost of our U-SPEC algorithm.

The hybrid representative selection takes $O(p^2dt)$ time. The affinity construction takes $O(N(p^{\frac{1}{2}}d+Kd+K^2))$ time. The eigen-decomposition takes $O(NK(K+k)+p^3)$ time. The $k$-means discretization takes $O(Nk^2t)$ time. With consideration to $k,K\ll p\ll N$, the overall time complexity of U-SPEC is $O(N(p^{\frac{1}{2}}d+K^2+Kk+Kd+k^2t))$, where $O(Np^{\frac{1}{2}}d)$ is the dominant term. Table~\ref{table:cmp_complexity} provides a comparison of time complexity of our U-SPEC algorithm against several other large-scale spectral clustering algorithms.

Besides the time cost, the memory cost of U-SPEC can be either $O(NK)$ or $O(Np^\frac{1}{2})$, which depends on the actual implementation of the $K$-nearest representative approximation. As the $K$-nearest representative approximation for the $N$ objects are independent of each other, one strategy is to perform approximation for the $N$ objects one after the other (i.e., in a serial processing manner), where the time cost is dominated by the storage of the cross-affinity matrix with $NK$ non-zero entries. Another strategy is to first construct an affinity matrix between the $N$ objects and the $z_1=\lfloor{p}^{\frac{1}{2}}\rfloor$ rep-cluster centers and then approximate the $K$-nearest representatives for the $N$ objects in a batch processing manner. For some matrix-oriented software, such as MATLAB, it will be much faster to perform the approximation in a batch processing manner (with optimized matrix computation) than in a serial processing manner.
To facilitate the matrix computation, our implementation of U-SPEC actually takes $O(Np^\frac{1}{2})$ memory. Similarly, the LSC algorithm \cite{cai15_LSC} also has a theoretically minimum memory cost of $O(NK)$, but the implementation\footnote{www.cad.zju.edu.cn/home/dengcai/Data/Clustering.html} provided by the authors actually takes $O(Np)$ memory, which is also due to the matrix-computation consideration.

\begin{table}
\centering
\caption{Comparison of the time complexity of several large-scale spectral clustering methods.}
\label{table:cmp_complexity}
\begin{threeparttable}
\begin{tabular}{p{1.33cm}<{\centering}p{1.65cm}<{\centering}p{1.6cm}<{\centering}p{2.435cm}<{\centering}}
\toprule
Method        &Representative selection     &Affinity construction &Eigen-decomposition\\
\midrule
\scriptsize
Nystr\"{o}m \cite{chen11_nystrom} &/    &$O(Npd)$     &$O(Np+p^3)$\\
LSC-R \cite{cai15_LSC}           &/    &$O(Npd)$     &$O(Np^2+p^3)$\\
LSC-K \cite{cai15_LSC}           &$O(Npdt)$    &$O(Npd)$     &$O(Np^2+p^3)$\\
U-SPEC                         &$O(p^2dt)$    &$O(Np^{\frac{1}{2}}d)$     &$O(NK(K+k)+p^3)$\\
\bottomrule
\end{tabular}
\begin{tablenotes}
\item[*] The final $k$-means discretization is $O(Nk^2t)$ for each method.
\end{tablenotes}
\end{threeparttable}
\end{table}

\subsection{Ultra-Scalable Ensemble Clustering (U-SENC)}
\label{sec:USENC}

Starting from U-SPEC, this section proposes the U-SENC algorithm to integrate multiple U-SPEC's into a unified ensemble clustering framework, aiming to further enhance the clustering robustness while maintaining high efficiency.

\subsubsection{Ensemble Generation via Multiple U-SPEC's}
\label{sec:ensemble_generation}

Ensemble clustering has been a popular research topic in recent years, due to its promising ability in enhancing clustering robustness by incorporating multiple base clusterers \cite{wu15_TKDE,Huang16_TKDE,liu17_bioinformatics,liu17_tkde,huang17_tcyb}. The general ensemble clustering process consists of two phases. The first phase is the ensemble generation, which involves producing a set of diverse and high-quality base clusterings. The second phase is the consensus function, which involves combining multiple base clusterings into a better and more robust consensus clustering.

In ensemble generation, the previous ensemble clustering algorithms mostly use the $k$-means method to generate an ensemble of multiple base clusterings \cite{wu15_TKDE,Huang16_TKDE,liu17_bioinformatics,liu17_tkde,huang17_tcyb}. Though $k$-means has the advantage of high efficiency, it typically favors spherical distribution and lacks the ability to properly partition nonlinearly separable datasets. Some researchers have exploited the spectral clustering technique in ensemble generation \cite{Yu16_tkde_incremental,Yu17_tkde}, but the large computational cost of conventional spectral clustering significantly restricts its feasibility for scalable applications.

To address this, we utilize multiple instances of U-SPEC as the multiple base clusterers in our ensemble clustering framework. To generate an ensemble of $m$ base clusterings, a set of $m$ U-SPEC clusterers are required, which are denoted as U-SPEC$_1,$U-SPEC$_2,\cdots,$U-SPEC$_m$. The diversity which is highly desired in ensemble generation is incorporated from two aspects. First, the set of representatives for each base clusterer is independently obtained by the hybrid selection strategy. There are two components in hybrid selection, i.e., random pre-selection and $k$-means based post-selection, both of which are non-deterministic and can bring in diversity for the multiple base clusterers. Second, the number of clusters for each base clustering is randomly selected to further enhance the diversity. Formally, given the dataset $\mathcal{X}$, the set of $p'$ candidate representatives for the $i$-th base clusterer (i.e., U-SPEC$_i$) are randomly selected from $\mathcal{X}$. Then the $k$-means is used to  partition the $p'$ candidates into $p$ clusters. After that, the $p$ cluster centers will be used as the set of $p$ representatives for U-SPEC$_i$, denoted as
\begin{align}
\mathcal{R}^i=\{r_1^i, r_2^i, \cdots, r_p^i\}.
\end{align}
With the representatives obtained, the sparse affinity sub-matrix $B^i$ for U-SPEC$_i$ can be built between the dataset $\mathcal{X}$ and the representative set $\mathcal{R}^i$ via fast approximation of $K$-nearest representatives.

By treating $\mathcal{X}\bigcup\mathcal{R}^i$ as the node set and $B^i$ as the cross-affinity matrix, the bipartite graph $G^i$ is built and its first $k^i$ eigenvectors are then computed via transfer cut \cite{CVPR12_Li}. Note that the number of clusters $k^i$ is randomly selected as
\begin{align}
k^i=\lfloor\tau(k_{max}-k_{min})\rfloor+k_{min},
\end{align}
where $\tau\in[0,1]$ is a random variable and $k_{max}$ and $k_{min}$ are respectively the upper bound and lower bound of the cluster number. Then, the obtained $k^i$ eigenvectors are stacked to form a new matrix, upon which the $k$-means is applied to construct the base clustering result for U-SPEC$_i$. With the $m$ U-SPEC clusterers, the ensemble of $m$ base clusterings can be generated, which are represented as
\begin{align}
\Pi=\{\pi^1,\pi^2,\cdots,\pi^m\},
\end{align}
where $\pi^i$ denotes the $i$-th base clustering.

\subsubsection{Consensus Function with Bipartite Graph}
\label{sec:consensus_function}

Having obtained the set of multiple base clusterings, this section presents the consensus function with bipartite graph for obtaining the consensus clustering.

Each base clustering consists of a certain number of clusters. For clarity, we denote the set of clusters in the ensemble of $m$ base clusterings as
\begin{align}
\mathcal{C}=\{C_1,C_2,\cdots,C_{k_c}\},
\end{align}
where $C_i$ is the $i$-th cluster and $k_c$ is the total number of clusters in $\Pi$. It is obvious that $k_c=\sum_{i=1}^m k^i$.

By treating both objects and clusters as graph nodes, the bipartite graph for the ensemble $\Pi$ is defined as
\begin{align}
\tilde{G}=\{\mathcal{X},\mathcal{C},\tilde{B}\},
\end{align}
where $\mathcal{X}\bigcup\mathcal{C}$ is the node set and $\tilde{B}$ is the cross-affinity matrix. In this bipartite graph, a (non-zero) edge exists between two nodes if and only if one node is an object and the other one is the cluster that contains it. Formally, the cross-affinity matrix is constructed as follows:
\begin{align}
\tilde{B} &= \{\tilde{b}_{ij}\}_{N\times k_c},\\
\tilde{b}_{ij} &= \begin{cases}1,&\text{if $x_i\in C_j$},\\
0,&\text{otherwise.}
\end{cases}
\end{align}

Inside the same base clustering, there is no intersection between two different clusters, i.e., $\forall i'\neq j'$, if $C_{i'}\in\pi^i$ and $C_{j'}\in\pi^i$, then $C_{i'}\bigcap C_{j'}=\emptyset$. Obviously, each object belongs to one and only one cluster in each base clustering, and thus each object belongs exactly to $m$ clusters in the ensemble of $m$ base clusterings. Therefore, there are exactly $m$ non-zero entries in each row of $\tilde{B}$. Although the cross-affinity matrix $\tilde{B}$ is an $N\times k_c$ matrix, it can be stored as a sparse matrix with $O(Nm)$ memory, which corresponds to the exactly $Nm$ non-zero entries in $\tilde{B}$. Besides the memory cost, the time cost of building the sparse matrix $\tilde{B}$ is $O(Nm)$.

As shown in Section~\ref{sec:bipartite_partition}, solving the eigen-problem for the bipartite graph $\tilde{G}$ can be equivalent to solving the eigen-problem for a much smaller graph $G_{\mathcal{C}}=\{\mathcal{C},E_{\mathcal{C}}\}$, that is
\begin{align}
\label{eq:ensemble_eigen_problem_reduced}
{L}_{\mathcal{C}}\tilde{v}=\tilde{\lambda} {D}_{\mathcal{C}}\tilde{v},
\end{align}
where $E_{\mathcal{C}}=\tilde{B}^\top {\tilde{D}_{\mathcal{X}}}^{-1}\tilde{B}$ is the affinity matrix, $\tilde{D}_{\mathcal{X}}\in \mathbb{R}^{N\times N}$ is a diagonal matrix with its $(i,i)$-th entry being the sum of the $i$-th row of $\tilde{B}$, ${L}_{\mathcal{C}}={D}_{\mathcal{C}}-E_{\mathcal{C}}$ is the graph Laplacian, and ${D}_{\mathcal{C}}\in \mathbb{R}^{k_c\times k_c}$ is the degree matrix of $G_{\mathcal{C}}$.

Let $\tilde{v}_1,\tilde{v}_2,\cdots,\tilde{v}_k$ denote the first $k$ eigenvectors for the eigen-problem~(\ref{eq:ensemble_eigen_problem_reduced}), which can be computed with a time cost of $O({k_c}^3)$. Based on the $k$ eigenvectors for $G_{\mathcal{C}}$, the first $k$ eigenvectors (denoted as $\tilde{u}_1,\tilde{u}_2,\cdots,\tilde{u}_k$) for the bipartite graph $\tilde{G}$ can be computed with $O(Nm(m+k))$ time (see Eqs.~(\ref{eq:uv_1}), (\ref{eq:uv_2}), and (\ref{eq:uv_3})). Finally, by stacking the $k$ eigenvectors to form a new matrix, the consensus clustering result in U-SENC can be obtained by $k$-means discretization with $O(Nk^2t)$ time.

\subsubsection{Computational Complexity}
\label{USENC_complexity}

This section summarizes the time and memory cost of the proposed U-SENC algorithm.

The ensemble generation of the U-SENC algorithm takes $O(Nm(p^{\frac{1}{2}}d+K^2+Kk+Kd+k^2t))$ time. The consensus function of U-SENC takes $O(N(m^2+mk+k^2t)+{k_c}^3)$ time. With consideration to $m,k,K\ll p\ll N$, the dominant term of the overall time complexity of U-SENC is $O(Nmp^{\frac{1}{2}}d)$.

Meanwhile, the memory costs of the ensemble generation and the consensus function of our U-SENC algorithm are respectively $O(Np^{\frac{1}{2}})$ and $O(Nm)$.

\section{Experiments}
\label{sec:experiment}

In this section, we conduct experiments on a variety of real and synthetic datasets to compare the proposed U-SPEC and U-SENC algorithms against several state-of-the-art spectral clustering and ensemble clustering algorithms.

All experiments are conducted in Matlab 2016b on a PC with an Intel i5-6600 CPU and 64GB of RAM.

\subsection{Datasets and Evaluation Measures}

\begin{table}[!t]
\centering
\caption{Description of the real and synthetic datasets.}
\label{table:datasets}
\begin{center}
\begin{tabular}{p{1cm}<{\centering}|p{1.5cm}<{\centering}|p{1.5cm}<{\centering}p{1.2cm}<{\centering}p{1.2cm}<{\centering}}
\toprule
\multicolumn{2}{c|}{Dataset}         &\#Object     &Dimension      &\#Class\\
\midrule
\multirow{5}{*}{\emph{Real}}  &\emph{PenDigits}    &10,992   &16   &10\\
&\emph{USPS}    &11,000   &256   &10\\
&\emph{Letters}    &20,000   &16   &26\\
&\emph{MNIST}    &70,000   &784   &10\\
&\emph{Covertype}    &581,012   &54   &7\\
\midrule
\multirow{5}{*}{\emph{Synthetic}}  &\emph{TB-1M}    &1,000,000   &2   &2\\
&\emph{SF-2M}    &2,000,000   &2   &4\\
&\emph{CC-5M}    &5,000,000   &2   &3\\
&\emph{CG-10M}   &10,000,000   &2   &11\\
&\emph{Flower-20M}    &20,000,000   &2   &13\\
\bottomrule
\end{tabular}
\end{center}
\end{table}

\begin{figure}
\begin{center}
{\subfigure[\emph{TB-1M} ($0.1\%$)]
{\includegraphics[width=0.31\columnwidth]{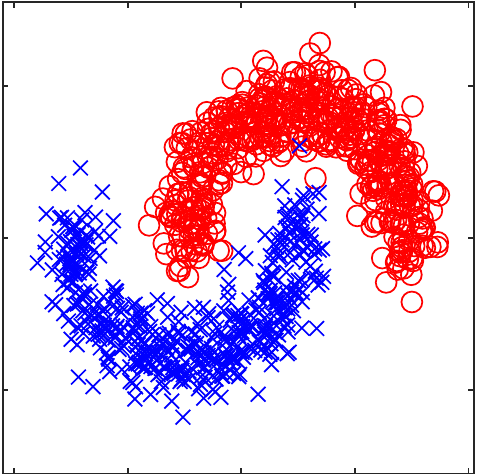}}}
{\subfigure[\emph{SF-2M} ($0.1\%$)]
{\includegraphics[width=0.31\columnwidth]{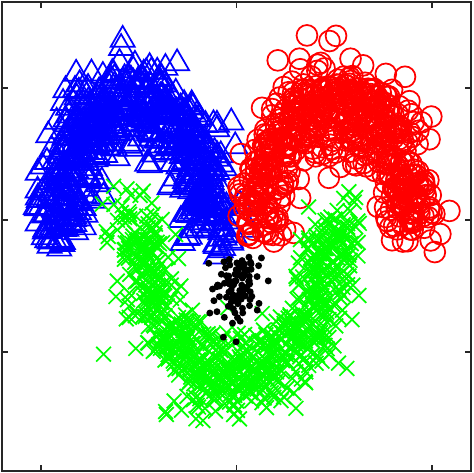}}}
{\subfigure[\emph{CC-5M} ($0.1\%$)]
{\includegraphics[width=0.31\columnwidth]{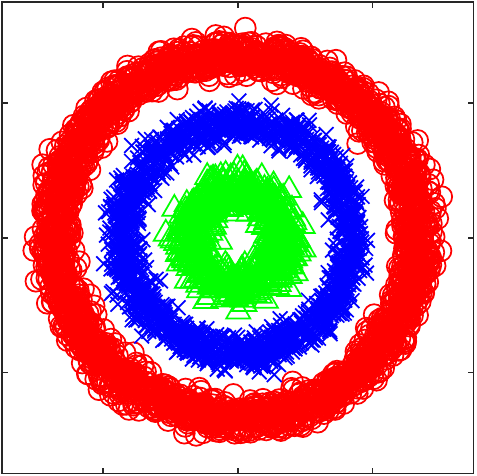}}}
{\subfigure[\emph{CG-10M} ($0.1\%$)]
{\includegraphics[width=0.31\columnwidth]{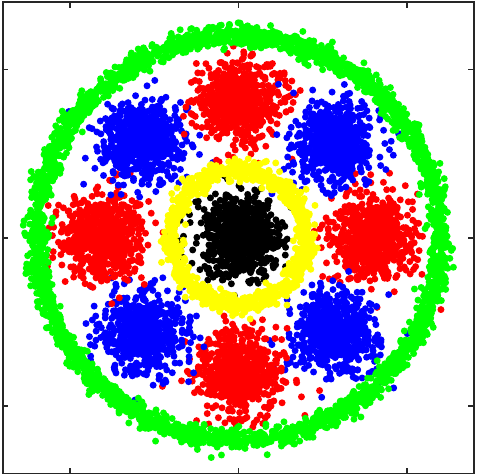}}}
{\subfigure[\emph{Flower-20M} ($0.1\%$)]
{\includegraphics[width=0.31\columnwidth]{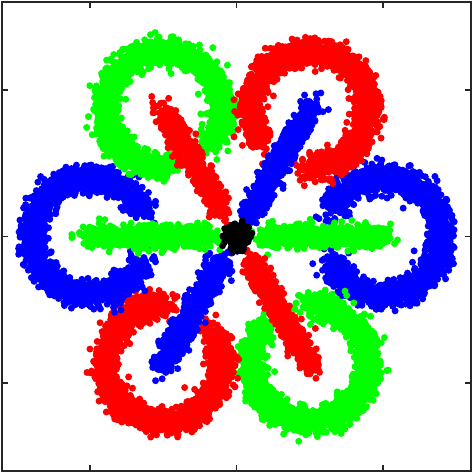}}}
\caption{Illustration of the five synthetic datasets. Note that only a $0.1\%$ subset of each dataset is plotted.}
\label{fig:fiveSynDS}
\end{center}
\end{figure}

Our experiments are conducted on ten large-scale datasets (including five real datasets and five synthetic datasets), whose data sizes range from ten thousand to as large as twenty million. Specifically, the five real datasets are \emph{PenDigits} \cite{Bache+Lichman:2013}, \emph{USPS} \cite{SamRoweisHomepage}, \emph{Letters} \cite{Bache+Lichman:2013}, \emph{MNIST} \cite{SamRoweisHomepage}, and \emph{Covertype} \cite{Bache+Lichman:2013}. The five synthetic datasets are \emph{Two Bananas-1M} (\emph{TB-1M}), \emph{Smiling Face-2M} (\emph{SF-2M}), \emph{Concentric Circles-5M} (\emph{CC-5M}), \emph{Circles and Gaussians-10M} (\emph{CG-10M}), and \emph{Flower-20M}. The details of the datasets are provided in Table~\ref{table:datasets} and Fig.~\ref{fig:fiveSynDS}.

\begin{table*}
\centering
\caption{Average NMI(\%) scores (over 20 runs) by our methods and the baseline spectral clustering methods
(The best score in each row is in bold).}
\label{table:compare_spectrals_nmi}
\begin{threeparttable}
\begin{tabular}{|m{1.7cm}<{\centering}||m{1.3cm}<{\centering}||m{1.12cm}<{\centering}m{1.12cm}<{\centering}m{1.12cm}<{\centering}m{1.12cm}<{\centering}m{1.12cm}<{\centering}m{1.12cm}<{\centering}m{1.25cm}<{\centering}|m{1.12cm}<{\centering}m{1.25cm}<{\centering}|}
\hline
\emph{Dataset} &$k$-means &SC &ESCG &Nystr\"{o}m &LSC-K &LSC-R &FastESC   &EulerSC &U-SPEC &U-SENC\\
\hline
\hline
\emph{PenDigits}	&66.66$_{\pm1.76}$	&59.36$_{\pm0.00}$	&76.41$_{\pm2.26}$	&65.67$_{\pm1.16}$	&79.73$_{\pm2.09}$	&78.13$_{\pm2.20}$	&65.31$_{\pm0.71}$	&58.59$_{\pm0.73}$	&80.30$_{\pm2.18}$	&\textbf{85.34}$_{\pm0.91}$\\
\hline
\emph{USPS}	&44.11$_{\pm1.24}$	&63.44$_{\pm0.01}$	&48.41$_{\pm3.53}$	&44.91$_{\pm1.28}$	&66.86$_{\pm1.58}$	&58.64$_{\pm1.31}$	&41.36$_{\pm1.80}$	&40.31$_{\pm1.91}$	&63.47$_{\pm0.97}$	&\textbf{73.89}$_{\pm1.82}$\\
\hline
\emph{Letters}	&34.86$_{\pm0.60}$	&10.43$_{\pm0.50}$	&35.80$_{\pm1.72}$	&39.02$_{\pm0.83}$	&43.41$_{\pm0.81}$	&40.98$_{\pm0.93}$	&35.92$_{\pm1.41}$	&31.76$_{\pm0.92}$	&42.53$_{\pm1.32}$	&\textbf{45.90}$_{\pm0.58}$\\
\hline
\emph{MNIST}	&48.91$_{\pm2.00}$	&74.07$_{\pm0.00}$	&55.75$_{\pm4.62}$	&47.78$_{\pm1.17}$	&73.97$_{\pm1.46}$	&62.16$_{\pm2.22}$	&43.44$_{\pm1.85}$	&8.93$_{\pm1.22}$	&67.43$_{\pm1.55}$	&\textbf{75.02}$_{\pm0.81}$\\
\hline
\emph{Covertype}	&6.17$_{\pm0.00}$	&N/A	&N/A	&6.93$_{\pm0.07}$	&6.75$_{\pm0.10}$	&6.69$_{\pm0.12}$	&\textbf{9.15}$_{\pm1.00}$	&0.01$_{\pm0.00}$	&6.97$_{\pm0.16}$	&9.13$_{\pm1.21}$\\
\hline
\emph{TB-1M}	&25.71$_{\pm0.00}$	&N/A	&N/A	&24.06$_{\pm0.01}$	&0.10$_{\pm0.11}$	&0.20$_{\pm0.24}$	&24.01$_{\pm2.72}$	&25.94$_{\pm0.01}$	&95.86$_{\pm0.48}$	&\textbf{97.48}$_{\pm0.05}$\\
\hline
\emph{SF-2M}	&47.34$_{\pm0.23}$	&N/A	&N/A	&46.66$_{\pm0.02}$	&66.45$_{\pm6.15}$	&58.34$_{\pm6.92}$	&52.03$_{\pm0.95}$	&47.35$_{\pm2.19}$	&75.59$_{\pm2.12}$	&\textbf{77.02}$_{\pm2.32}$\\
\hline
\emph{CC-5M}	&0.00$_{\pm0.00}$	&N/A	&N/A	&N/A	&N/A	&N/A	&N/A	&0.00$_{\pm0.00}$	&99.87$_{\pm0.01}$	&\textbf{99.91}$_{\pm0.00}$\\
\hline
\emph{CG-10M}	&63.20$_{\pm1.59}$	&N/A	&N/A	&N/A	&N/A	&N/A	&N/A	&16.19$_{\pm0.21}$	&78.82$_{\pm1.61}$	&\textbf{89.57}$_{\pm3.96}$\\
\hline
\emph{Flower-20M}	&64.19$_{\pm2.56}$	&N/A	&N/A	&N/A	&N/A	&N/A	&N/A	&26.61$_{\pm0.86}$	&86.86$_{\pm2.05}$	&\textbf{92.47}$_{\pm2.45}$\\
\hline
\hline
Avg. score	&-	&N/A	&N/A	&N/A	&N/A	&N/A	&N/A	&25.57	&69.77	&\textbf{74.57}\\
\hline
N-Avg. score	&-	&N/A	&N/A	&N/A	&N/A	&N/A	&N/A	&33.94	&91.71	&\textbf{99.98}\\
\hline
\hline
Avg. rank	&-	&5.90	&6.00	&5.20	&3.70	&4.60	&5.20	&6.00	&2.50	&\textbf{1.10}\\
\hline
\end{tabular}
\begin{tablenotes}
\item[*] Note that N/A indicates the out-of-memory error.
\item[**] The $k$-means method is listed for reference only; it doesn't participate in the comparison of the spectral methods.
\end{tablenotes}
\end{threeparttable}
\end{table*}

\begin{table*}
\centering
\caption{Average CA(\%) scores (over 20 runs) by our methods and the baseline spectral clustering methods
(The best score in each row is in bold).}
\label{table:compare_spectrals_ca}
\begin{threeparttable}
\begin{tabular}{|m{1.7cm}<{\centering}||m{1.3cm}<{\centering}||m{1.12cm}<{\centering}m{1.12cm}<{\centering}m{1.12cm}<{\centering}m{1.12cm}<{\centering}m{1.12cm}<{\centering}m{1.12cm}<{\centering}m{1.25cm}<{\centering}|m{1.12cm}<{\centering}m{1.25cm}<{\centering}|}
\hline
\emph{Dataset} &$k$-means &SC &ESCG &Nystr\"{o}m &LSC-K &LSC-R &FastESC   &EulerSC &U-SPEC &U-SENC\\
\hline
\hline
\emph{PenDigits}	&71.57$_{\pm3.12}$	&56.44$_{\pm0.00}$	&77.21$_{\pm3.81}$	&71.13$_{\pm2.07}$	&83.07$_{\pm3.21}$	&81.82$_{\pm3.17}$	&69.97$_{\pm1.15}$	&65.85$_{\pm1.87}$	&84.17$_{\pm3.26}$	&\textbf{88.56}$_{\pm0.61}$\\
\hline
\emph{USPS}	&47.25$_{\pm2.57}$	&62.74$_{\pm0.02}$	&53.47$_{\pm3.94}$	&51.09$_{\pm1.93}$	&68.42$_{\pm2.39}$	&60.78$_{\pm2.18}$	&48.80$_{\pm1.76}$	&47.79$_{\pm2.41}$	&63.76$_{\pm1.35}$	&\textbf{78.17}$_{\pm3.05}$\\
\hline
\emph{Letters}	&28.15$_{\pm0.97}$	&12.42$_{\pm0.46}$	&30.37$_{\pm1.75}$	&32.05$_{\pm0.91}$	&35.45$_{\pm1.34}$	&33.86$_{\pm1.13}$	&29.32$_{\pm1.51}$	&28.08$_{\pm1.44}$	&35.71$_{\pm1.47}$	&\textbf{37.74}$_{\pm1.06}$\\
\hline
\emph{MNIST}	&58.48$_{\pm2.67}$	&74.46$_{\pm0.00}$	&63.32$_{\pm4.64}$	&59.72$_{\pm1.75}$	&79.45$_{\pm1.02}$	&69.24$_{\pm2.75}$	&55.93$_{\pm2.41}$	&24.06$_{\pm1.53}$	&74.31$_{\pm2.28}$	&\textbf{80.58}$_{\pm1.75}$\\
\hline
\emph{Covertype}	&49.05$_{\pm0.00}$	&N/A	&N/A	&49.21$_{\pm0.11}$	&49.45$_{\pm0.16}$	&49.32$_{\pm0.25}$	&48.88$_{\pm0.18}$	&48.76$_{\pm0.00}$	&49.76$_{\pm0.35}$	&\textbf{50.73}$_{\pm0.62}$\\
\hline
\emph{TB-1M}	&78.93$_{\pm0.00}$	&N/A	&N/A	&78.04$_{\pm0.01}$	&51.54$_{\pm1.13}$	&52.09$_{\pm1.58}$	&77.97$_{\pm1.52}$	&79.04$_{\pm0.00}$	&99.55$_{\pm0.06}$	&\textbf{99.75}$_{\pm0.01}$\\
\hline
\emph{SF-2M}	&74.33$_{\pm2.14}$	&N/A	&N/A	&69.58$_{\pm0.05}$	&85.34$_{\pm5.70}$	&78.26$_{\pm7.43}$	&74.13$_{\pm0.32}$	&76.93$_{\pm2.17}$	&\textbf{93.60}$_{\pm1.00}$	&93.46$_{\pm2.27}$\\
\hline
\emph{CC-5M}	&52.96$_{\pm0.00}$	&N/A	&N/A	&N/A	&N/A	&N/A	&N/A	&52.96$_{\pm0.00}$	&\textbf{99.99}$_{\pm0.00}$	&\textbf{99.99}$_{\pm0.00}$\\
\hline
\emph{CG-10M}	&63.14$_{\pm2.42}$	&N/A	&N/A	&N/A	&N/A	&N/A	&N/A	&32.81$_{\pm0.67}$	&81.32$_{\pm2.00}$	&\textbf{93.99}$_{\pm3.25}$\\
\hline
\emph{Flower-20M}	&60.85$_{\pm3.33}$	&N/A	&N/A	&N/A	&N/A	&N/A	&N/A	&33.75$_{\pm0.56}$	&88.89$_{\pm2.85}$	&\textbf{93.79}$_{\pm3.21}$\\
\hline
\hline
Avg. score	&-	&N/A	&N/A	&N/A	&N/A	&N/A	&N/A	&49.00	&77.11	&\textbf{81.68}\\
\hline
N-Avg. score	&-	&N/A	&N/A	&N/A	&N/A	&N/A	&N/A	&62.12	&94.26	&\textbf{99.99}\\
\hline
\hline
Avg. rank	&-	&6.10	&5.90	&5.30	&3.50	&4.40	&5.90	&5.80	&2.10	&\textbf{1.10}\\
\hline
\end{tabular}
\end{threeparttable}
\end{table*}

\begin{table*}
\centering
\caption{Time costs(s) of our methods and the baseline spectral clustering methods.}
\label{table:compare_spectrals_time}
\begin{threeparttable}
\begin{tabular}{|m{1.7cm}<{\centering}||m{1.3cm}<{\centering}||m{1.12cm}<{\centering}m{1.12cm}<{\centering}m{1.12cm}<{\centering}m{1.12cm}<{\centering}m{1.12cm}<{\centering}m{1.12cm}<{\centering}m{1.25cm}<{\centering}|m{1.12cm}<{\centering}m{1.25cm}<{\centering}|}
\hline
\emph{Dataset} &$k$-means &SC &ESCG &Nystr\"{o}m &LSC-K &LSC-R &FastESC   &EulerSC &U-SPEC &U-SENC\\
\hline
\emph{PenDigits}		&0.06	&7.37	&1.63	&1.98	&1.25	&\textbf{0.49}	&0.73	&1.47	&1.01	&19.13\\
\hline
\emph{USPS}		&0.32	&9.56	&9.63	&1.92	&1.70	&\textbf{0.75}	&0.94	&8.20	&1.59	&29.17\\
\hline
\emph{Letters}		&0.72	&3.85	&7.74	&2.69	&3.89	&2.88	&1.86	&23.39	&\textbf{1.44}	&21.44\\
\hline
\emph{MNIST}		&8.79	&1,231.68	&1,211.54	&6.40	&16.51	&6.38	&\textbf{3.82}	&125.35	&7.48	&131.60\\
\hline
\emph{Covertype}		&13.19	&N/A	&N/A	&33.11	&101.12	&53.46	&19.55	&116.96	&\textbf{14.08}	&174.49\\
\hline
\emph{TB-1M}		&3.25	&N/A	&N/A	&105.15	&109.23	&35.92	&21.79	&\textbf{6.27}	&10.47	&318.29\\
\hline
\emph{SF-2M}		&31.26	&N/A	&N/A	&226.77	&254.98	&102.55	&51.07	&80.44	&\textbf{27.06}	&658.82\\
\hline
\emph{CC-5M}		&94.76	&N/A	&N/A	&N/A	&N/A	&N/A	&N/A	&132.35	&\textbf{46.65}	&1,726.40\\
\hline
\emph{CG-10M}		&281.84	&N/A	&N/A	&N/A	&N/A	&N/A	&N/A	&963.29	&\textbf{318.93}	&3,603.08\\
\hline
\emph{Flower-20M}		&579.06	&N/A	&N/A	&N/A	&N/A	&N/A	&N/A	&3,397.57	&\textbf{764.09}	&7,225.83\\
\hline
\end{tabular}
\end{threeparttable}
\end{table*}

To evaluate the clustering results by different algorithms, two widely used evaluation measures are adopted, namely, normalized mutual information (NMI) \cite{strehl02} and clustering accuracy (CA) \cite{Nguyen07_icdm}. To rule out the factor of \emph{getting lucky occasionally}, in each experiment, every test method will be conducted 20 times and their average NMI, CA, and time costs will be reported. Note that larger values of NMI and CA indicate better clustering results.

\subsection{Baseline Methods and Experimental Settings}

\begin{table*}
\centering
\caption{Average NMI(\%) scores (over 20 runs) by our methods and the baseline ensemble clustering methods
(The best score in each row is in bold).}
\label{table:compare_ens_nmi}
\begin{threeparttable}
\begin{tabular}{|m{1.7cm}<{\centering}||m{1.38cm}<{\centering}||m{1.18cm}<{\centering}m{1.18cm}<{\centering}m{1.18cm}<{\centering}m{1.18cm}<{\centering}m{1.18cm}<{\centering}m{1.18cm}<{\centering}m{1.31cm}<{\centering}|m{1.31cm}<{\centering}|}
\hline
\emph{Dataset} &U-SPEC &EAC &WCT &KCC &PTGP &ECC &SEC &LWGP  &U-SENC\\
\hline
\hline
\emph{PenDigits}	&80.30$_{\pm2.18}$	&76.31$_{\pm2.70}$	&77.69$_{\pm2.54}$	&58.92$_{\pm3.47}$	&75.58$_{\pm2.26}$	&57.64$_{\pm4.14}$	&47.07$_{\pm7.53}$	&77.54$_{\pm1.87}$	&\textbf{85.34}$_{\pm0.91}$\\
\hline
\emph{USPS}	&63.47$_{\pm0.97}$	&59.02$_{\pm1.69}$	&58.40$_{\pm2.15}$	&49.24$_{\pm2.98}$	&59.63$_{\pm1.76}$	&48.89$_{\pm1.80}$	&39.00$_{\pm3.83}$	&57.55$_{\pm1.78}$	&\textbf{73.89}$_{\pm1.82}$\\
\hline
\emph{Letters}	&42.53$_{\pm1.32}$	&37.19$_{\pm0.50}$	&36.59$_{\pm0.95}$	&33.64$_{\pm1.03}$	&38.09$_{\pm0.66}$	&34.59$_{\pm0.68}$	&31.81$_{\pm2.01}$	&37.09$_{\pm0.75}$	&\textbf{45.90}$_{\pm0.58}$\\
\hline
\emph{MNIST}	&67.43$_{\pm1.55}$	&66.19$_{\pm1.49}$	&65.60$_{\pm0.96}$	&54.34$_{\pm3.38}$	&59.93$_{\pm2.23}$	&56.01$_{\pm2.25}$	&34.19$_{\pm4.61}$	&65.06$_{\pm0.95}$	&\textbf{75.02}$_{\pm0.81}$\\
\hline
\emph{Covertype}	&6.97$_{\pm0.16}$	&N/A	&N/A	&5.86$_{\pm1.84}$	&6.42$_{\pm0.44}$	&5.70$_{\pm0.77}$	&5.26$_{\pm2.82}$	&7.44$_{\pm0.31}$	&\textbf{9.13}$_{\pm1.21}$\\
\hline
\emph{TB-1M}	&95.86$_{\pm0.48}$	&N/A	&N/A	&23.36$_{\pm1.62}$	&34.20$_{\pm2.51}$	&26.91$_{\pm2.13}$	&10.62$_{\pm4.64}$	&96.80$_{\pm1.90}$	&\textbf{97.48}$_{\pm0.05}$\\
\hline
\emph{SF-2M}	&75.59$_{\pm2.12}$	&N/A	&N/A	&42.72$_{\pm7.11}$	&45.17$_{\pm2.66}$	&41.61$_{\pm6.01}$	&27.05$_{\pm7.73}$	&69.88$_{\pm4.45}$	&\textbf{77.02}$_{\pm2.32}$\\
\hline
\emph{CC-5M}	&99.87$_{\pm0.01}$	&N/A	&N/A	&33.36$_{\pm12.65}$	&0.41$_{\pm0.86}$	&31.62$_{\pm14.99}$	&17.05$_{\pm6.90}$	&98.18$_{\pm7.75}$	&\textbf{99.91}$_{\pm0.00}$\\
\hline
\emph{CG-10M}	&78.82$_{\pm1.61}$	&N/A	&N/A	&64.78$_{\pm5.08}$	&63.75$_{\pm0.61}$	&62.79$_{\pm4.91}$	&49.70$_{\pm6.08}$	&78.08$_{\pm2.43}$	&\textbf{89.57}$_{\pm3.96}$\\
\hline
\emph{Flower-20M}	&86.86$_{\pm2.05}$	&N/A	&N/A	&61.18$_{\pm2.43}$	&67.92$_{\pm1.99}$	&60.61$_{\pm2.37}$	&50.37$_{\pm6.32}$	&78.55$_{\pm2.31}$	&\textbf{92.47}$_{\pm2.45}$\\
\hline
\hline
Avg. score	&-	&N/A	&N/A	&42.74	&45.11	&42.64	&31.21	&66.62	&\textbf{74.57}\\
\hline
N-Avg. score	&-	&N/A	&N/A	&59.69	&64.12	&59.51	&45.35	&87.82	&\textbf{100.00}\\
\hline
\hline
Avg. rank	&-	&5.40	&5.60	&4.90	&3.60	&5.40	&6.70	&2.80	&\textbf{1.00}\\
\hline
\end{tabular}
\begin{tablenotes}
\item[*] The U-SPEC is listed for reference only; it doesn't participate in the comparison of the ensemble methods.
\end{tablenotes}
\end{threeparttable}
\end{table*}

\begin{table*}
\centering
\caption{Average CA(\%) scores (over 20 runs) by our methods and the baseline ensemble clustering methods
(The best score in each row is in bold).}
\label{table:compare_ens_ca}
\begin{threeparttable}
\begin{tabular}{|m{1.7cm}<{\centering}||m{1.38cm}<{\centering}||m{1.18cm}<{\centering}m{1.18cm}<{\centering}m{1.18cm}<{\centering}m{1.18cm}<{\centering}m{1.18cm}<{\centering}m{1.18cm}<{\centering}m{1.31cm}<{\centering}|m{1.31cm}<{\centering}|}
\hline
\emph{Dataset} &U-SPEC &EAC &WCT &KCC &PTGP &ECC &SEC &LWGP  &U-SENC\\
\hline
\hline
\emph{PenDigits}	&84.17$_{\pm3.26}$	&81.04$_{\pm4.02}$	&82.97$_{\pm3.17}$	&63.33$_{\pm4.06}$	&78.33$_{\pm2.91}$	&62.36$_{\pm4.12}$	&51.60$_{\pm5.93}$	&81.96$_{\pm2.77}$	&\textbf{88.56}$_{\pm0.61}$\\
\hline
\emph{USPS}	&63.76$_{\pm1.35}$	&63.39$_{\pm2.76}$	&62.72$_{\pm3.14}$	&53.46$_{\pm3.51}$	&62.68$_{\pm1.92}$	&53.67$_{\pm2.21}$	&45.38$_{\pm3.20}$	&59.73$_{\pm3.30}$	&\textbf{78.17}$_{\pm3.05}$\\
\hline
\emph{Letters}	&35.71$_{\pm1.47}$	&30.28$_{\pm0.58}$	&30.17$_{\pm1.01}$	&26.90$_{\pm1.23}$	&31.50$_{\pm0.89}$	&27.53$_{\pm0.72}$	&26.12$_{\pm1.93}$	&30.76$_{\pm0.84}$	&\textbf{37.74}$_{\pm1.06}$\\
\hline
\emph{MNIST}	&74.31$_{\pm2.28}$	&73.12$_{\pm2.73}$	&70.73$_{\pm1.76}$	&59.86$_{\pm5.11}$	&65.06$_{\pm2.75}$	&61.18$_{\pm3.58}$	&43.13$_{\pm4.88}$	&71.98$_{\pm1.67}$	&\textbf{80.58}$_{\pm1.75}$\\
\hline
\emph{Covertype}	&49.76$_{\pm0.35}$	&N/A	&N/A	&49.54$_{\pm0.58}$	&49.11$_{\pm0.30}$	&49.68$_{\pm0.40}$	&49.86$_{\pm0.94}$	&49.50$_{\pm0.28}$	&\textbf{50.73}$_{\pm0.62}$\\
\hline
\emph{TB-1M}	&99.55$_{\pm0.06}$	&N/A	&N/A	&70.05$_{\pm1.21}$	&82.94$_{\pm1.08}$	&72.50$_{\pm1.48}$	&60.12$_{\pm3.64}$	&99.65$_{\pm0.31}$	&\textbf{99.75}$_{\pm0.01}$\\
\hline
\emph{SF-2M}	&93.60$_{\pm1.00}$	&N/A	&N/A	&67.12$_{\pm5.41}$	&73.46$_{\pm1.76}$	&66.90$_{\pm6.15}$	&55.91$_{\pm5.71}$	&88.71$_{\pm3.28}$	&\textbf{93.46}$_{\pm2.27}$\\
\hline
\emph{CC-5M}	&99.99$_{\pm0.00}$	&N/A	&N/A	&66.76$_{\pm6.24}$	&52.96$_{\pm0.00}$	&62.71$_{\pm5.38}$	&61.91$_{\pm5.49}$	&99.30$_{\pm3.07}$	&\textbf{99.99}$_{\pm0.00}$\\
\hline
\emph{CG-10M}	&81.32$_{\pm2.00}$	&N/A	&N/A	&66.96$_{\pm5.60}$	&63.36$_{\pm1.26}$	&64.74$_{\pm6.80}$	&58.19$_{\pm4.69}$	&81.95$_{\pm3.93}$	&\textbf{93.99}$_{\pm3.25}$\\
\hline
\emph{Flower-20M}	&88.89$_{\pm2.85}$	&N/A	&N/A	&57.78$_{\pm3.37}$	&63.83$_{\pm2.34}$	&56.69$_{\pm2.35}$	&50.70$_{\pm5.02}$	&81.37$_{\pm2.69}$	&\textbf{93.79}$_{\pm3.21}$\\
\hline
\hline
Avg. score	&-	&N/A	&N/A	&58.18	&62.32	&57.80	&50.29	&74.49	&\textbf{81.68}\\
\hline
N-Avg. score	&-	&N/A	&N/A	&72.48	&77.98	&72.22	&63.53	&90.54	&\textbf{100.00}\\
\hline
\hline
Avg. rank	&-	&5.40	&5.60	&5.00	&4.20	&5.00	&6.30	&2.90	&\textbf{1.00}\\
\hline
\end{tabular}
\end{threeparttable}
\end{table*}

\begin{table*}
\centering
\caption{Time costs(s) of our methods and the baseline ensemble clustering methods.}
\label{table:compare_ens_time}
\begin{threeparttable}
\begin{tabular}{|m{1.7cm}<{\centering}||m{1.38cm}<{\centering}||m{1.18cm}<{\centering}m{1.18cm}<{\centering}m{1.18cm}<{\centering}m{1.18cm}<{\centering}m{1.18cm}<{\centering}m{1.18cm}<{\centering}m{1.31cm}<{\centering}|m{1.31cm}<{\centering}|}
\hline
\emph{Dataset} &U-SPEC &EAC &WCT &KCC &PTGP &ECC &SEC &LWGP  &U-SENC\\
\hline
\hline
\emph{PenDigits}		&1.01	&8.89	&47.01	&8.97	&11.94	&13.56	&\textbf{5.27}	&5.46	&19.13\\
\hline
\emph{USPS}		&1.59	&13.11	&48.45	&15.87	&59.71	&23.53	&\textbf{10.15}	&10.25	&29.17\\
\hline
\emph{Letters}		&1.44	&29.60	&177.11	&33.91	&137.46	&53.04	&16.06	&\textbf{15.58}	&21.44\\
\hline
\emph{MNIST}		&7.48	&576.71	&3,435.19	&315.58	&2,205.18	&417.10	&260.96	&259.91	&\textbf{131.60}\\
\hline
\emph{Covertype}		&14.08	&N/A	&N/A	&954.89	&7,919.02	&1,482.43	&712.84	&685.89	&\textbf{174.49}\\
\hline
\emph{TB-1M}		&10.47	&N/A	&N/A	&1,308.54	&1,276.82	&2,100.02	&1,000.30	&989.10	&\textbf{318.29}\\
\hline
\emph{SF-2M}		&27.06	&N/A	&N/A	&2,908.34	&2,493.99	&4,714.16	&2,160.46	&2,105.82	&\textbf{658.82}\\
\hline
\emph{CC-5M}		&46.65	&N/A	&N/A	&6,833.38	&5,027.91	&11,202.43	&5,130.84	&5,070.21	&\textbf{1,726.40}\\
\hline
\emph{CG-10M}		&318.93	&N/A	&N/A	&17,344.29	&11,578.11	&27,492.40	&10,938.88	&10,700.38	&\textbf{3,603.08}\\
\hline
\emph{Flower-20M}		&764.09	&N/A	&N/A	&34,869.83	&21,198.87	&54,913.10	&21,696.29	&21,378.63	&\textbf{7,225.83}\\
\hline
\end{tabular}
\end{threeparttable}
\end{table*}

In the experiments, we first compare our algorithms against the classical $k$-means algorithm \cite{cai11_litekmeans} as well as seven spectral clustering algorithms (including the original algorithm and six large-scale algorithms). The baseline spectral clustering algorithms are listed as follows:

\begin{enumerate}
  \item \textbf{SC} \cite{vonLuxburg2007}: original spectral clustering.
  \item \textbf{ESCG} \cite{Liu13_escg}: efficient spectral clustering on graphs.
  \item \textbf{Nystr\"{o}m} \cite{chen11_nystrom}: Nystr\"{o}m spectral clustering.
  \item \textbf{LSC-K} \cite{cai15_LSC}: landmark based spectral clustering using $k$-means based landmark selection.
  \item \textbf{LSC-R} \cite{cai15_LSC}: landmark based spectral clustering using random landmark selection.
  \item \textbf{FastESC} \cite{he18_tcyb}: fast explicit spectral clustering.
  \item \textbf{EulerSC} \cite{wu17_Euler}: Euler spectral clustering.
\end{enumerate}

Besides these large-scale spectral clustering algorithms, we also compare our algorithms against seven ensemble clustering algorithms, which are listed as follows:

\begin{enumerate}
  \item \textbf{EAC} \cite{Fred05_EAC}: evidence accumulation clustering.
  \item \textbf{WCT} \cite{iam_on11_linkbased}: weighted connected triple method.
  \item \textbf{KCC} \cite{wu15_TKDE}: $k$-means based consensus clustering.
  \item \textbf{PTGP} \cite{Huang16_TKDE}: probability trajectory based graph partitioning.
  \item \textbf{ECC} \cite{liu17_bioinformatics}: entropy based consensus clustering.
  \item \textbf{SEC} \cite{liu17_tkde}: spectral ensemble clustering.
  \item \textbf{LWGP} \cite{huang17_tcyb}: locally weighted graph partitioning.
\end{enumerate}

There are several common parameters among the above-mentioned algorithms. In our experiments, we comply with the following experimental settings:
\begin{itemize}
  \item The SC and ESCG methods need to take the $N\times N$ affinity matrix as input. The affinity matrix is constructed using the same Gaussian kernel as Eq.~(\ref{eq:gaussian_kernel}) with $K$-nearest neighbors.
  \item The U-SPEC, U-SENC, Nystr\"{o}m, LSC-K, and LSC-R methods have a common parameter $p$. In the experiments, $p=1000$ is used for these methods. Their performances with varying $p$ will be further evaluated in Section~\ref{sec:para_p}.
  \item The U-SPEC, U-SENC, LSC-K, and LSC-R methods have a common parameter $K$. In the experiments, $K=5$ is used. Their performances with varying $K$ will be further evaluated in Section~\ref{sec:para_K}.
  \item For the seven ensemble clustering methods, the base clusterings are generated by $k$-means as suggested by their papers \cite{Fred05_EAC,iam_on11_linkbased,wu15_TKDE,Huang16_TKDE,liu17_bioinformatics,liu17_tkde,huang17_tcyb}. The number of clusters in each base clustering is randomly selected in $[20,60]$. The number of base clusterings, i.e., $m$, is set to $20$. Their performances with varying $m$ will be further evaluated in Section~\ref{sec:para_M}.
  \item The true number of classes on each dataset is used as the number of clusters for all the test methods.
  \item Besides these common parameters, the other parameters in the baseline methods will be set as suggested by the corresponding papers.
\end{itemize}

\subsection{Comparison with Spectral Clustering Methods}
\label{sec:cmp_spectral}

In this section, we compare our U-SPEC and U-SENC algorithms with several state-of-the-art large-scale spectral clustering algorithms.

As the data sizes range from ten thousand to twenty million, most of the baseline algorithms are not computationally feasible for ten-million-level datasets. Specifically, we use N/A to indicate the out-of-memory error in the results. As shown in Tables~\ref{table:compare_spectrals_nmi} and \ref{table:compare_spectrals_ca}, the SC and ESCG methods are not able to handle the datasets large than MNIST (which consists of 70,000 objects), due to the memory consumption of constructing and manipulating the $N\times N$ affinity matrix. The Nystr\"{o}m, LSC-K, LSC-R, and FastESC methods can at most partition a dataset with two million objects, and cannot deal with datasets larger than that. Out of the total of nine spectral clustering methods, only three methods (i.e., U-SPEC, U-SENC, and EulerSC) can deal with all of the benchmark datasets. As shown in Tables~\ref{table:compare_spectrals_nmi} and \ref{table:compare_spectrals_ca}, our U-SENC and U-SPEC methods achieve the best and the second best scores, respectively, on most of the ten benchmark datasets.

In Tables~\ref{table:compare_spectrals_nmi} and \ref{table:compare_spectrals_ca}, we also provide the average score, normalized average score (N-Avg. score), and average rank of each method across the ten datasets. To obtain the normalized average score, the scores in each row will first be divided by the maximum score in this row, where it is obvious that the maximum score will become $100\%$. Then we take the average of these normalized rows as the normalized average score. Note that if a baseline method cannot process all the datasets, it will not have the average score and normalized average score information, but it will still have the average rank information. For example, if only three methods are efficient enough to process the \emph{CC-5M} dataset, then all the other \emph{infeasible} methods will be treated as equally ranked in the fourth position on this dataset. As shown in Tables~\ref{table:compare_spectrals_nmi} and \ref{table:compare_spectrals_ca}, our U-SENC method ranks in the first position on nine out of the ten datasets, and achieves an average rank of 1.10 w.r.t. both NMI and CA. Our U-SPEC method achieves an average rank of 2.40 w.r.t. NMI and 2.00 w.r.t. CA. In terms of average score and normalized average score, our U-SENC and U-SPEC methods also significantly outperform the other methods.

Table~\ref{table:compare_spectrals_time} reports the time costs of different methods on the benchmark datasets. The U-SPEC shows superior efficiency on most of the datasets, especially on the datasets larger than one million. The U-SENC requires a larger time cost than U-SPEC, but it still provides better scalability than most of the baseline methods and scales well for ten-million-level datasets due to its memory efficiency. As U-SENC is a spectral clustering algorithm and also an ensemble clustering algorithm, in the following, we will further compare it with other state-of-the-art ensemble clustering algorithms.

\subsection{Comparison with Ensemble Clustering Methods}
\label{sec:cmp_ensemble}

In this section, we compare our algorithms with several state-of-the-art ensemble clustering algorithms.

Note that U-SPEC is not an ensemble clustering algorithm; its clustering results are provided in Tables~\ref{table:compare_ens_nmi}, \ref{table:compare_ens_ca}, and \ref{table:compare_ens_time} for reference only. As shown in Tables~\ref{table:compare_ens_nmi} and \ref{table:compare_ens_ca}, our U-SENC algorithm obtains the highest NMI and CA scores on all of the ten datasets. In terms of average score across the ten datasets, U-SENC achieves the best average NMI($\%$) and CA($\%$) scores of $74.57$ and $81.68$, respectively while the second best ensemble clustering method (i.e., LWGP) only achieves average NMI($\%$) and CA($\%$) scores of $66.62$ and $74.49$, respectively. Similar advantages of U-SENC can also be observed in the normalized average scores. In terms of average rank, U-SENC obtains an average rank of 1.00 w.r.t. both NMI and CA, while the second best method obtains an average rank of 2.80 w.r.t. NMI and 2.90 w.r.t. CA.

In Table~\ref{table:compare_ens_time}, the time costs of different ensemble clustering methods are provided. As can be seen in Table~\ref{table:compare_ens_time}, the proposed U-SENC method has shown its advantage in efficiency over the other ensemble clustering methods, especially on the large-scale datasets whose data sizes go beyond millions.

\subsection{Parameters Analysis}
\label{sec:para_analysis}

\begin{table}
\centering
\caption{Average NMI(\%), CA(\%), and time costs(s) over 20 runs by different methods with varying number of representatives $p$.}
\label{table:compare_para_p}
\begin{threeparttable}
\begin{tabular}{m{0.75cm}<{\centering}|m{1.45cm}<{\centering}m{1.45cm}<{\centering}m{1.45cm}<{\centering}m{1.55cm}<{\centering}}
\toprule
\emph{Dataset}  &\emph{MNIST}  &\emph{Covertype}  &\emph{TB-1M}  &\emph{SF-2M}\\
\midrule
\multirow{1}{*}{NMI}
&\includegraphics[width=1.7cm]{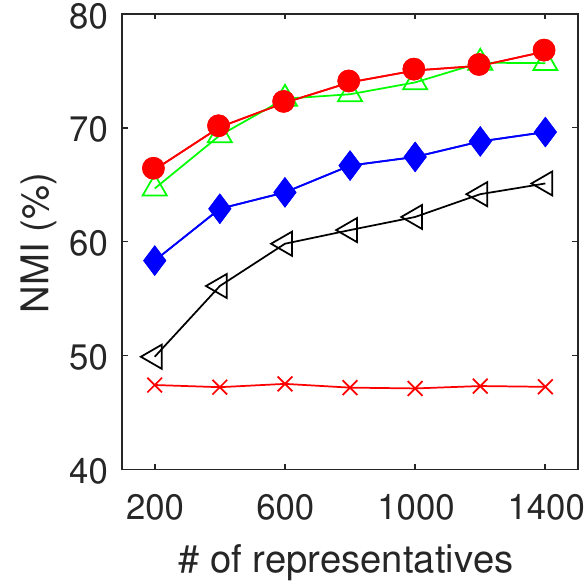}
&\includegraphics[width=1.7cm]{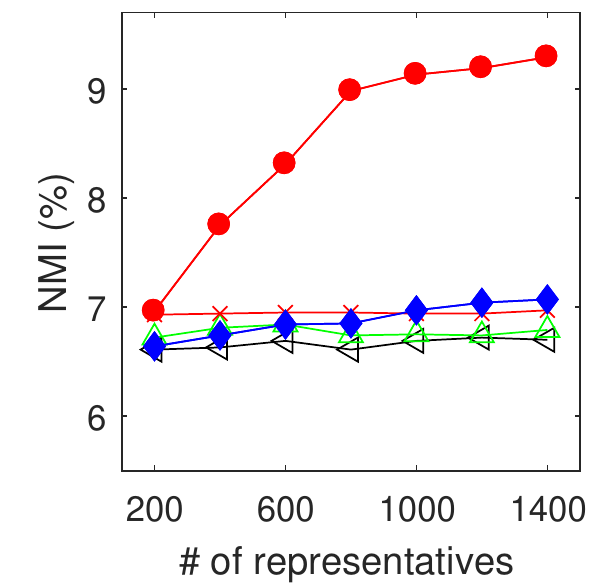}
&\includegraphics[width=1.7cm]{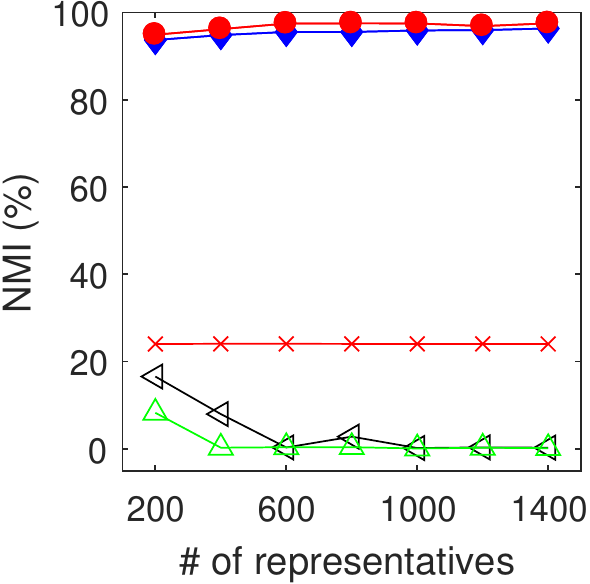}
&\includegraphics[width=1.7cm]{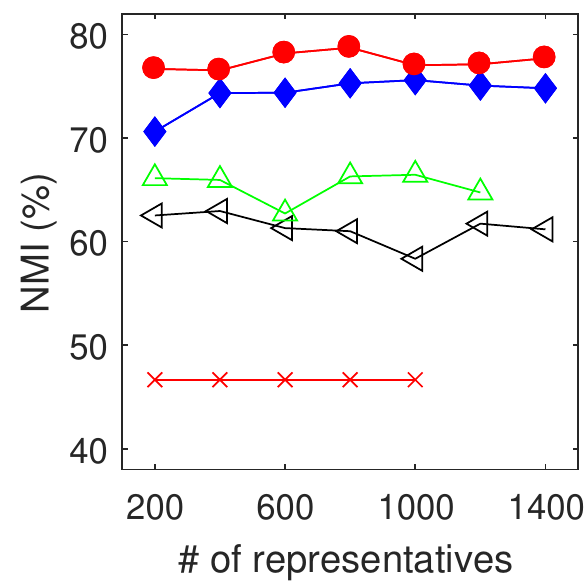}\\
CA
&\includegraphics[width=1.7cm]{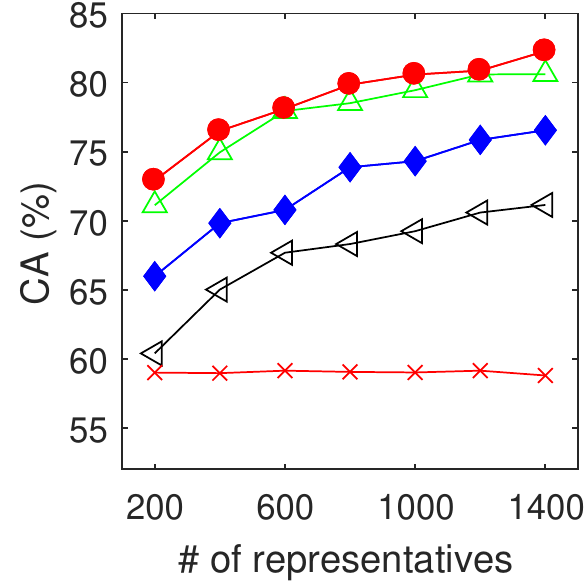}
&\includegraphics[width=1.7cm]{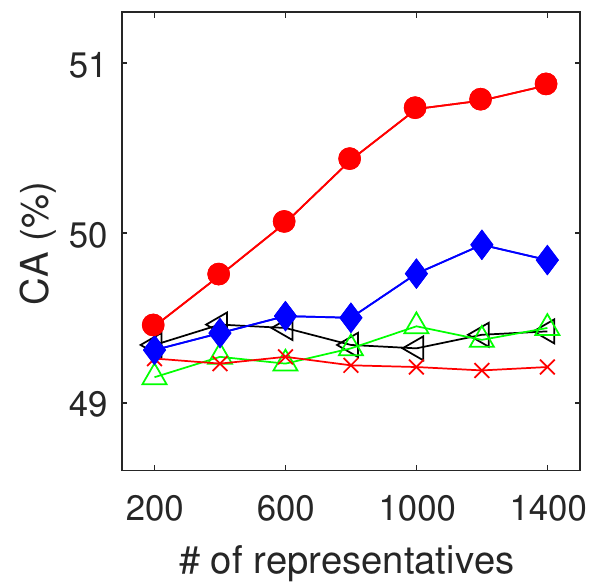}
&\includegraphics[width=1.7cm]{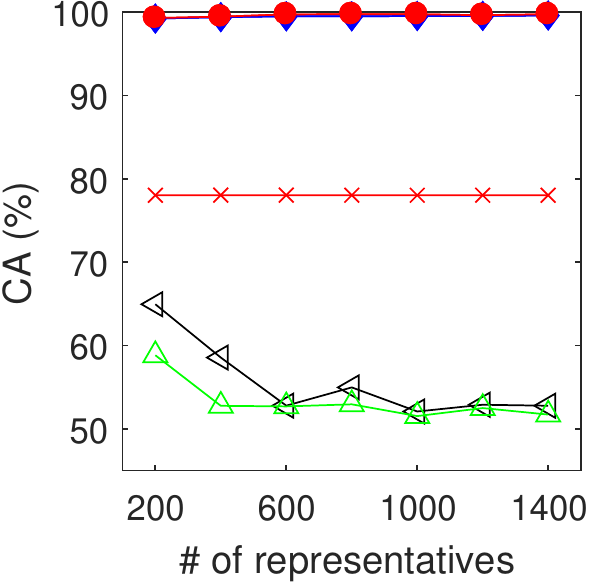}
&\includegraphics[width=1.7cm]{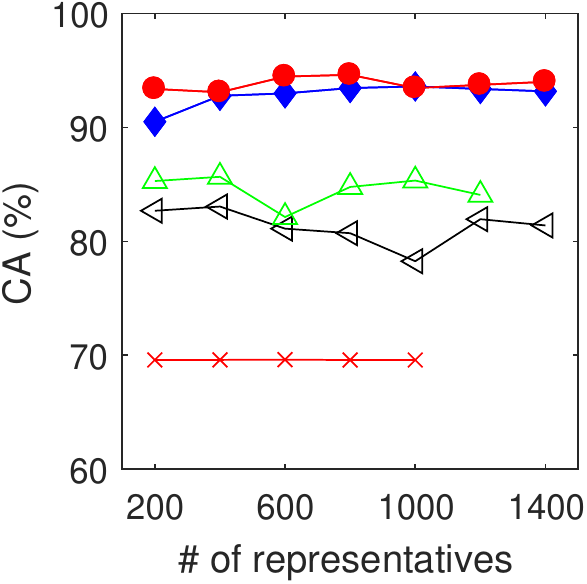}\\
Time cost
&\includegraphics[width=1.7cm]{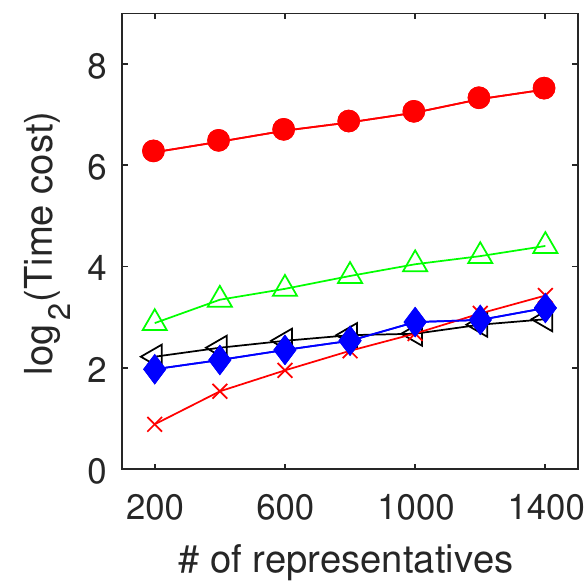}
&\includegraphics[width=1.7cm]{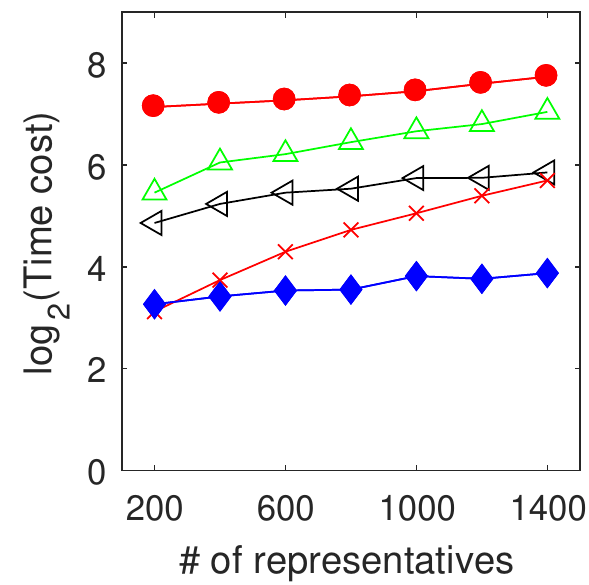}
&\includegraphics[width=1.7cm]{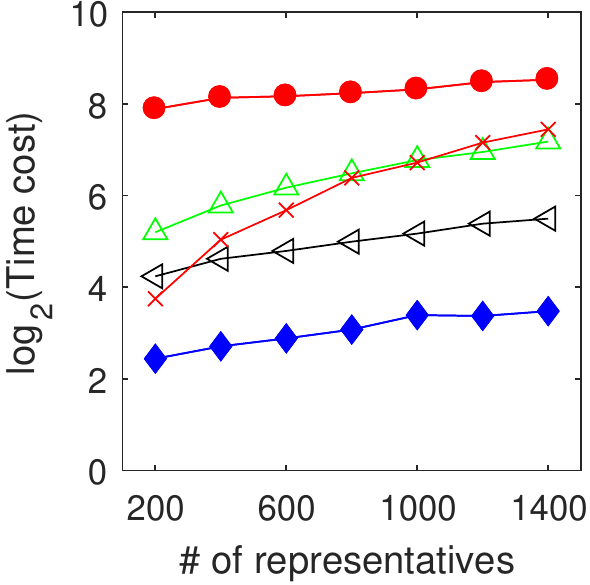}
&\includegraphics[width=1.7cm]{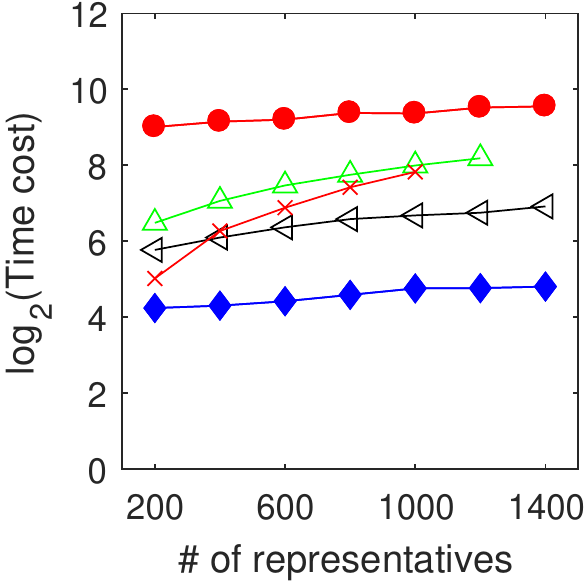}\\
&\multicolumn{4}{c}{\includegraphics[width=6cm]{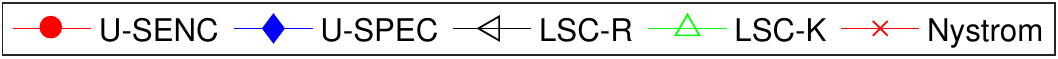}}\\
\bottomrule
\end{tabular}
\begin{tablenotes}
\item[*] On the \emph{SF-2M} dataset, LSC-K cannot handle $\geq 1400$ representatives (or landmarks), while Nystr\"{o}m cannot handle $\geq 1200$ representatives (or landmarks), due to the memory bottleneck.
\end{tablenotes}
\end{threeparttable}
\end{table}

\begin{table}
\centering
\caption{Average NMI(\%), CA(\%), and time costs(s) over 20 runs by different methods with varying number of nearest representatives $K$.}
\label{table:compare_para_Knn}
\begin{threeparttable}
\begin{tabular}{m{0.75cm}<{\centering}|m{1.45cm}<{\centering}m{1.45cm}<{\centering}m{1.45cm}<{\centering}m{1.55cm}<{\centering}}
\toprule
\emph{Dataset}  &\emph{MNIST}  &\emph{Covertype}  &\emph{TB-1M}  &\emph{SF-2M}\\
\midrule
\multirow{1}{*}{NMI}
&\includegraphics[width=1.7cm]{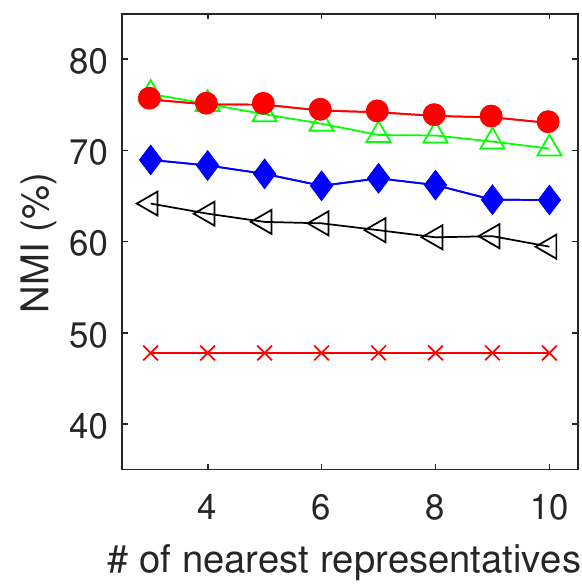}
&\includegraphics[width=1.7cm]{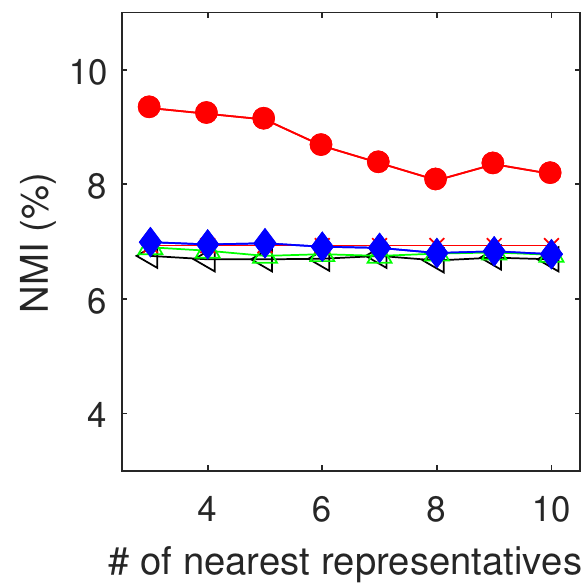}
&\includegraphics[width=1.7cm]{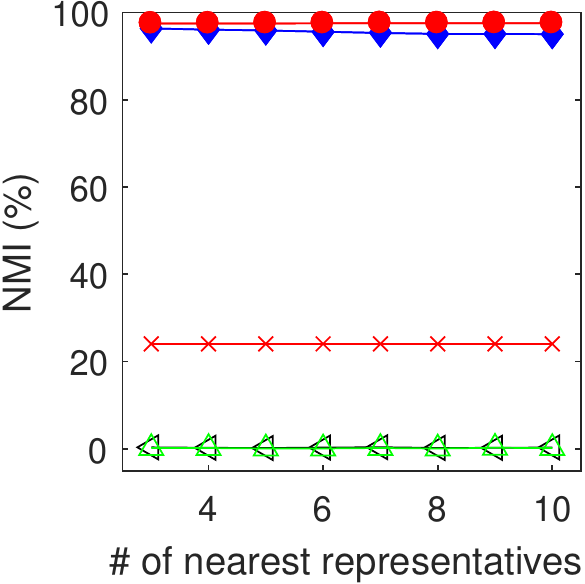}
&\includegraphics[width=1.7cm]{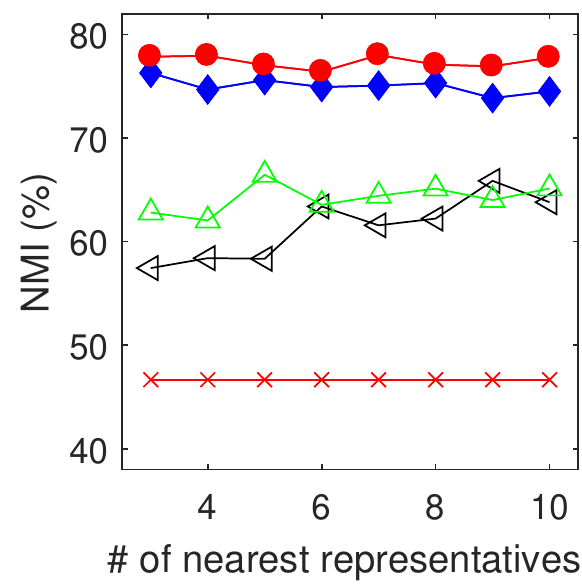}\\
CA
&\includegraphics[width=1.7cm]{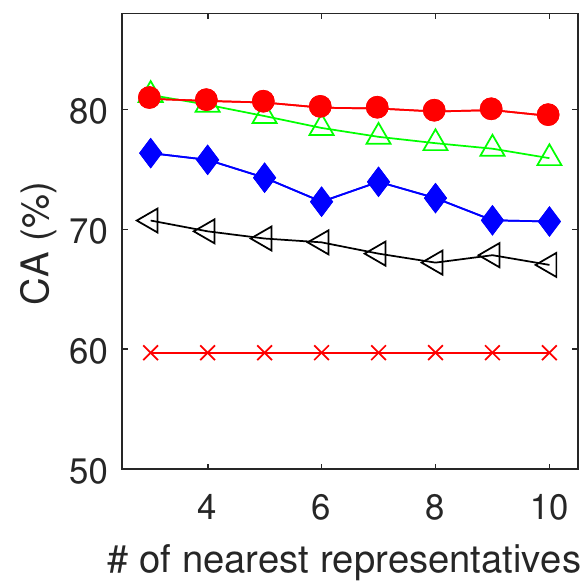}
&\includegraphics[width=1.7cm]{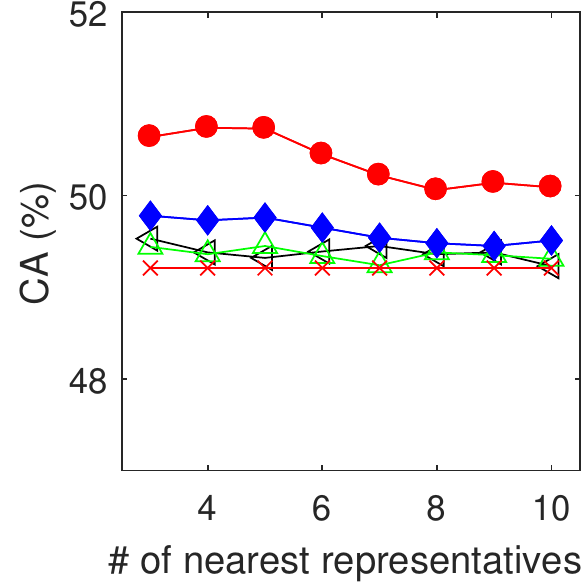}
&\includegraphics[width=1.7cm]{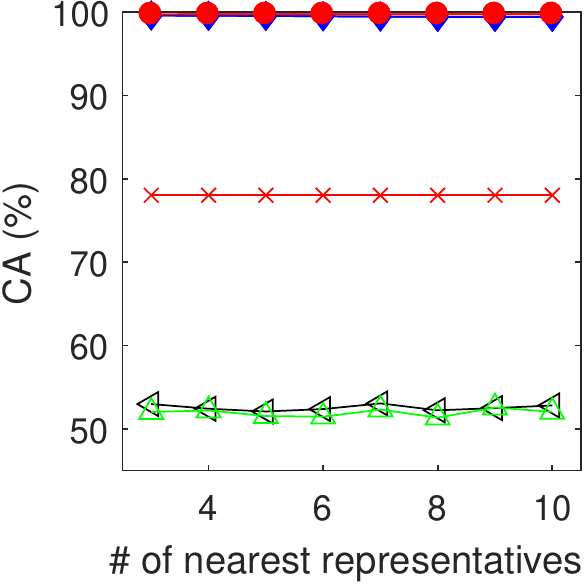}
&\includegraphics[width=1.7cm]{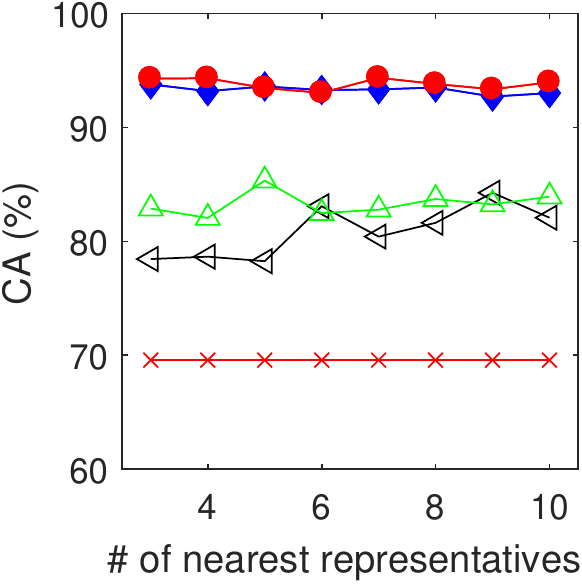}\\
Time cost
&\includegraphics[width=1.7cm]{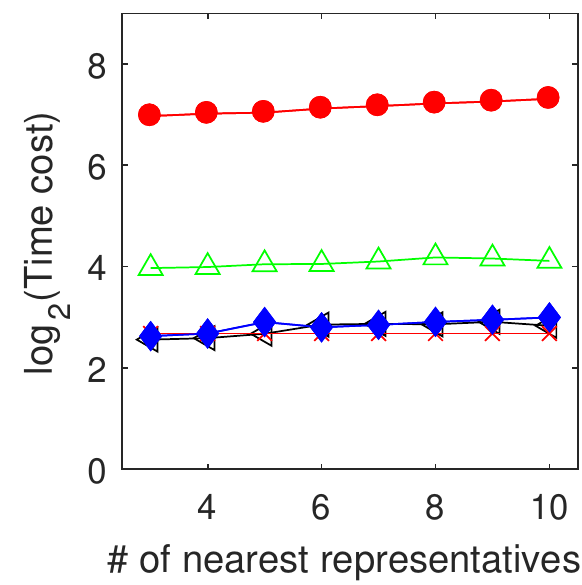}
&\includegraphics[width=1.7cm]{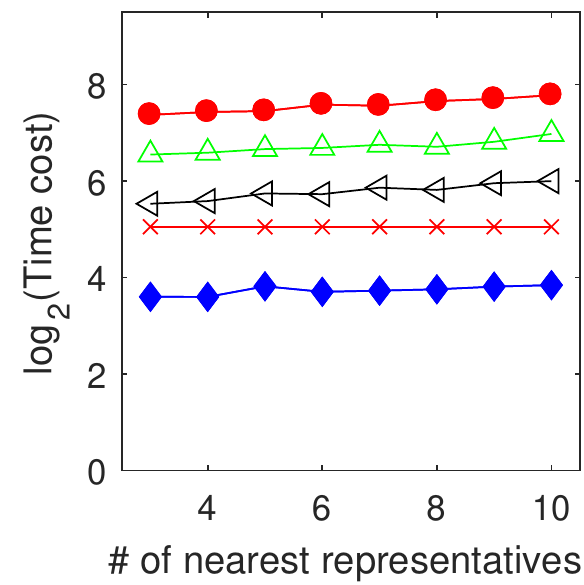}
&\includegraphics[width=1.7cm]{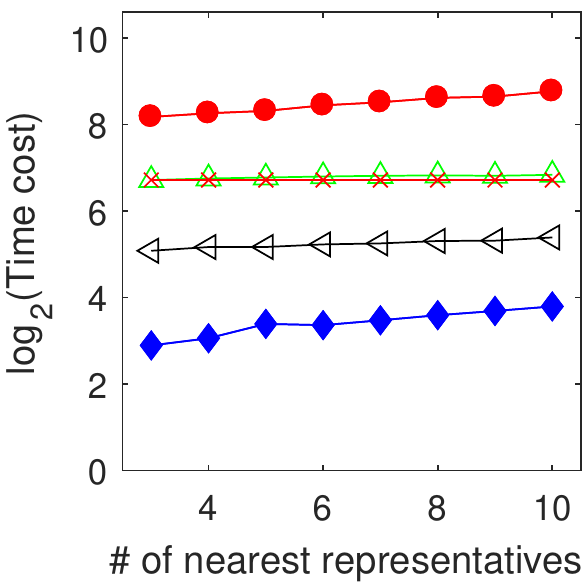}
&\includegraphics[width=1.7cm]{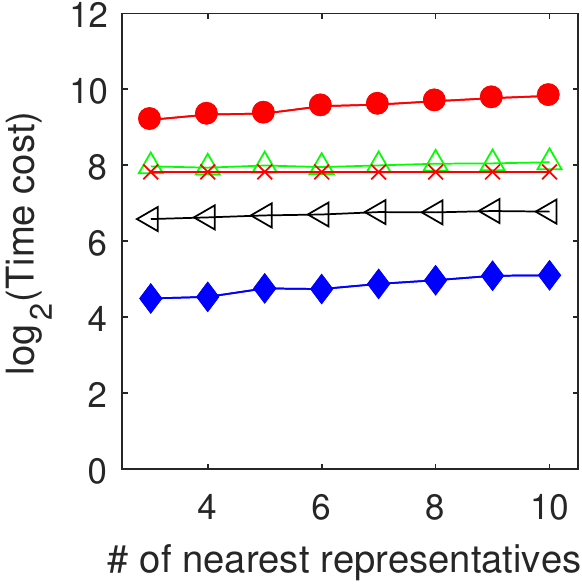}\\
&\multicolumn{4}{c}{\includegraphics[width=6cm]{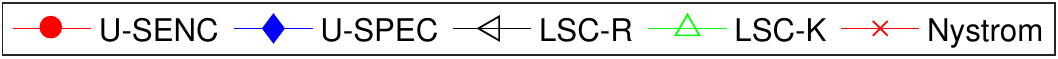}}\\
\bottomrule
\end{tabular}
\end{threeparttable}
\end{table}

\begin{table}
\centering
\caption{Average NMI(\%), CA(\%), and time costs(s) over 20 runs by different methods with varying ensemble size $m$.}
\label{table:compare_para_Msize}
\begin{threeparttable}
\begin{tabular}{m{0.75cm}<{\centering}|m{1.45cm}<{\centering}m{1.45cm}<{\centering}m{1.45cm}<{\centering}m{1.55cm}<{\centering}}
\toprule
\emph{Dataset}  &\emph{MNIST}  &\emph{Covertype}  &\emph{TB-1M}  &\emph{SF-2M}\\
\midrule
\multirow{1}{*}{NMI}
&\includegraphics[width=1.7cm]{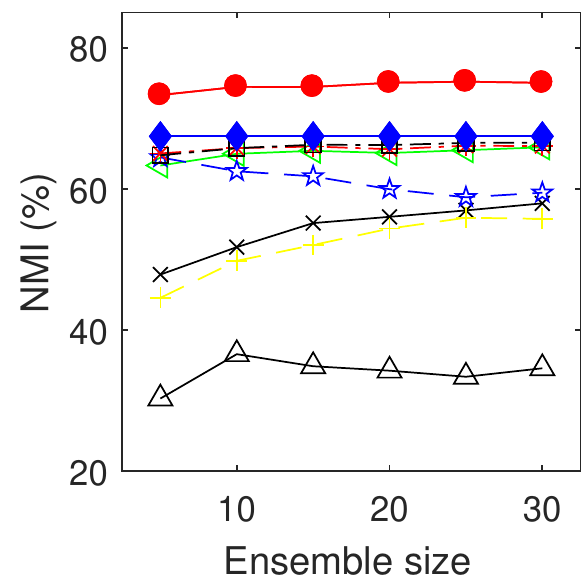}
&\includegraphics[width=1.7cm]{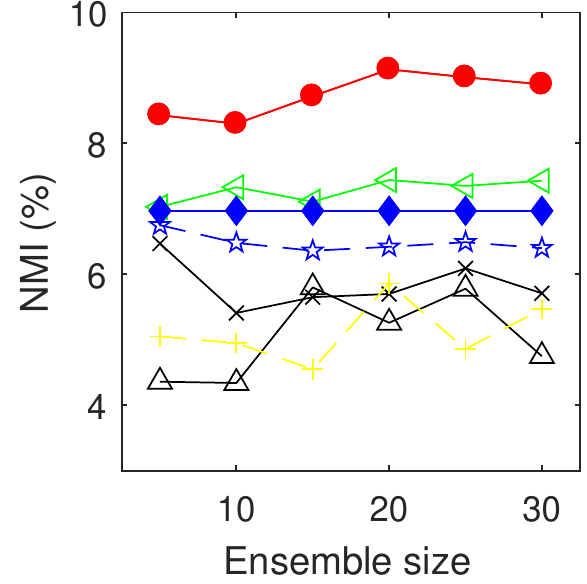}
&\includegraphics[width=1.7cm]{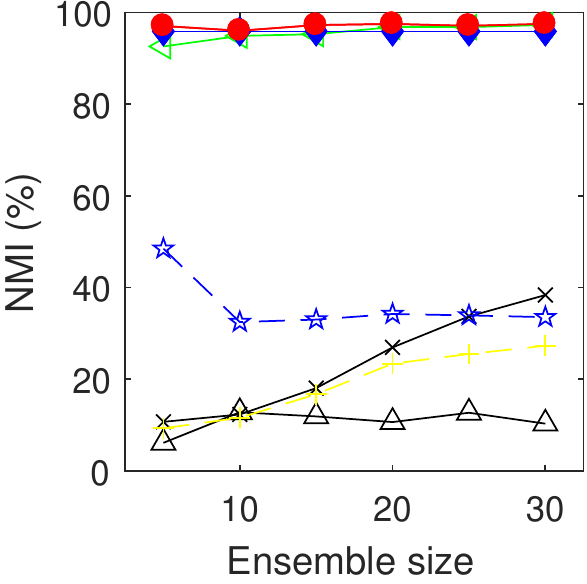}
&\includegraphics[width=1.7cm]{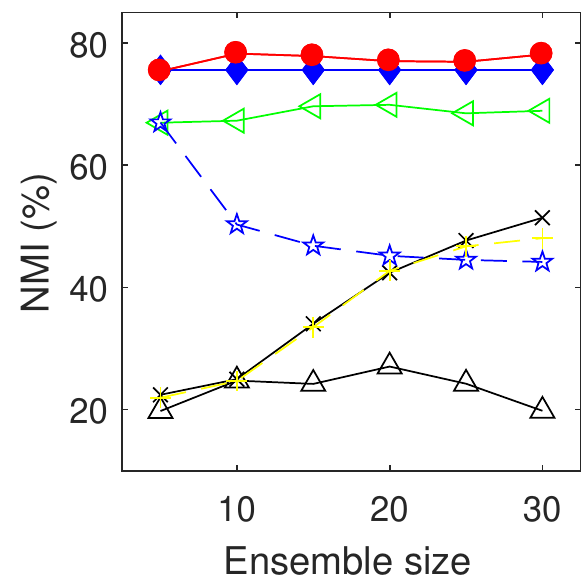}\\
CA
&\includegraphics[width=1.7cm]{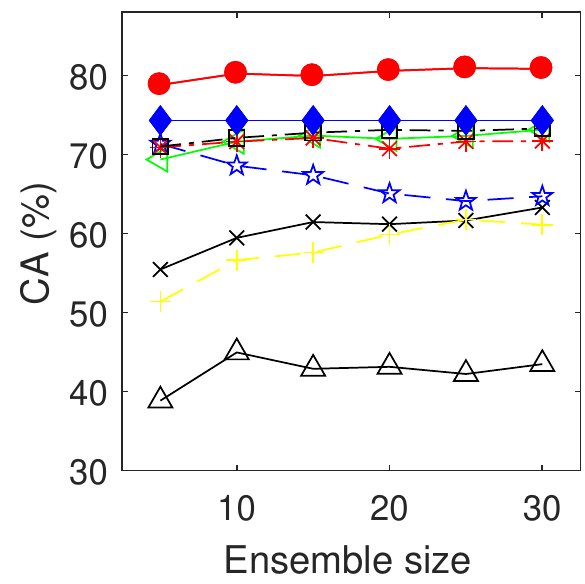}
&\includegraphics[width=1.7cm]{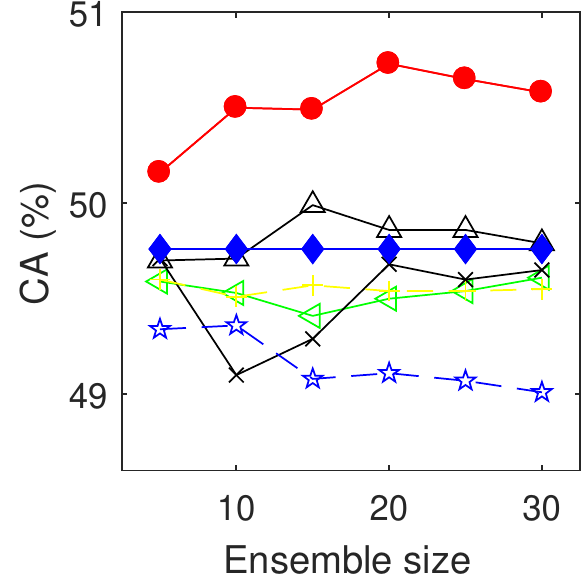}
&\includegraphics[width=1.7cm]{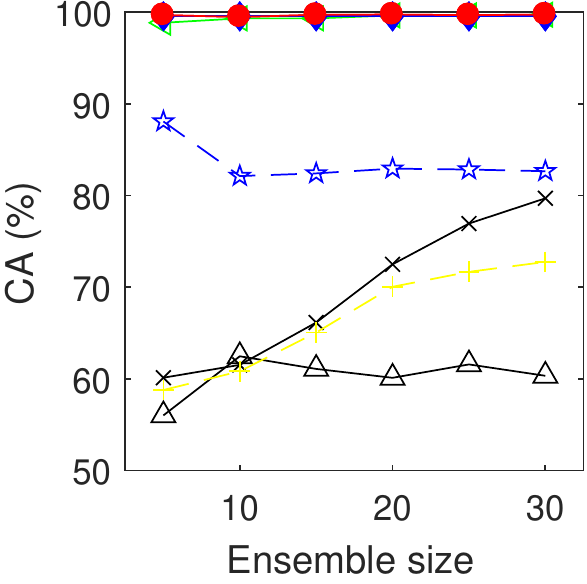}
&\includegraphics[width=1.7cm]{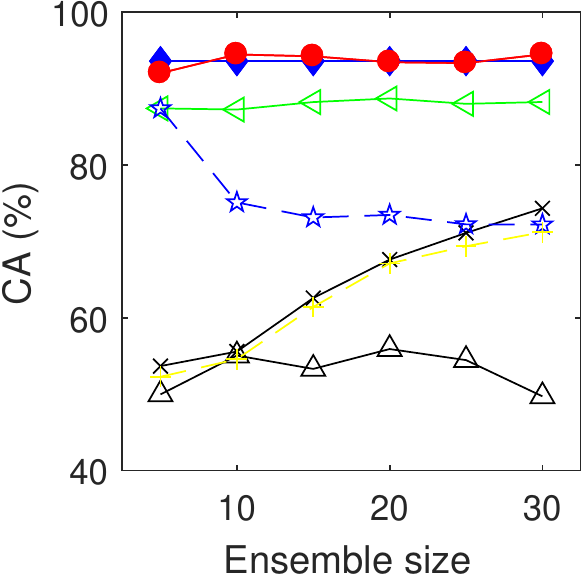}\\
Time cost
&\includegraphics[width=1.7cm]{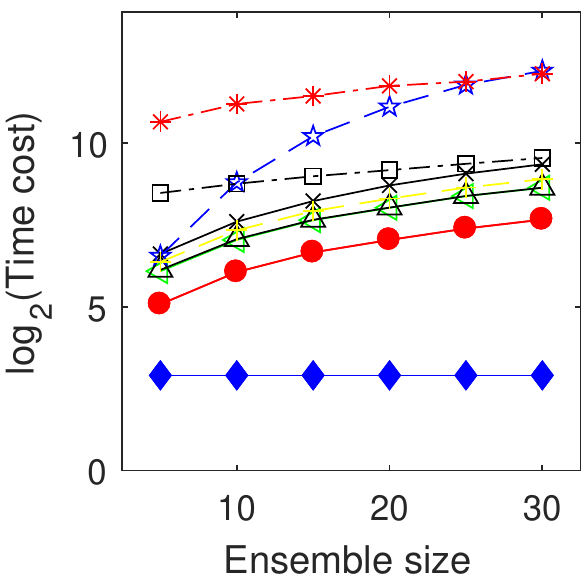}
&\includegraphics[width=1.7cm]{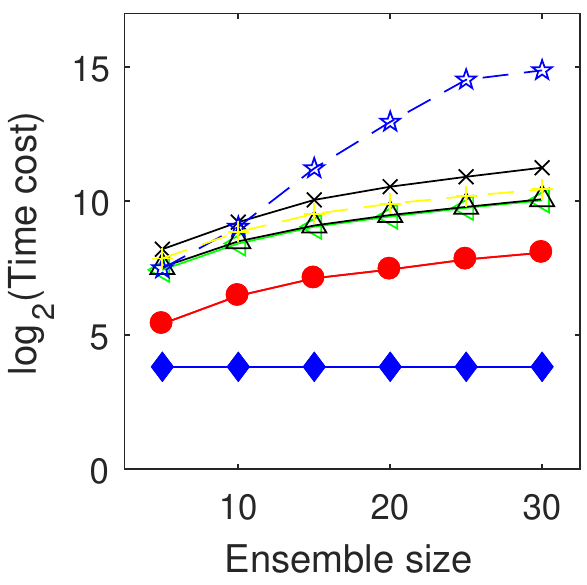}
&\includegraphics[width=1.7cm]{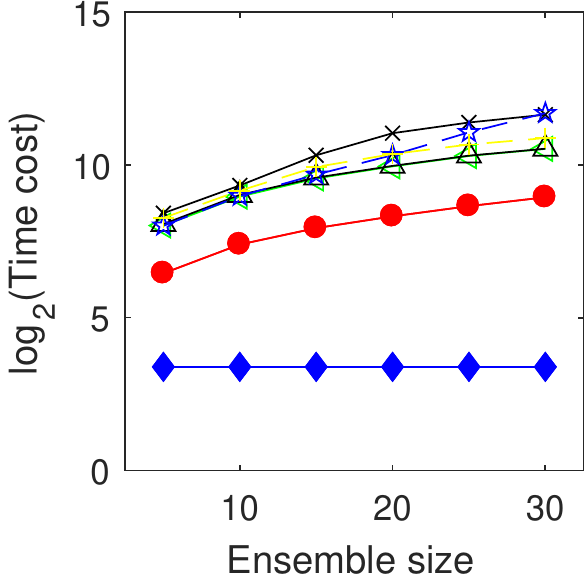}
&\includegraphics[width=1.7cm]{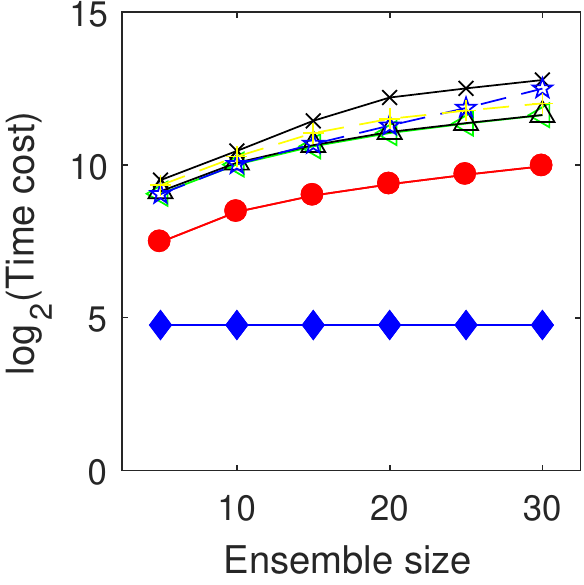}\\
&\multicolumn{4}{c}{\includegraphics[width=7.3cm]{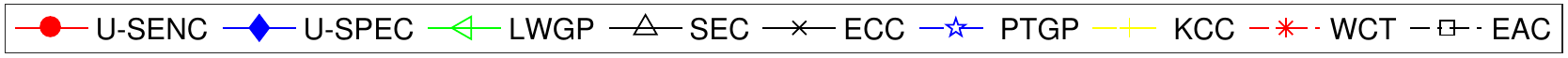}}\\
\bottomrule
\end{tabular}
\end{threeparttable}
\end{table}

In this section, we evaluate the performances of our algorithms and several baseline algorithms with varying parameters. Because some important baseline methods (such as Nystr\"{o}m, LSC-K, and LSC-R) can not go beyond two-million-level datasets, in order to fairly test the influence of some common parameters among them, we perform the parameter analysis on four benchmark datasets, namely, \emph{MNIST}, \emph{Covertype}, \emph{TB-1M}, and \emph{SF-2M}, which are the largest four datasets whose sizes are no larger than two million.

\subsubsection{Number of Representatives $p$}
\label{sec:para_p}

The parameter $p$ denotes the number of representatives (or landmarks), which is a common parameter in the sub-matrix based spectral clustering methods, such as Nystr\"{o}m, LSC-K, LSC-R, and our U-SPEC and U-SENC methods. As can be seen in Table~\ref{table:compare_para_p}, a larger $p$ generally leads to better performance, but also brings in an increasing time cost. In terms of NMI and CA, our U-SENC method consistently outperforms the other methods with varying parameter $p$ on all of the four datasets. The LSC-K outperforms U-SPEC on the \emph{MNIST} dataset. But on all the other three datasets, U-SPEC achieves better or significantly better NMI and CA scores than LSC-K. In terms of computational cost, the LSC-K and Nystr\"{o}m methods cannot deal with $p\geq 1,400$ representatives on the \emph{SF-2M} dataset with two million objects. On the benchmark datasets, U-SPEC is overall the fastest method with varying parameter $p$ (as shown in Table~\ref{table:compare_para_p}).

\subsubsection{Number of Nearest Representatives $K$}
\label{sec:para_K}

The parameter $K$ denotes the number of nearest representatives (or landmarks), which is a common parameter in LSC-K, LSC-R, and our U-SPEC and U-SENC methods. Note that the Nystr\"{o}m method doesn't have such a parameter $K$, but we still illustrate the performance of Nystr\"{o}m in Table~\ref{table:compare_para_Knn} just to use Nystr\"{o}m as a benchmark here. As illustrated in Table~\ref{table:compare_para_Knn}, on the \emph{MNIST} dataset, U-SENC and LSC-K are respectively the best and the second best methods w.r.t. NMI and CA, while U-SPEC is the third best method. On all of the other three benchmark datasets, U-SENC and U-SPEC are overall the best two methods w.r.t. both NMI and CA with varying parameter $K$ (as shown in Table~\ref{table:compare_para_Knn}).

\subsubsection{Ensemble Size $m$}
\label{sec:para_M}

The parameter $m$ denotes the number of base clusterings, which is a common parameter in all of the ensemble clustering methods, including U-SENC as well as the baseline ensemble clustering methods. Note that U-SPEC is not an ensemble clustering method and doesn't have the parameter $m$, but we still illustrate the performance of U-SPEC in Table~\ref{table:compare_para_Msize} for reference only. As shown in Table~\ref{table:compare_para_Msize}, U-SENC outperforms, or even significantly outperforms, the other ensemble clustering methods w.r.t. both NMI and CA on the benchmark datasets with varying ensemble size $m$. Meanwhile, U-SENC consistently requires a lower computational cost than the other ensemble clustering methods.

\subsection{Influence of Representative Selection Strategies}
\label{sec:cmpSelStrat}

In this section, we compare the performances of our algorithms using different representative selection strategies. Specifically, Table~\ref{table:compare_sel_strategies_USPEC} illustrates the performances of U-SPEC using hybrid selection (U-SPEC-H), U-SPEC using random selection (U-SPEC-R), and U-SPEC using $k$-means based selection (U-SPEC-K), whereas Table~\ref{table:compare_sel_strategies_USENC} illustrates the performances of U-SENC using hybrid selection (U-SENC-H), U-SENC using random selection (U-SENC-R), and U-SENC using $k$-means based selection (U-SENC-K). As shown in Tables~\ref{table:compare_sel_strategies_USPEC} and \ref{table:compare_sel_strategies_USENC}, the random representative selection is very efficient compared to $k$-means based selection, but may degrade the clustering quality due to its inherent instability. The $k$-means based selection generally leads to better clustering quality than random selection, but brings in a much larger computational cost. Compared to random selection and $k$-means based selection, our hybrid selection strategy strikes a balance between efficiency and clustering robustness. It achieves comparable efficiency to the random selection and significantly better efficiency than the $k$-means based selection, and also yields competitive clustering quality as compared to the $k$-means based selection.

\begin{table}
\centering
\caption{The NMI(\%), CA(\%), and time costs(s) by U-SPEC using different representative selection strategies (\textbf{H}: hybrid selection; \textbf{R}: random selection; \textbf{K}: $K$-means based selection).}
\label{table:compare_sel_strategies_USPEC}
\begin{threeparttable}
\begin{tabular}{m{0.75cm}<{\centering}|m{1.45cm}<{\centering}m{1.45cm}<{\centering}m{1.45cm}<{\centering}m{1.55cm}<{\centering}}
\toprule
\emph{Dataset}  &\emph{MNIST}  &\emph{Covertype}  &\emph{TB-1M}  &\emph{SF-2M}\\
\midrule
\multirow{1}{*}{NMI}
&\includegraphics[width=1.7cm]{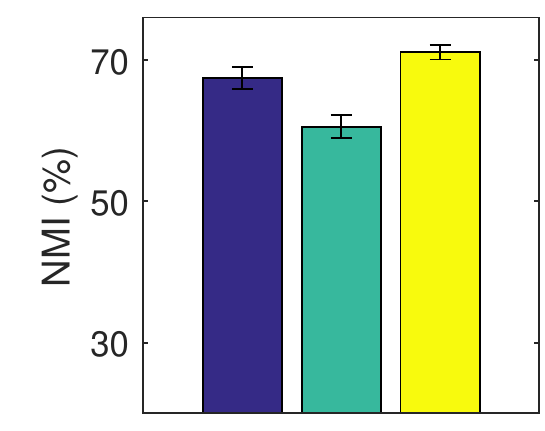}
&\includegraphics[width=1.7cm]{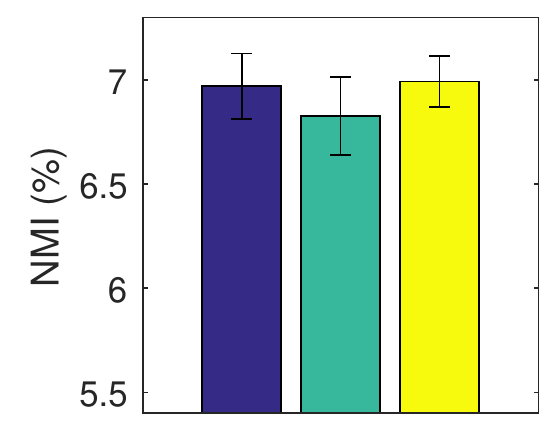}
&\includegraphics[width=1.7cm]{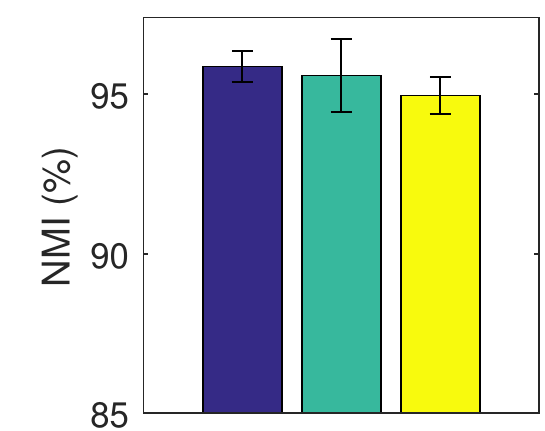}
&\includegraphics[width=1.7cm]{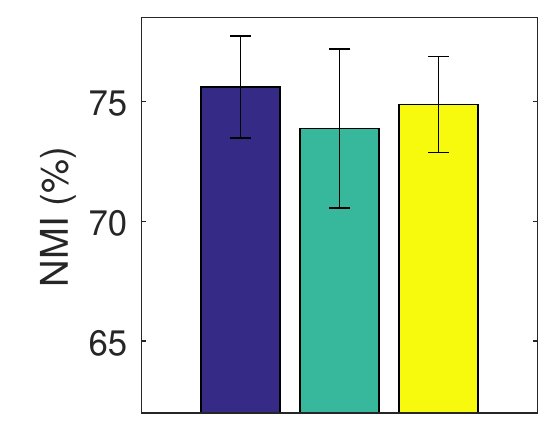}\\
CA
&\includegraphics[width=1.7cm]{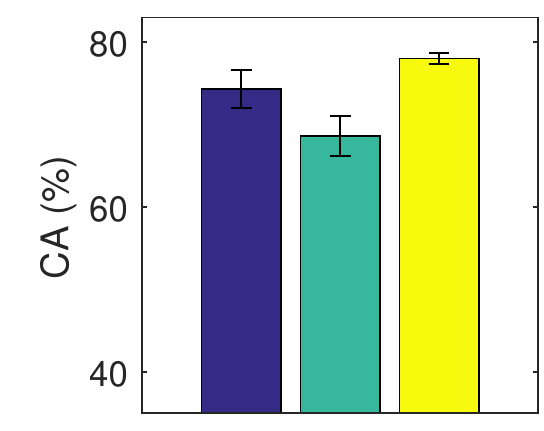}
&\includegraphics[width=1.7cm]{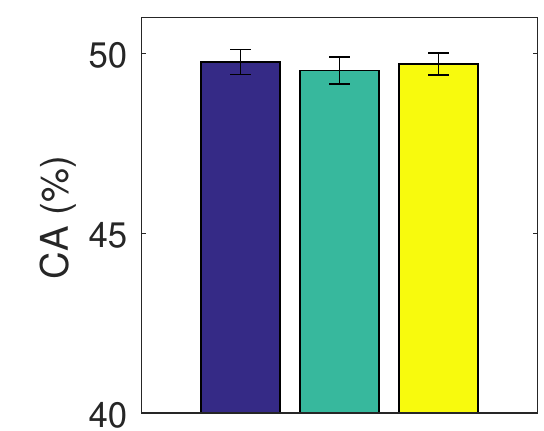}
&\includegraphics[width=1.7cm]{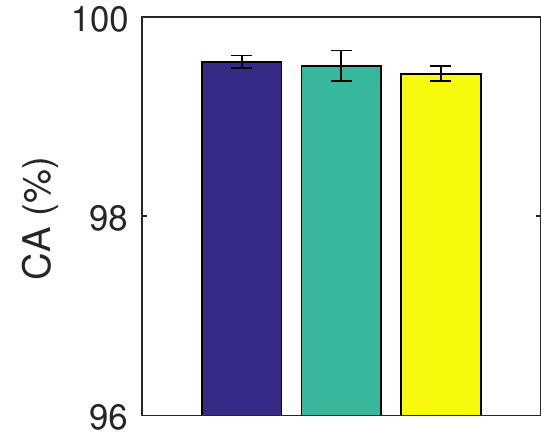}
&\includegraphics[width=1.7cm]{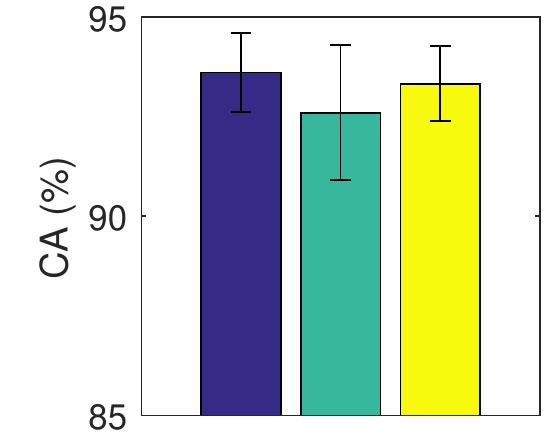}\\
Time cost
&\includegraphics[width=1.7cm]{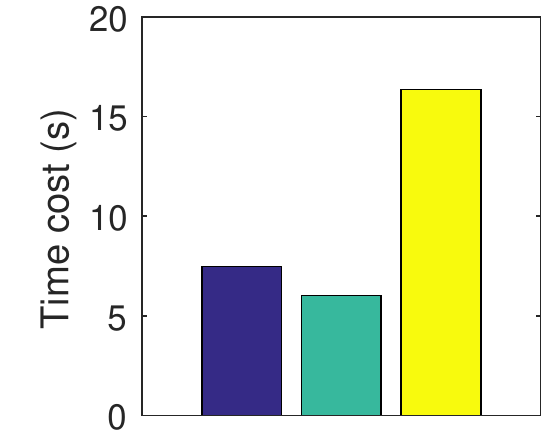}
&\includegraphics[width=1.7cm]{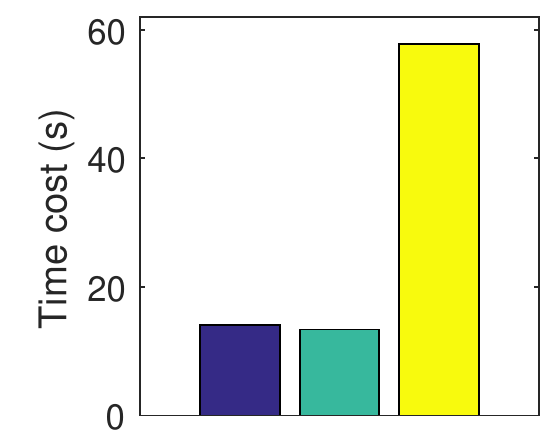}
&\includegraphics[width=1.7cm]{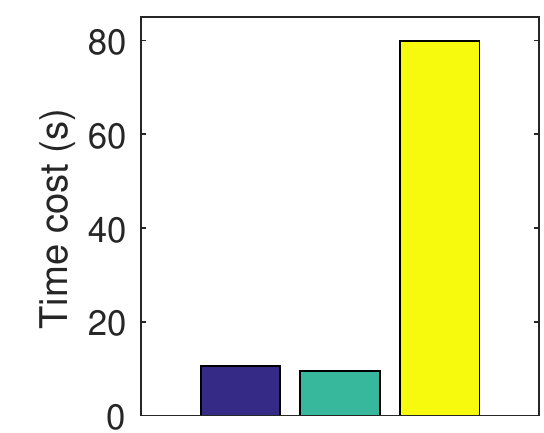}
&\includegraphics[width=1.7cm]{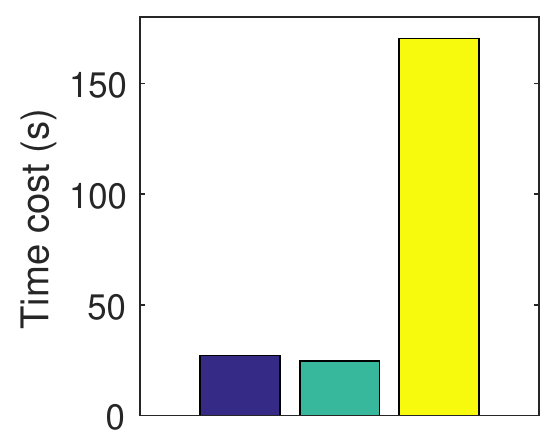}\\
&\multicolumn{4}{c}{\includegraphics[width=4.455cm]{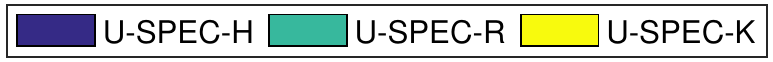}}\\
\bottomrule
\end{tabular}
\end{threeparttable}
\end{table}

\begin{table}
\centering
\caption{The NMI(\%), CA(\%), and time costs(s) by U-SENC using different representative selection strategies (\textbf{H}: hybrid selection; \textbf{R}: random selection; \textbf{K}: $K$-means based selection).}
\label{table:compare_sel_strategies_USENC}
\begin{threeparttable}
\begin{tabular}{m{0.75cm}<{\centering}|m{1.45cm}<{\centering}m{1.45cm}<{\centering}m{1.45cm}<{\centering}m{1.55cm}<{\centering}}
\toprule
\emph{Dataset}  &\emph{MNIST}  &\emph{Covertype}  &\emph{TB-1M}  &\emph{SF-2M}\\
\midrule
\multirow{1}{*}{NMI}
&\includegraphics[width=1.7cm]{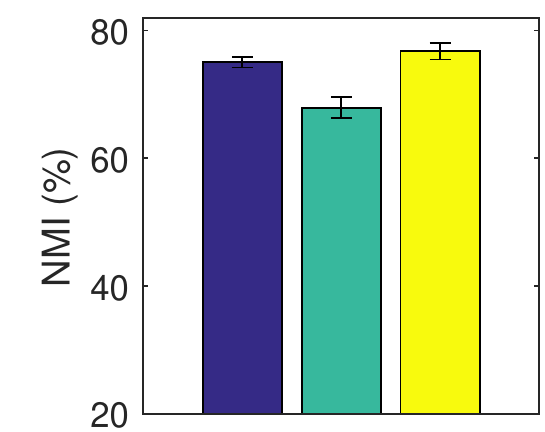}
&\includegraphics[width=1.7cm]{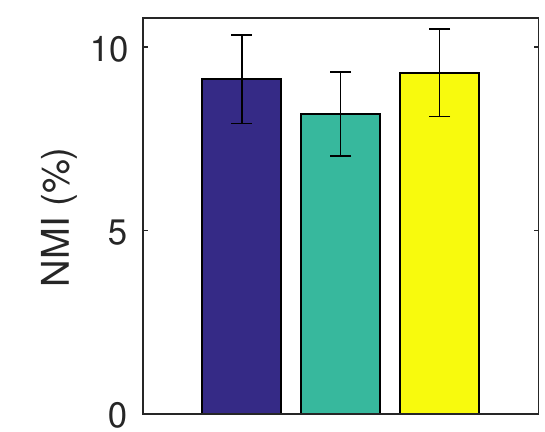}
&\includegraphics[width=1.7cm]{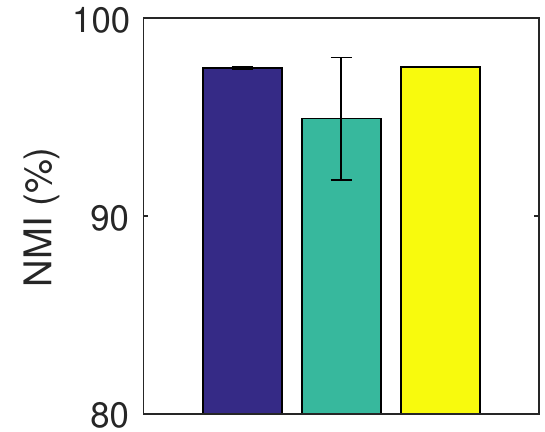}
&\includegraphics[width=1.7cm]{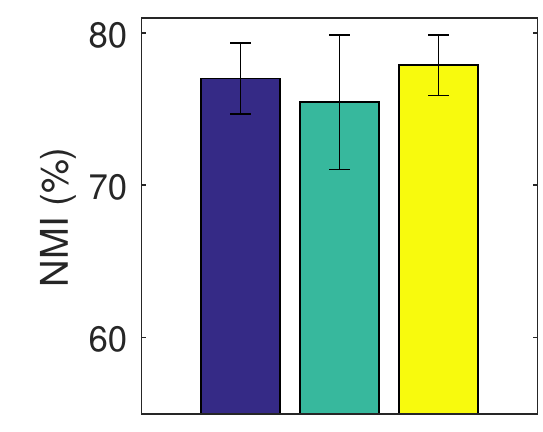}\\
CA
&\includegraphics[width=1.7cm]{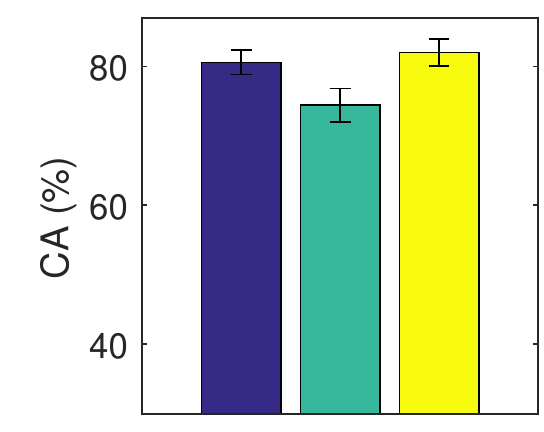}
&\includegraphics[width=1.7cm]{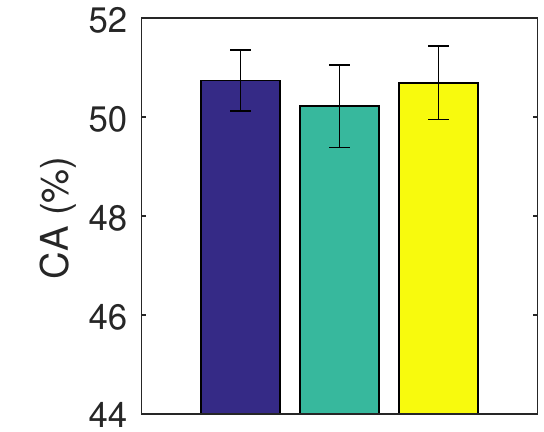}
&\includegraphics[width=1.7cm]{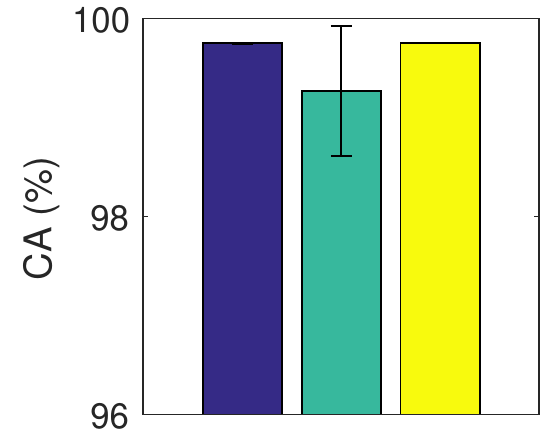}
&\includegraphics[width=1.7cm]{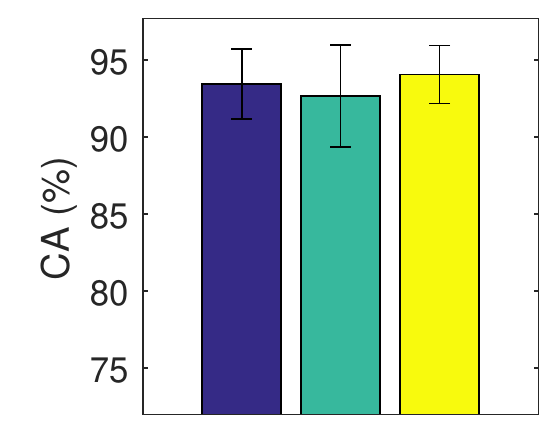}\\
Time cost
&\includegraphics[width=1.7cm]{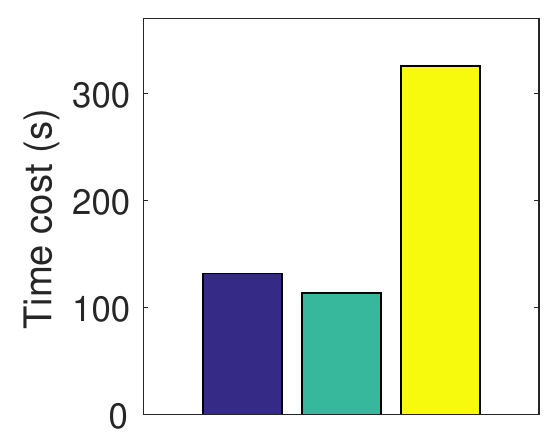}
&\includegraphics[width=1.7cm]{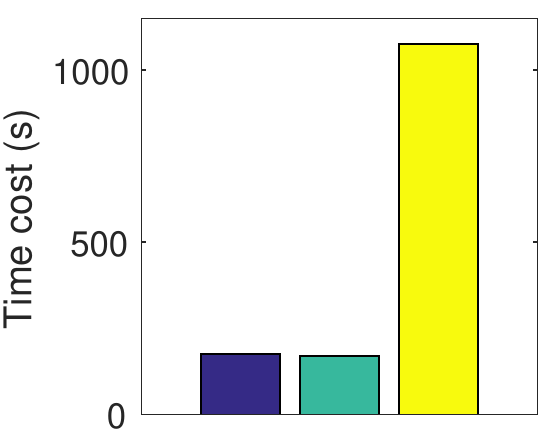}
&\includegraphics[width=1.7cm]{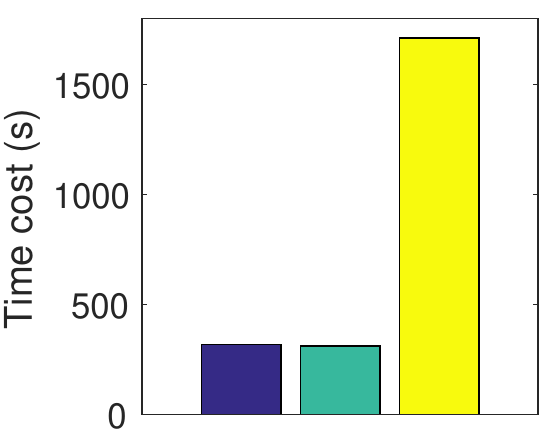}
&\includegraphics[width=1.7cm]{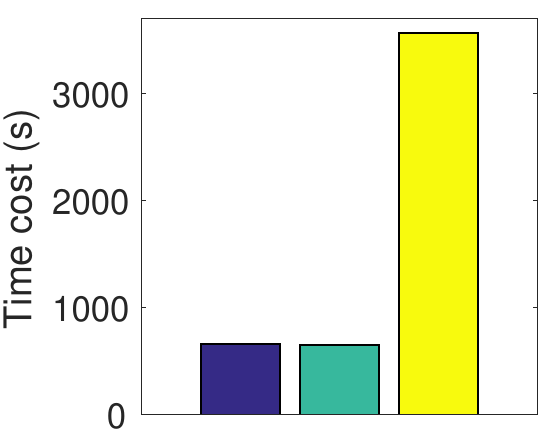}\\
&\multicolumn{4}{c}{\includegraphics[width=4.455cm]{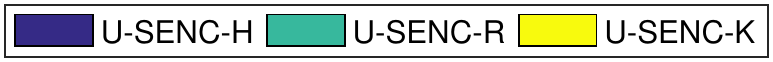}}\\
\bottomrule
\end{tabular}
\end{threeparttable}
\end{table}

\subsection{Influence of Approximate $K$-Nearest Neighbors}
\label{sec:cmpApproxKNN}

In this section, we compare our algorithms using \textbf{A}pproximate $K$-nearest representatives against using \textbf{E}xact $K$-nearest representatives, where four variants are evaluated, i.e., U-SPEC(A), U-SPEC(E), U-SENC(A), and U-SENC(E). The purpose of using approximate $K$-nearest representatives (see Section~\ref{sec:approx_knn}) is to alleviate the time and memory cost of the affinity sub-matrix construction while maintaining the overall clustering quality. As shown in Tables~\ref{table:compare_approxKNN_USPEC} and \ref{table:compare_approxKNN_USENC}, using approximate $K$-nearest representatives can achieve comparable clustering quality (w.r.t. NMI and CA) with using exact $K$-nearest representatives while alleviating the computational cost. As our approximation of $K$-nearest representatives reduces the time complexity from $O(Npd)$ to $O(Np^{\frac{1}{2}}d)$, the improvement in efficiency is more significant for high-dimensional datasets, such as the \emph{MNIST} dataset, whose dimension is 784. Even for the low-dimensional datasets, such as \emph{TB-1M} and \emph{SF-2M}, the use of approximate $K$-nearest representatives can still consistently reduce the time cost. Besides the time efficiency, the approximate $K$-nearest representatives also alleviate the memory burden. Specifically, on a machine with 64GB memory, the computation of conventional $K$-nearest representatives can hardly go beyond five million objects, whereas the proposed approximation method for $K$-nearest representatives can scale well for even ten-million-level datasets.

\begin{table}
\centering
\caption{The NMI(\%), CA(\%), and time costs(s) by U-SPEC using \textbf{A}pproximate $K$-nearest representatives against \textbf{E}xact $K$-nearest representatives.}
\label{table:compare_approxKNN_USPEC}
\begin{threeparttable}
\begin{tabular}{m{0.75cm}<{\centering}|m{1.45cm}<{\centering}m{1.45cm}<{\centering}m{1.45cm}<{\centering}m{1.55cm}<{\centering}}
\toprule
\emph{Dataset}  &\emph{MNIST}  &\emph{Covertype}  &\emph{TB-1M}  &\emph{SF-2M}\\
\midrule
\multirow{1}{*}{NMI}
&\includegraphics[width=1.7cm]{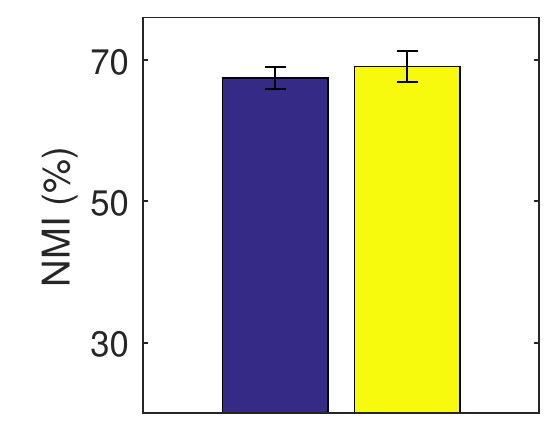}
&\includegraphics[width=1.7cm]{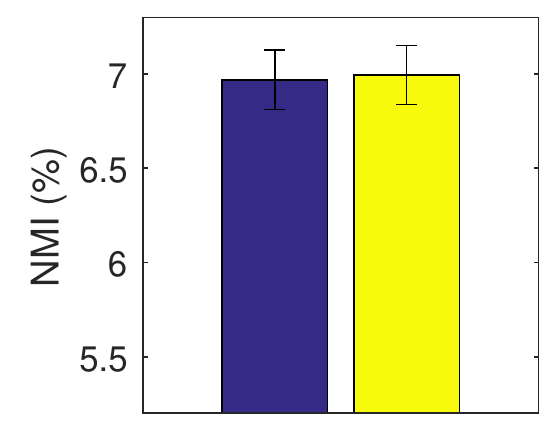}
&\includegraphics[width=1.7cm]{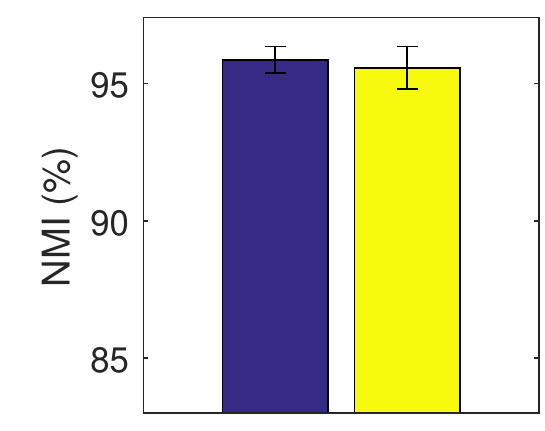}
&\includegraphics[width=1.7cm]{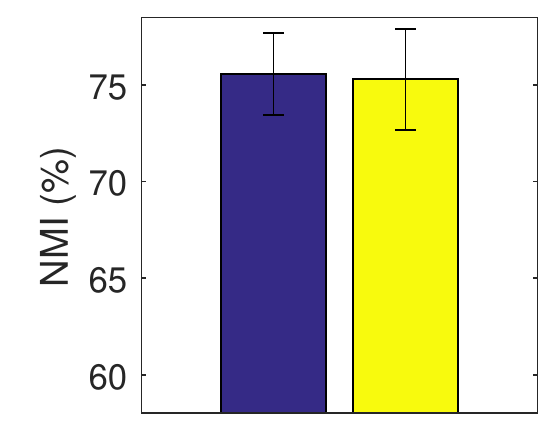}\\
CA
&\includegraphics[width=1.7cm]{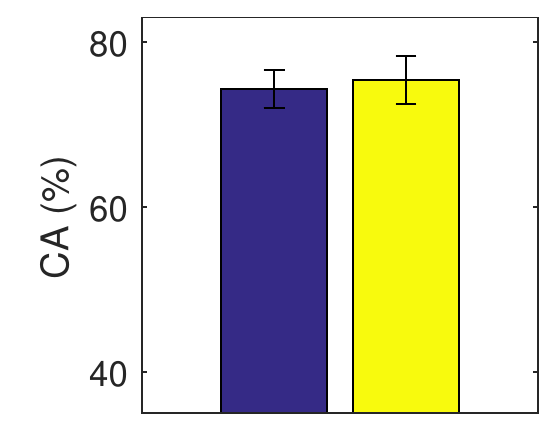}
&\includegraphics[width=1.7cm]{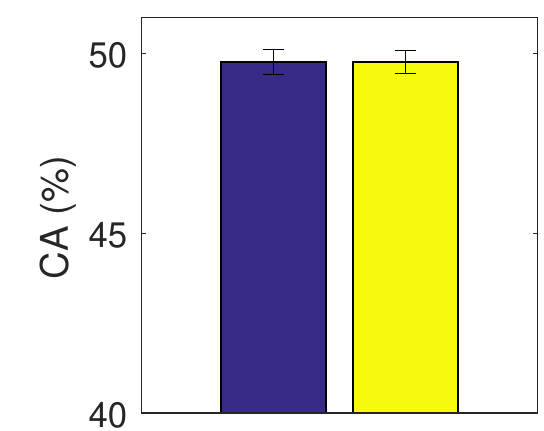}
&\includegraphics[width=1.7cm]{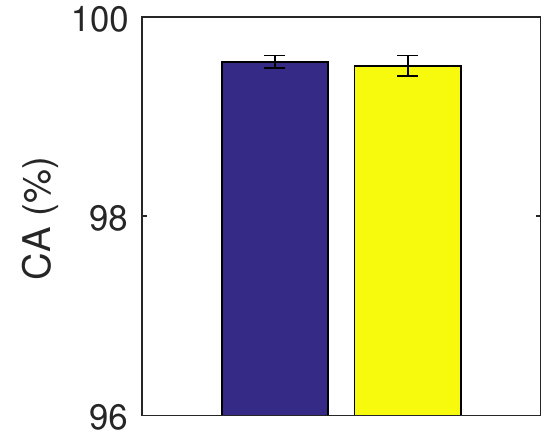}
&\includegraphics[width=1.7cm]{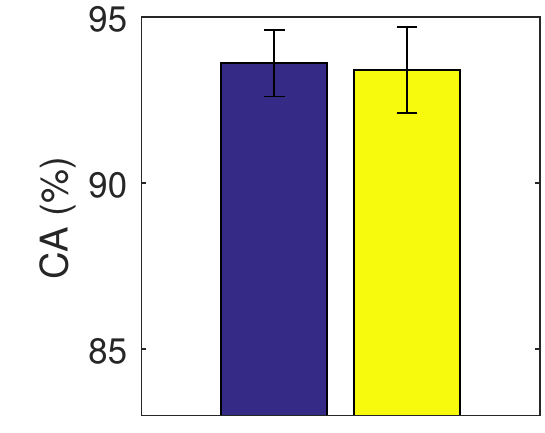}\\
Time cost
&\includegraphics[width=1.7cm]{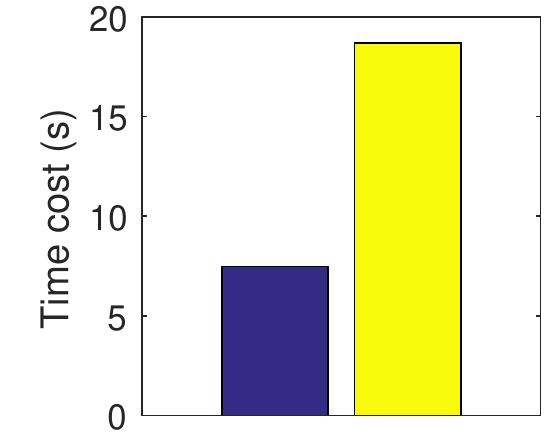}
&\includegraphics[width=1.7cm]{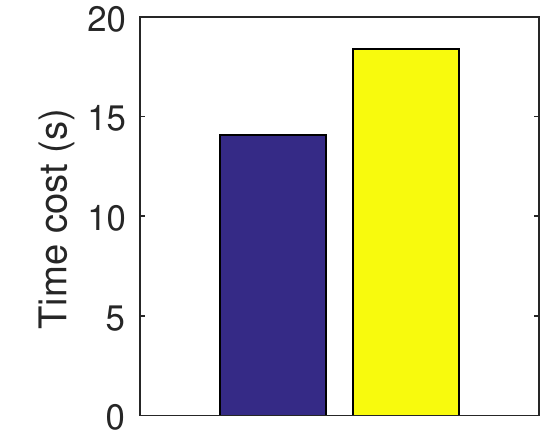}
&\includegraphics[width=1.7cm]{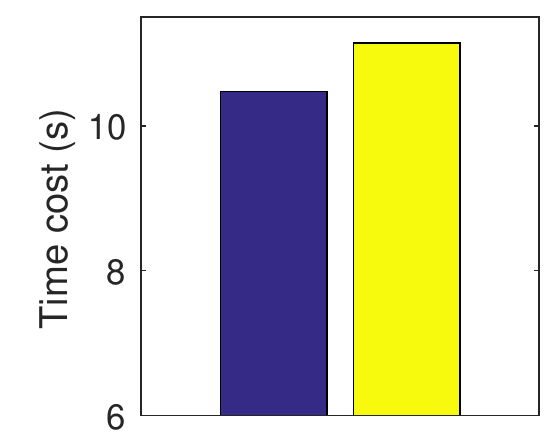}
&\includegraphics[width=1.7cm]{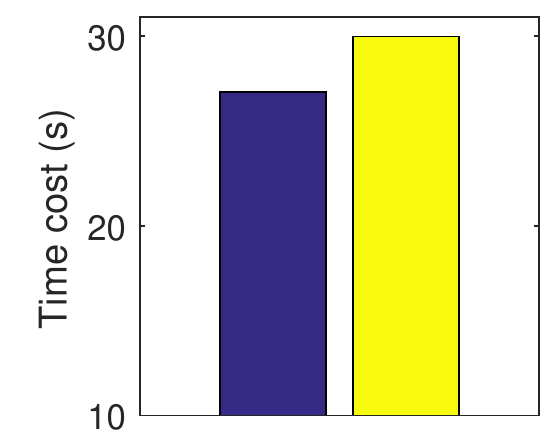}\\
&\multicolumn{4}{c}{\includegraphics[width=3.1cm]{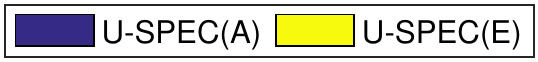}}\\
\bottomrule
\end{tabular}
\end{threeparttable}
\end{table}

\begin{table}
\centering
\caption{The NMI(\%), CA(\%), and time costs(s) by U-SENC using \textbf{A}pproximate $K$-nearest representatives against \textbf{E}xact $K$-nearest representatives.}
\label{table:compare_approxKNN_USENC}
\begin{threeparttable}
\begin{tabular}{m{0.75cm}<{\centering}|m{1.45cm}<{\centering}m{1.45cm}<{\centering}m{1.45cm}<{\centering}m{1.55cm}<{\centering}}
\toprule
\emph{Dataset}  &\emph{MNIST}  &\emph{Covertype}  &\emph{TB-1M}  &\emph{SF-2M}\\
\midrule
\multirow{1}{*}{NMI}
&\includegraphics[width=1.7cm]{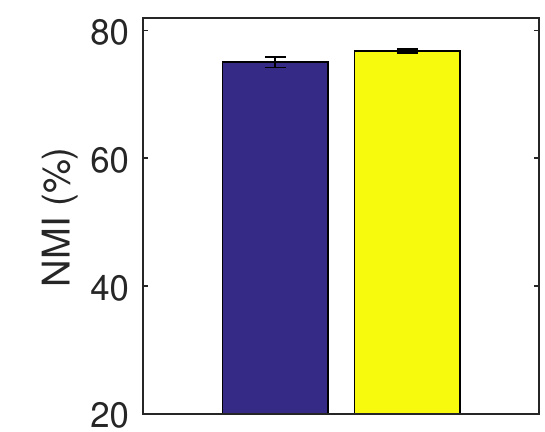}
&\includegraphics[width=1.7cm]{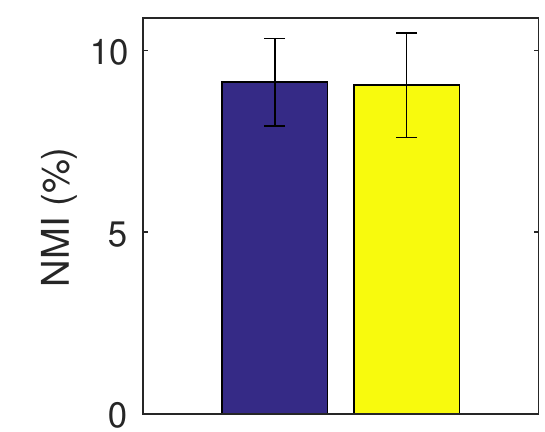}
&\includegraphics[width=1.7cm]{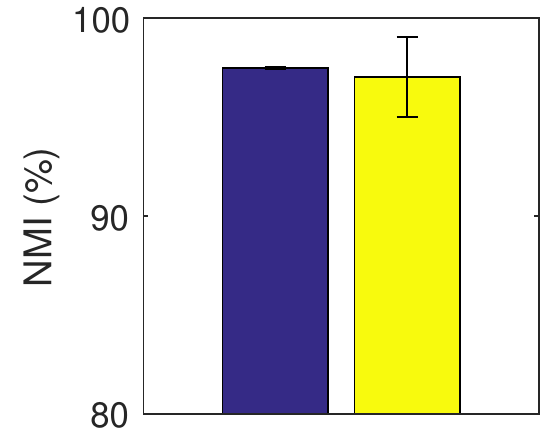}
&\includegraphics[width=1.7cm]{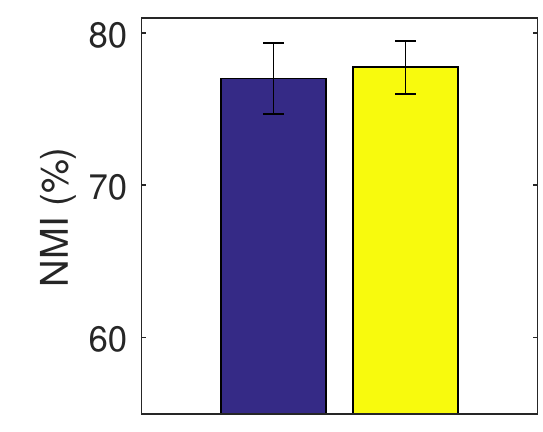}\\
CA
&\includegraphics[width=1.7cm]{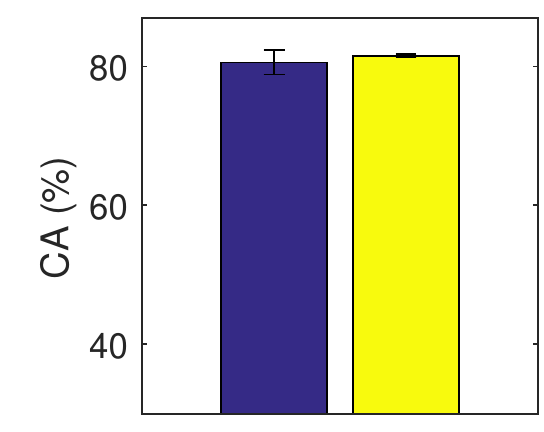}
&\includegraphics[width=1.7cm]{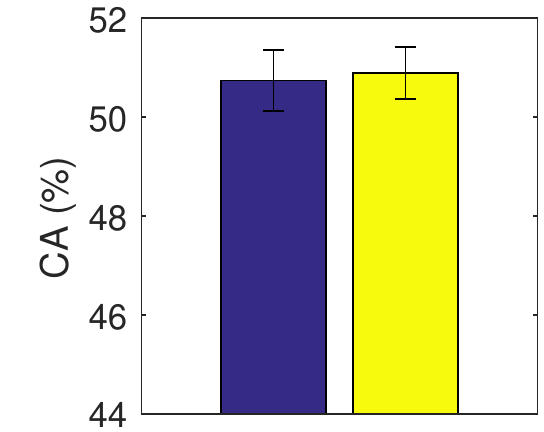}
&\includegraphics[width=1.7cm]{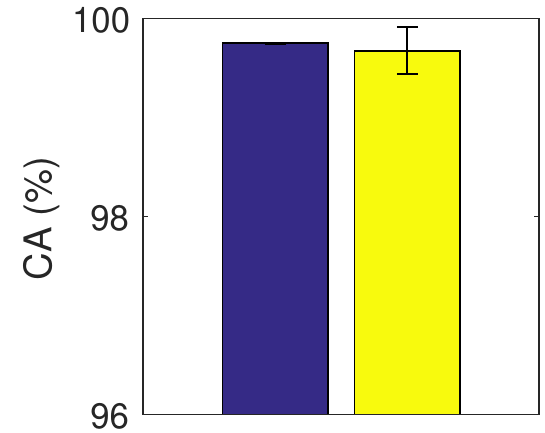}
&\includegraphics[width=1.7cm]{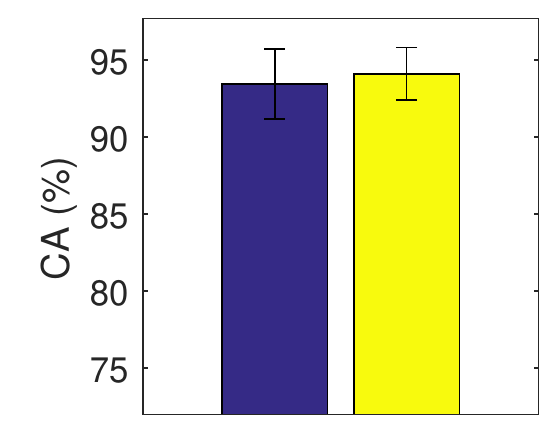}\\
Time cost
&\includegraphics[width=1.7cm]{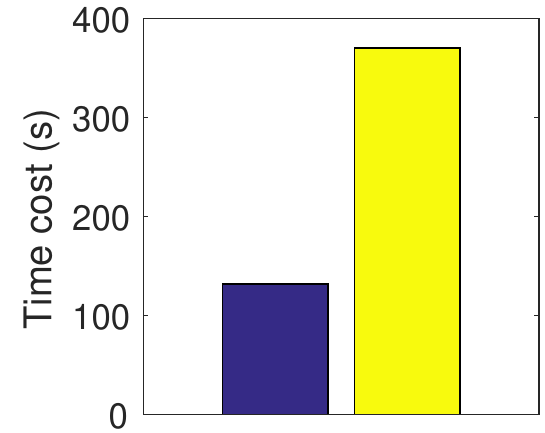}
&\includegraphics[width=1.7cm]{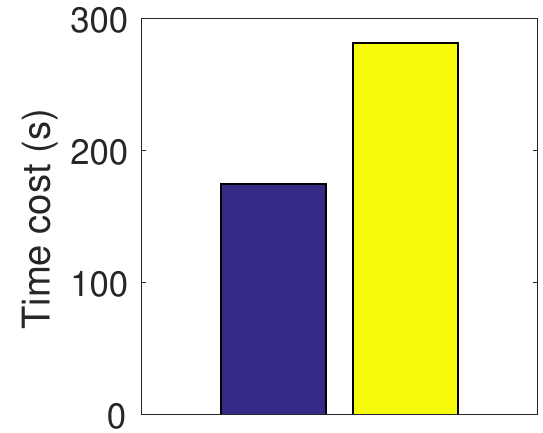}
&\includegraphics[width=1.7cm]{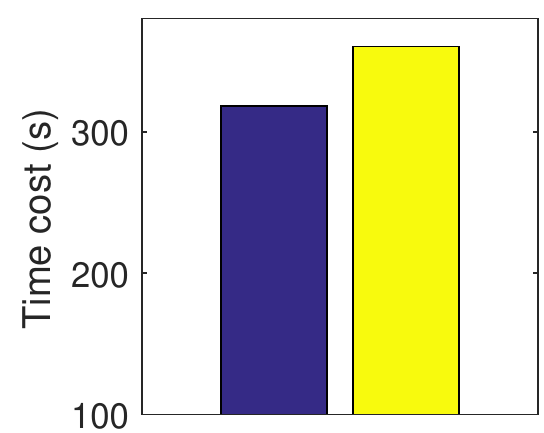}
&\includegraphics[width=1.7cm]{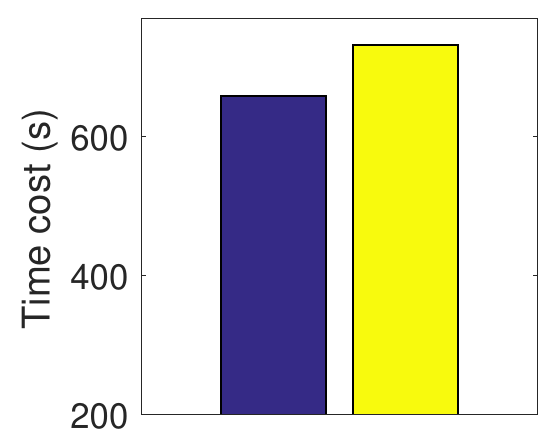}\\
&\multicolumn{4}{c}{\includegraphics[width=3.1cm]{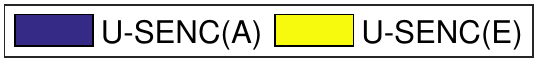}}\\
\bottomrule
\end{tabular}
\end{threeparttable}
\end{table}

\section{Conclusion}
\label{sec:conclusion}
This paper proposes two large-scale clustering algorithms, termed ultra-scalable spectral clustering (U-SPEC) and ultra-scalable ensemble clustering (U-SENC), respectively. In U-SPEC, a new hybrid representative selection strategy is designed to strike a balance between the efficiency of random selection and the effectiveness of $k$-means based selection. Then a new approximation method for $K$-nearest representatives is presented to efficiently construct a bipartite graph between the original data objects and the set of representatives, upon which the transfer cut can be utilized to obtain the clustering result. Starting from the U-SPEC algorithm, we further integrate multiple U-SPEC clusterers into a unified ensemble clustering framework and propose the U-SENC algorithm. Specifically, multiple U-SPEC's are exploited in the ensemble generation phase to produce an ensemble of diverse and high-quality base clusterings. The multiple base clusterings are incorporated into a new bipartite graph, which treats both objects and base clusters as graph nodes and is then efficiently partitioned to achieve the final consensus clustering. Extensive experiments have been conducted on ten large-scale datasets, which demonstrate the scalability and robustness of our algorithms.

\ifCLASSOPTIONcompsoc
  \section*{Acknowledgments}
\else
  \section*{Acknowledgment}
\fi

This project was supported by NSFC (61602189, 61876193 \& 61876104), National Key Research and Development Program of China (2016YFB1001003), and Guangdong Natural Science Funds for Distinguished Young Scholars (2016A030306014).

\ifCLASSOPTIONcaptionsoff
  \newpage
\fi

\bibliographystyle{IEEEtran}
\bibliography{tkde_2018}

\begin{IEEEbiography}[{\includegraphics[width=1in,height=1.25in,clip,keepaspectratio]{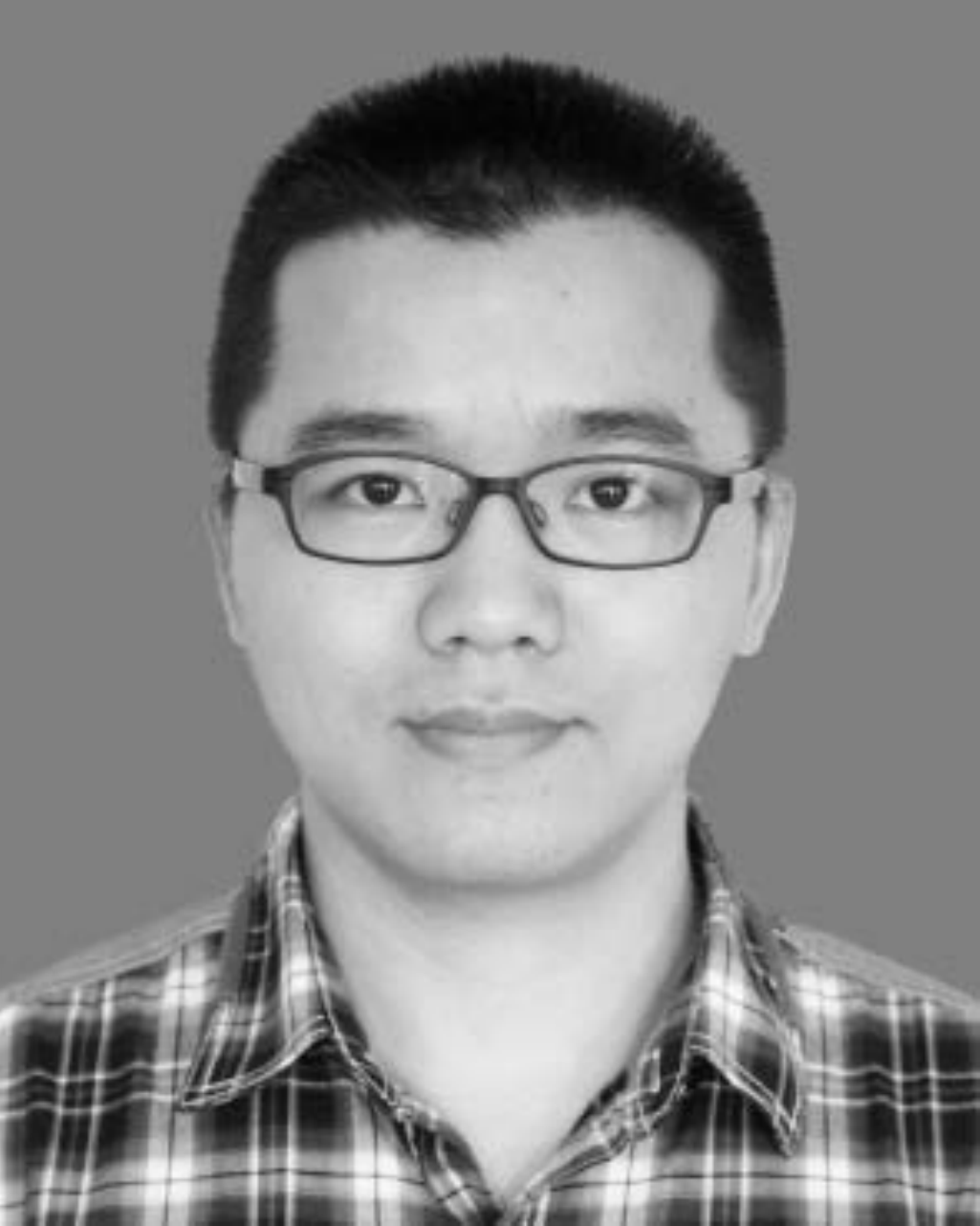}}]{Dong Huang}
received the B.S. degree in computer science from South China University of Technology, Guangzhou, China, in 2009, and the M.Sc. and Ph.D. degrees in computer science from Sun Yat-sen University, Guangzhou, in 2011 and 2015, respectively. He joined South China Agricultural University in 2015, where he is currently an Associate Professor with the College of Mathematics and Informatics. From July 2017 to July 2018, he was a visiting fellow with the School of Computer Science and Engineering, Nanyang Technological University, Singapore. His research interests are in the areas of data mining and machine learning, and more specifically focus on ensemble clustering, multi-view clustering, and large-scale clustering.
\end{IEEEbiography}

\begin{IEEEbiography}[{\includegraphics[width=1in,height=1.25in,clip,keepaspectratio]{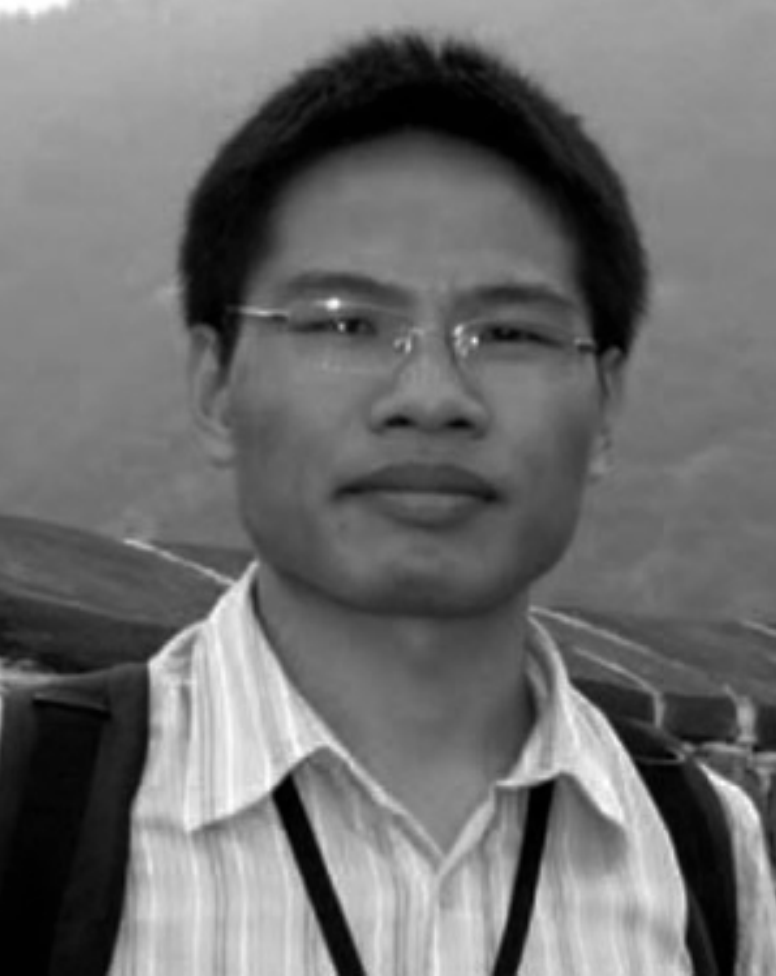}}]{Chang-Dong Wang}
received the Ph.D. degree in computer
science in 2013 from Sun Yat-sen University,
China. He is a visiting student at University of Illinois at Chicago
from Jan. 2012 to Nov. 2012. He joined Sun Yat-sen University in
2013 as an assistant professor and now he is currently an associate professor with School of Data and Computer Science.
His research interests include machine learning
and data mining. He has published over 90 papers in
international journals and conferences such as IEEE TPAMI, IEEE
TKDE, IEEE TCYB, IEEE TSMC-C, Pattern Recognition, KAIS, AAAI, ICDM, CIKM and SDM. His ICDM 2010 paper won the Honorable Mention for Best
Research Paper Awards. He won 2012 Microsoft Research Fellowship Nomination Award. He was awarded 2015 Chinese Association for Artificial Intelligence (CAAI) Outstanding Dissertation.
\end{IEEEbiography}

\begin{IEEEbiography}[{\includegraphics[width=1in,height=1.25in,clip,keepaspectratio]{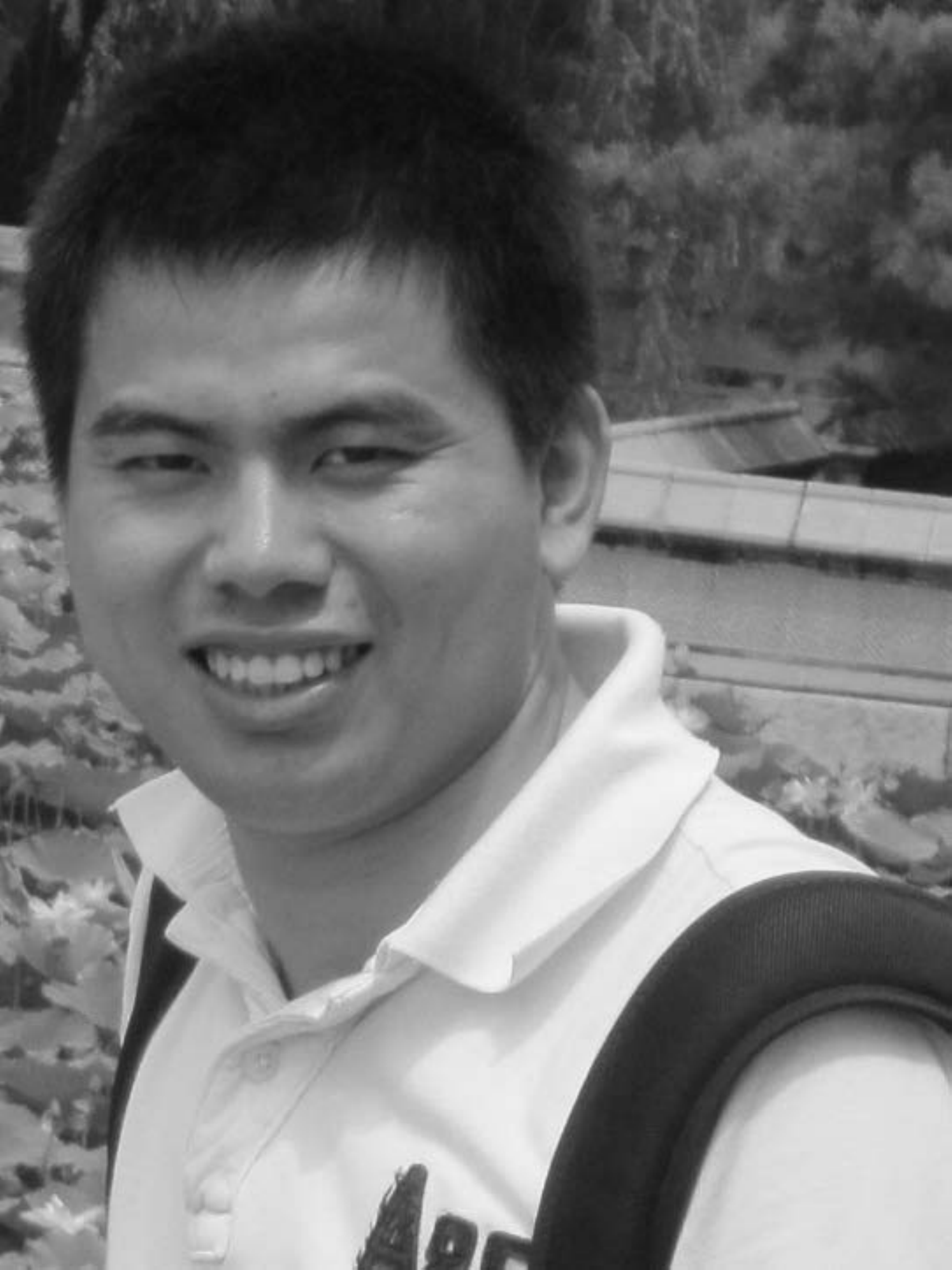}}]{Jian-Sheng Wu}
received the B.S. and Ph.D. degrees from the School of Information Science and Technology from Sun Yat-sen University, Guangzhou, China, in 2009 and 2015, respectively. He joined the School of Information Engineering, Nanchang University, in 2015. His current research interests include machine learning and data mining, especially focusing on large scale data clustering.
\end{IEEEbiography}

\begin{IEEEbiography}[{\includegraphics[width=1in,height=1.25in,clip,keepaspectratio]{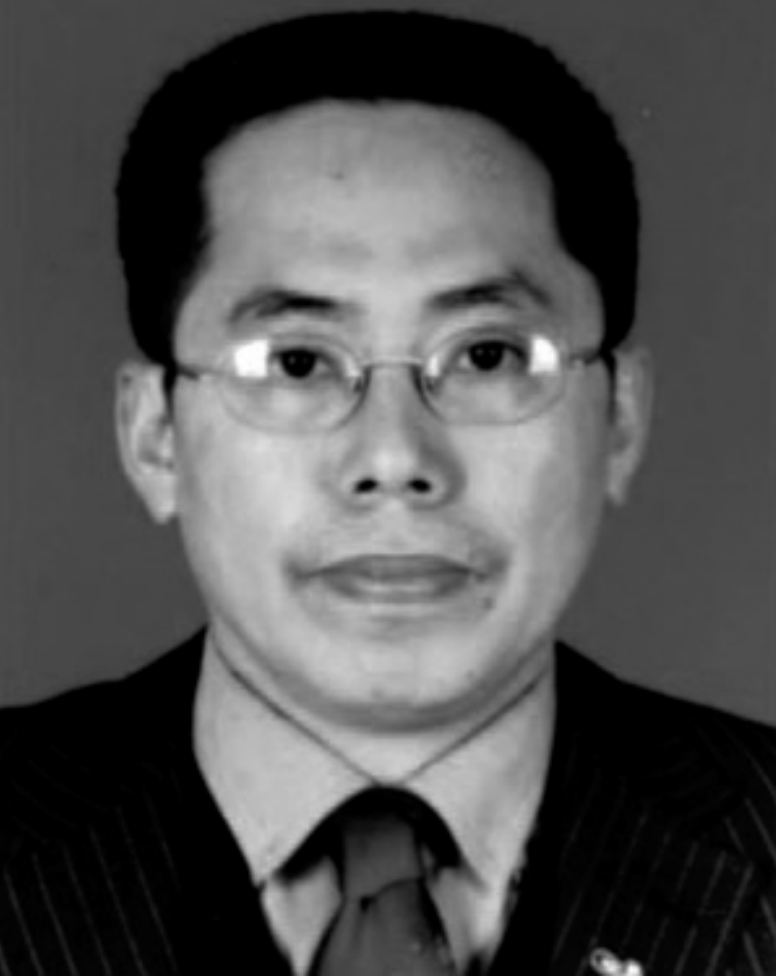}}]{Jian-Huang Lai}
received the M.Sc. degree in applied mathematics in 1989 and the Ph.D. degree in mathematics in 1999 from Sun Yat-sen University, China. He joined Sun Yat-sen University in 1989 as an Assistant Professor, where he is currently a Professor with the School of Data and Computer Science. His current research interests include the areas of digital image processing, pattern recognition, multimedia communication, wavelet and its applications. He has published more than 200 scientific papers in the international journals and conferences on image processing and pattern recognition, such as IEEE TPAMI, IEEE TKDE, IEEE TNN, IEEE TIP, IEEE TSMC-B, Pattern Recognition, ICCV, CVPR, IJCAI, ICDM and SDM. Prof. Lai serves as a Standing Member of the Image and Graphics Association of China, and also serves as a Standing Director of the Image and Graphics Association of Guangdong.
\end{IEEEbiography}

\begin{IEEEbiography}[{\includegraphics[width=1in,height=1.25in,clip,keepaspectratio]{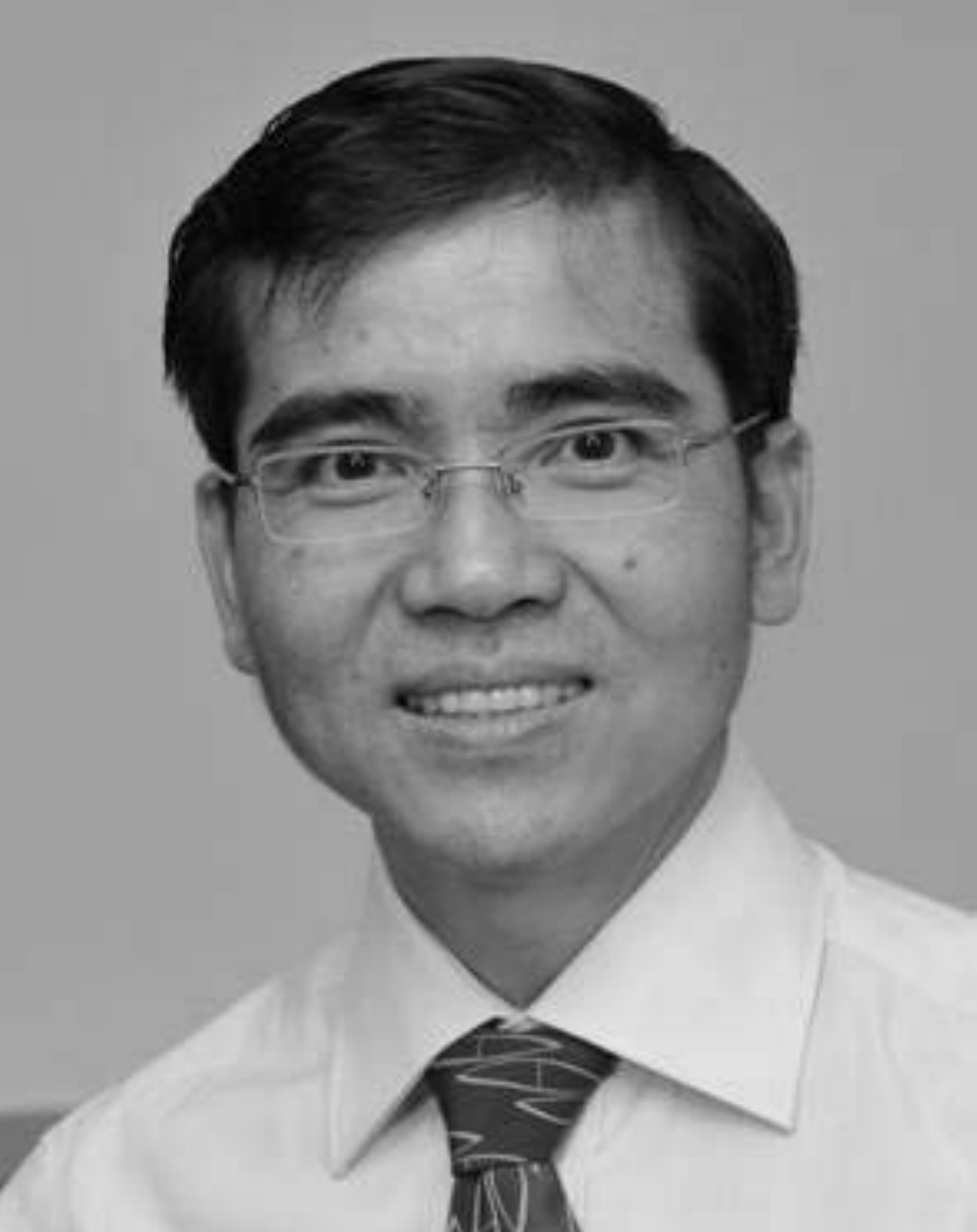}}]{Chee-Keong Kwoh} received the bachelor's degree in electrical engineering (first class) and the master's degree in industrial system engineering from the National University of Singapore in 1987 and 1991, respectively. He received the PhD degree from the Imperial College of Science, Technology and Medicine, University of London, in 1995. He has been with the School of Computer Science and Engineering, Nanyang Technological University (NTU) since 1993. He is the programme director of the MSc in Bioinformatics programme at NTU. His research interests include data mining, soft computing and graph-based inference; applications areas include bioinformatics and biomedical engineering. He has done significant research work in his research areas and has published many quality international conferences and journal papers. He is an editorial board member of the International Journal of Data Mining and Bioinformatics; the Scientific World Journal; Network Modeling and Analysis in Health Informatics and Bioinformatics; Theoretical Biology Insights; and Bioinformation. He has been a guest editor for many journals such as Journal of Mechanics in Medicine and Biology, the International Journal on Biomedical and Pharmaceutical Engineering and others. He has often been invited as an organizing member or referee and reviewer for a number of premier conferences and journals including GIW, IEEE BIBM, RECOMB, PRIB, BIBM, ICDM and iCBBE. He is a member of the Association for Medical and Bio-Informatics, Imperial College Alumni Association of Singapore. He has provided many services to professional bodies in Singapore and was conferred the Public Service Medal by the President of Singapore in 2008.
\end{IEEEbiography}

\end{document}